\documentclass[journal]{IEEEtran}
\usepackage{algorithm,algorithmic,amsbsy,amsmath,amssymb,epsfig,bbm,mathrsfs,fancyhdr,fancyvrb,subfigure,url,cite,multirow}
\usepackage{graphicx}
\usepackage{epstopdf}
\usepackage{setspace}
\usepackage{graphicx}
\usepackage{lscape}
\usepackage{multirow}
\usepackage{bbding}
\usepackage{array}
\usepackage{caption}
\usepackage{bm}
\allowdisplaybreaks
\usepackage{array}

\usepackage{booktabs}

\usepackage[table,xcdraw]{xcolor}

\newcommand{\beq}{\begin{equation}}
\newcommand{\eeq}{\end{equation}}
\newcommand{\beqn}{\begin{eqnarray}}
\newcommand{\eeqn}{\end{eqnarray}}

\usepackage{amsmath}

\providecommand{\theoremname}{\textbf{Theorem}}
\providecommand{\propositionname}{\textbf{Proposition}}
\providecommand{\remarkname}{\textbf{Remark}}
\providecommand{\lemmaname}{\textbf{Lemma}}
\providecommand{\corollaryname}{\textbf{Corollary}}
\providecommand{\Definition}{\textbf{Definition}}

\begin{document}

\title{Applications of Multi-Agent Reinforcement Learning in Future Internet: A Comprehensive Survey}
\author{Tianxu Li, Kun Zhu,~\IEEEmembership{Member,~IEEE}, Nguyen Cong Luong,~\IEEEmembership{Member,~IEEE}, Dusit Niyato,~\IEEEmembership{Fellow,~IEEE}, \\Qihui Wu,~\IEEEmembership{Senior Member,~IEEE}, Yang Zhang,~\IEEEmembership{Member,~IEEE}, and Bing Chen,~\IEEEmembership{Member,~IEEE}
%\thanks{This work is supported by National Natural Science Foundation of China under Grant 61701230 and Grant BK20170805, and the Fundamental Research Funds for the Central Universities under NE2018107.}
\thanks{T. X. Li, K. Zhu, Q. H. Wu, Y. Zhang and B. Chen are with the College of Computer Science and Technology, Nanjing University of Aeronautics and Astronautics, Nanjing 210016, China (email: \{tianxuli, zhukun, wuqihui, yangzhang, cb\_china\}@nuaa.edu.cn).}
\thanks{N. C. Luong is with the Faculty of Computer Science, PHENIKAA University, Hanoi 12116, Vietnam (email: luong.nguyencong@phenikaa-uni.edu.vn).}
\thanks{D. Niyato is with School of Computer Science and Engineering, Nanyang Technological University, Singapore 639798 (email: dniyato@ntu.edu.sg).}
}\maketitle

\begin{abstract}
Future Internet involves several emerging technologies such as 5G and beyond 5G networks, vehicular networks, unmanned aerial vehicle (UAV) networks, and Internet of Things (IoTs). Moreover, the future Internet becomes heterogeneous and decentralized with a large number of involved network entities. Each entity may need to make its local decision to improve the network performance under dynamic and uncertain network environments. Standard learning algorithms such as single-agent Reinforcement Learning (RL) or Deep Reinforcement Learning (DRL) have been recently used to enable each network entity as an agent to learn an optimal decision-making policy adaptively through interacting with the unknown environments. However, such an algorithm fails to model the cooperations or competitions among network entities, and simply treats other entities as a part of the environment that may result in the non-stationarity issue. Multi-agent Reinforcement Learning (MARL) allows each network entity to learn its optimal policy by observing not only the environments but also other entities' policies. As a result, MARL can significantly improve the learning efficiency of the network entities, and it has been recently used to solve various issues in the emerging networks. In this paper, we thus review the applications of MARL in emerging networks. In particular, we provide a tutorial of MARL and a comprehensive survey of applications of MARL in next-generation Internet. In particular, we first introduce single-agent RL and MARL. Then, we review a number of applications of MARL to solve emerging issues in the future Internet. The issues consist of network access, transmit power control, computation offloading, content caching, packet routing, trajectory design for UAV-aided networks, and network security issues. Finally, we discuss the challenges, open issues, and future directions related to the applications of MARL in the future Internet.
\end{abstract}


\section{Introduction}

Future Internet involves various emerging networks such as 5G and beyond 5G networks, vehicular networks, unmanned aerial vehicle (UAV) networks, and Internet of Things (IoTs). Such networks usually have heterogeneous devices, dense users, frequent communication requests, and ubiquitous channel interference, which imposes significant challenges on improving the Quality of Service (QoS). To meet the rapidly growing QoS demands of network entities, e.g., mobile users, vehicles, UAVs, and IoT devices, and establish low-latency, ultra-reliable, and energy-efficient networks, Reinforcement Learning (RL)~\cite{sutton2018reinforcement} and Deep Reinforcement Learning (DRL)~\cite{li2017deep}, that combines Deep Neural Networks (DNNs) with RL, have been proposed as effective tools to provide Artificial Intelligence (AI)-enabled network solutions for key issues in the future internet such as network access~\cite{wang2018deep}, transmit power control~\cite{chen2018dqn}, task offloading~\cite{liu2018deepnap}, and content caching~\cite{he2017deep}. In particular, it enables the network entity as an agent to learn an optimal decision-making policy by interacting with dynamic and uncertain network environments, which is usually modeled as a Markov Decision Process (MDP)~\cite{Puterman1994MarkovDP}. However, future networks typically involve multiple network entities interacting with each other, and thus the single-agent RL or DRL may not be efficient and even applicable. The reason is that the single-agent RL or DRL can be only used for a single agent to learn its decision-making policy independently and simply treat other agents as a part of the environment. This means that the policy of the network entity, i.e., the agent, learned by the single-agent RL or DRL does not consider the impact of the policies of other network entities. This may cause the non-stationarity issue~\cite{hernandez2017survey} and significantly reduce the learning efficiency as well as network performance. 

In real-world network environments, each network entity typically needs to autonomously make local action decisions (e.g., the channel selection) depending on its local observation (e.g., the local Channel State Information (CSI)) since it may not observe the instantaneous global network state, especially in large scale networks. With the partial observation, the independent learning network entity will struggle to account for the policies of other entities and address the non-stationarity issue. Furthermore, cooperation among network entities is common in communication networks, which is of great significance for the establishment of decentralized and self-organizing networks. For example, cellular users may need to collaborate with other users to maximize the global network throughput. This motivates the applications of Multi-Agent Reinforcement Learning (MARL)~\cite{zhang2021multi} including Multi-Agent Deep Reinforcement Learning (MADRL) in the area of future Internet. MARL is an extension of RL to multi-agent environments to address the sequential decision-making problem involving multiple learning agents. In general, MARL is able to provide the following advantages:

\begin{itemize}

\item MARL can solve complex optimization issues in the area of future Internet such as transmit power control which is typically NP-hard and non-convex. Thus, with MARL, the network entities can learn optimal solutions to improve their network performance.

\item MARL enables the network entity to overcome the non-stationarity issue by taking other entities' states and actions into account. Thus, the network entity can learn a more stable policy with MARL compared with single-agent RL and DRL that do not account for other network entities' information.

\item MARL enables the network entity to learn the collaboration by communicating with other entities. Thus, the network entity such as a mobile user is capable of cooperating with other entities by using MARL to learn an optimal policy, e.g., channel selection, transmit power selection, and offloading and caching decisions, so as to achieve the global objective such as maximizing global network throughput or minimizing the overall task execution and content access latency.

\item MARL can provide decentralized decision-making solutions to the network entity through global optimization, e.g., through the centralized training with decentralized execution (CTDE) framework. Thus, the network entity only needs its local observation to make action decisions without the need to exchange information with other entities. This can significantly reduce the signaling overhead.

\item Some networking issues in the future internet such as spectrum access and computation offloading can be modeled as a non-cooperative game. MARL enables the network entity to achieve a Nash equilibrium in such a game by taking other competitors' actions into account. Thus, with MARL, each network entity can learn a non-cooperative policy that can maximize its own network performance, e.g., its own throughput.

\end{itemize}

\begin{table*}[]\scriptsize\centering
\renewcommand\arraystretch{1.5}
\caption{A summary of existing surveys on deep learning, DRL, and MARL in the area of future Internet}
\label{Survey}
\begin{tabular}{|c|m{10cm}<{\centering}|c|c|c|}
\hline

\multirow{2}{*}{\textbf{Work}} & \multirow{2}{*}{\textbf{Brief summary}} & \multicolumn{3}{c|}{\textbf{Scope}} \\ \cline{3-5}
                      &                                & deep learning    & DRL    & MARL                     \\ \hline
     ~\cite{Zhou2018IntelligentWC}                 &  A survey on the applications of machine learning for cognitive radio                            &  \CheckmarkBold                & \XSolidBrush       &  \XSolidBrush                              \\ \hline
      ~\cite{ota2017deep}                 &         A survey on the applications of deep learning for mobile multimedia                   &  \CheckmarkBold                &    \XSolidBrush    &  \XSolidBrush                              \\ \hline
     ~\cite{Zhang2019DeepLI}                 &      A survey on deep learning applications in mobile and wireless networking                          &   \CheckmarkBold               &  \CheckmarkBold      &      \XSolidBrush                          \\ \hline
    
     ~\cite{mohammadi2018deep}              &    A survey on the deep learning for IoT big data and streaming analytics                 &   \CheckmarkBold               &  \CheckmarkBold       &  \XSolidBrush                            \\ \hline
     ~\cite{mao2018deep}                 &    A comprehensive survey on the applications of deep learning in intelligent wireless networks                      &   \CheckmarkBold               &  \CheckmarkBold      &  \XSolidBrush                              \\ \hline
     ~\cite{qian2019survey}                 &    A survey on RL applications in communication networks                    &    \CheckmarkBold               &   \CheckmarkBold     &   \XSolidBrush                             \\ \hline
     ~\cite{9069178}                  &   An overview of model, applications and challenges of DRL for autonomous IoTs               &       \CheckmarkBold            &  \CheckmarkBold      &   \XSolidBrush                             \\ \hline
     ~\cite{8553654}                  &  A survey on DRL for mobile edge caching                              &       \CheckmarkBold            &   \CheckmarkBold     &  \XSolidBrush                             \\ \hline
     ~\cite{Luong2019ApplicationsOD}                  &   A comprehensive survey on the applications of DRL in communications and networking                            &     \CheckmarkBold             &  \CheckmarkBold      &   \XSolidBrush                             \\ \hline
     ~\cite{8766739}                  &      An overview of the potential applications of MARL in Vehicular Networks                          & \CheckmarkBold                  &   \CheckmarkBold     &  \CheckmarkBold                              \\ \hline
     Our survey                &      A comprehensive survey on applications of MARL in the future internet                          &     \CheckmarkBold              &    \CheckmarkBold    &  \CheckmarkBold                              \\ \hline
\end{tabular}
\end{table*}

\subsection{Related Existing Surveys and Scope of Our Survey}

\begin{figure*}[h]
    \centering
    \includegraphics[scale=0.6]{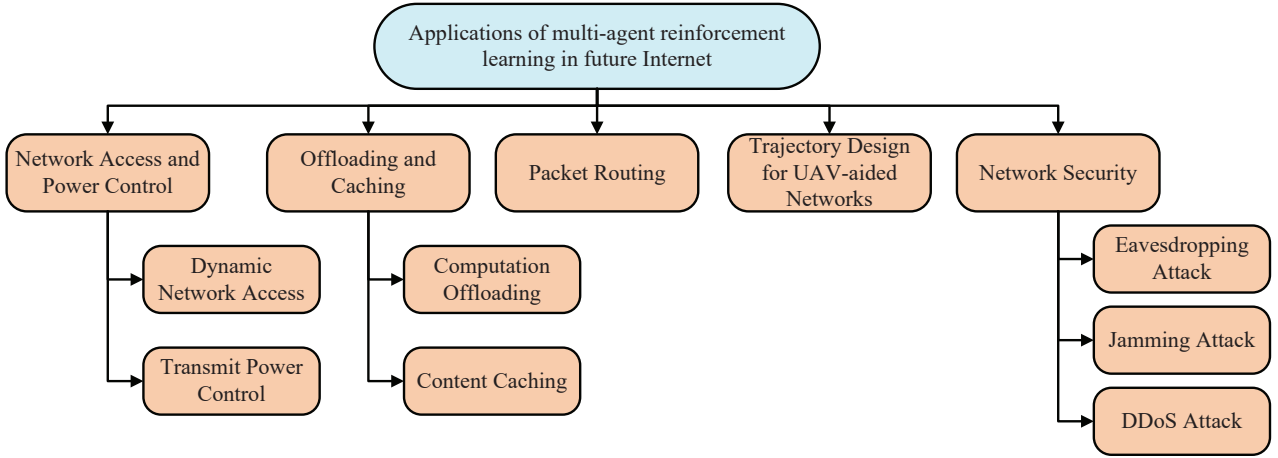}
    \caption{A classification of the applications of MARL in the future internet.}
    \label{classification}
\end{figure*}

Given the aforementioned advantages, MARL has been widely used to solve the key issues in the area of future Internet. However, to the best of our knowledge, there is no comprehensive survey reviewing the MARL-related applications in the future internet. As presented in Table \ref{Survey}, most of the related surveys discuss the applications of deep learning and single-agent DRL for the future Internet. For example, the authors in~\cite{Zhang2019DeepLI} provide a comprehensive survey on the applications of deep learning in mobile and wireless networking. Other surveys discuss the specific applications of deep learning for cognitive radio~\cite{Zhou2018IntelligentWC}, mobile multimedia~\cite{ota2017deep}, and IoT big data~\cite{mohammadi2018deep}. The authors in~\cite{Luong2019ApplicationsOD} present a comprehensive survey on the applications of single-agent DRL to solve the issues in the area of communications and networking. The authors in~\cite{8766739} provide some potential applications of MARL in vehicular networks, but they do not review the related works. Different from the existing surveys, our paper focuses on reviewing, analyzing, and comparing existing works of literature related to the applications of MARL to solve key issues in the future internet such as dynamic network access, transmit power control, computation offloading, content caching, packet routing, trajectory design for UAV-aided networks, and network security, which are summarized in Fig. \ref{classification}.

\begin{figure*}
\centering
\subfigure[]{
\begin{minipage}[t]{0.27\linewidth}
\centering
\includegraphics[width=2.0in]{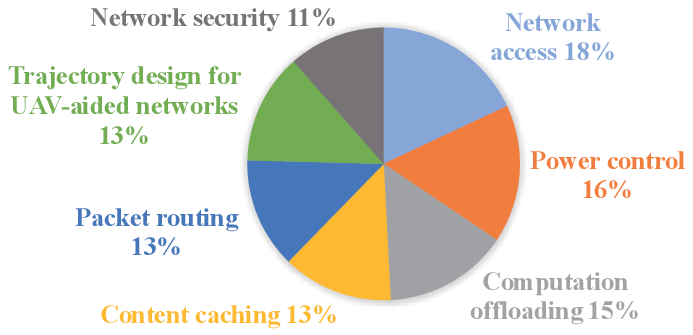}
%\caption{fig1}
\end{minipage}%
}%
\subfigure[]{
\begin{minipage}[t]{0.23\linewidth}
\centering
\includegraphics[width=1.7in]{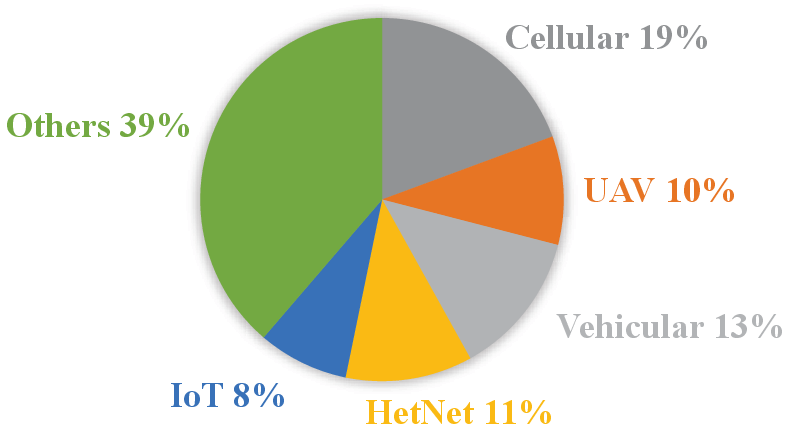}
%\caption{fig2}
\end{minipage}%
}%
\subfigure[]{
\begin{minipage}[t]{0.25\linewidth}
\centering
\includegraphics[width=1.9in]{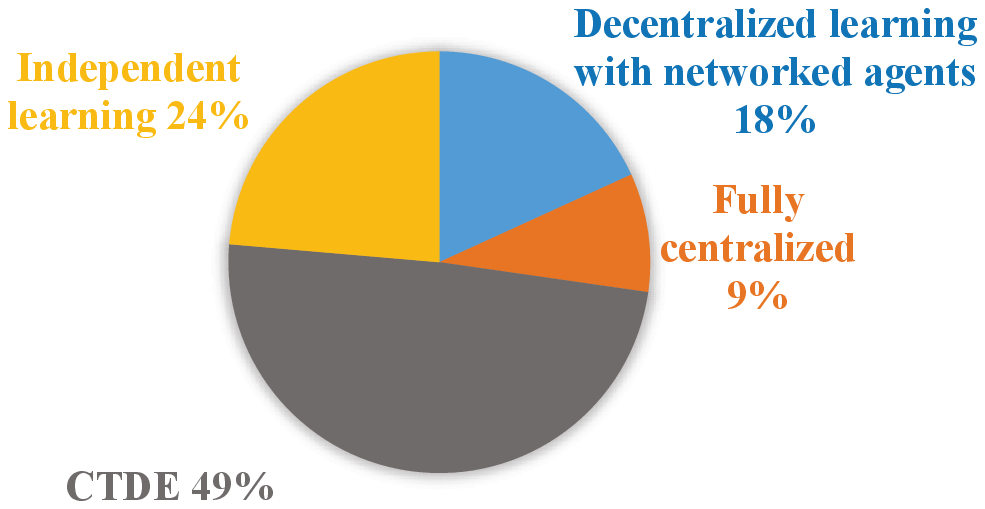}
%\caption{fig2}
\end{minipage}
}%
\subfigure[]{
\begin{minipage}[t]{0.25\linewidth}
\centering
\includegraphics[width=1.23in]{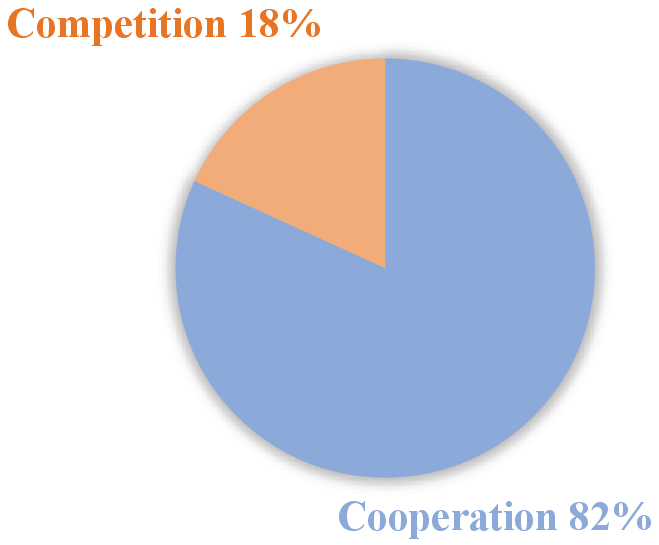}
%\caption{fig2}
\end{minipage}
}%
\centering
%\caption{}
\caption{Percentages of MARL-related works for (a) different issues, (b) different networks, with (c) different learning schemes, and (d) different learning settings.}
\label{Percentage}
\end{figure*}

The statistics in terms of the percentages of MARL-related works for different issues and different networks are shown in Fig. \ref{Percentage}(a) and Fig. \ref{Percentage}(b), respectively. Moreover, we show the percentages of related works with different learning schemes and settings in Fig. \ref{Percentage}(c) and Fig. \ref{Percentage}(d). As seen, network access and transmit power control issues receive more attention than other issues, and applications of MARL in cellular systems are common. Also, from Fig. \ref{Percentage}(c) and Fig. \ref{Percentage}(d), we observe that most works adopt the CTDE learning scheme and the cooperation setting to train network entities.

\subsection{Contributions and Organization of This Paper}
Our contributions are summarized as follows:

\begin{itemize}
\item We provide an introduction on single-agent RL. We present the definitions of MDPs and Partially Observable Markov Decision Processes (POMDPs) and demonstrate that they are not sufficient to model the sequential decision-making process involving multiple network entities, i.e., agents. The advantages and shortcomings of value-based or policy-based algorithms are analyzed, and we give some examples to show the potential of using these algorithms as the basic models to establish the corresponding multi-agent solutions to solve the networking issues in the future internet.

\item We provide a tutorial on MARL. We first discuss the challenges of multi-agent environments such as non-stationarity, scalability, and partial observability. Then, we provide and discuss different MARL learning schemes to overcome these challenges. We summarize and analyze the widely used algorithms for solving networking issues in the future internet, which are typically based on CTDE. We further discuss the value-decomposition MARL algorithms, which can be efficiently used to solve the credit assignment issue among multiple network entities, but they are rarely used in the existing works of literature related to networking.

\item We review the works of literature on the applications of MARL for solving emerging issues in the future internet. We further discuss the advantages as well as shortcomings of the MARL solutions.

\item We present the challenges, open issues, and future research directions on the applications of MARL for future Internet

\end{itemize}

\begin{figure*}[h]
    \centering
    \includegraphics[scale=0.6]{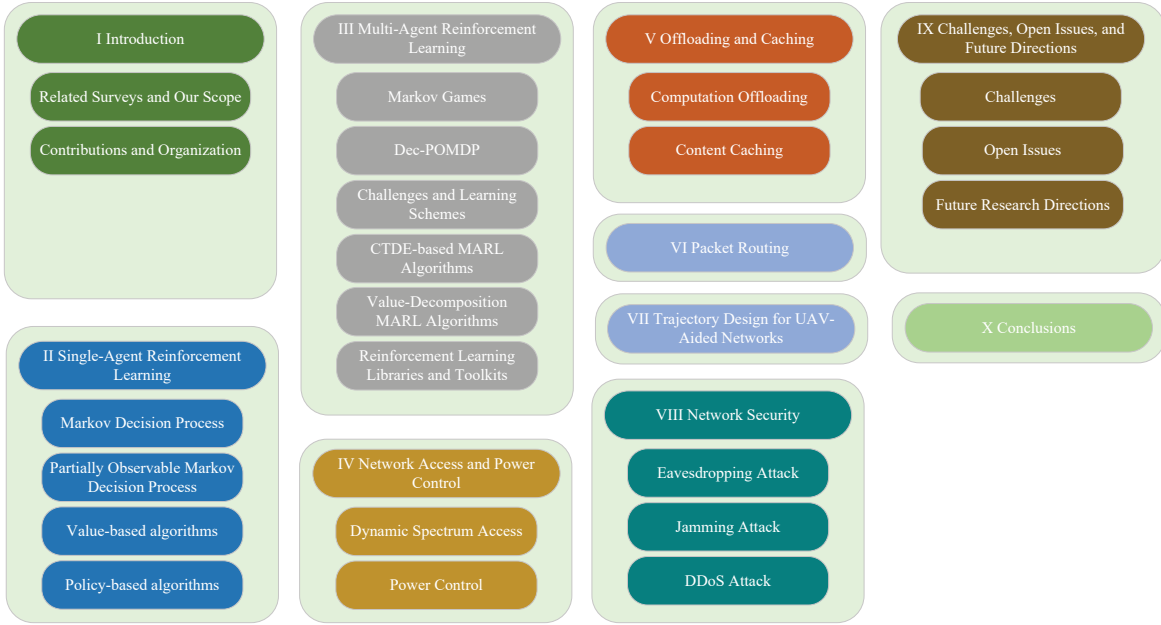}
    \caption{Structure of the survey.}
    \label{Structure}
\end{figure*}

The structure of our survey is shown in Fig. \ref{Structure}. The rest of this paper is organized as follows. We introduce the fundamentals of single-agent RL and MARL in Section \ref{single-agent RL} and Section \ref{MARL}, respectively. We present the applications of MARL for network access and power control in Section \ref{Network Access and Power Control}. Section \ref{Offloading and Caching} presents the applications of MARL for computation offloading and content caching. Section \ref{packet routing} reviews the MARL-related works for packet routing in various networks. The applications of MARL for trajectory design in UAV-aided networks and network security are discussed in Section \ref{Trajectory Design for UAVs} and Section \ref{Network Security}, respectively. In section \ref{Challenges_Open}, we discuss the challenges, open issues, and future research directions of applications of MARL for the future Internet. The conclusions are given in Section \ref{Conclusions}. For convenience, we provide a summary of abbreviations commonly used in this paper in Table \ref{tab:Abbreviation}.

\begin{table}[]\scriptsize\centering
\renewcommand\arraystretch{1.5}
\caption{Summary of Abbreviations}
\label{tab:Abbreviation}
\begin{tabular}{|m{2cm}<{\centering}|m{5.5cm}<{\centering}|}
\hline
\rowcolor[HTML]{9B9B9B}
\textbf{Abbreviation} & \textbf{Description}                                                       \\ \hline
A2C/A3C               & Advantage Actor-Critic/Asynchronous Advantage Actor-Critic                 \\ \hline
AoI                   & Age of Information                                                         \\ \hline
COMA                  & Counterfactual Multi-Agent Policy Gradients                                \\ \hline
CRN                    & Cognitive Radio Network                                                            \\ \hline
CSI                   & Channel State Information                                                  \\ \hline
CTDE                  & Centralized Training with Decentralized Execution                          \\ \hline
D2D                   & Device to Device                                                           \\ \hline
DDoS                   & Distributed Denial of Service                                                          \\ \hline
Dec-POMDP             & Decentralized Partially Observable Markov Decision Process                 \\ \hline
DQN/DDQN/D3QN         & Deep Q Network/Double Deep Q network/Dueling Double Deep Q Network         \\ \hline
DNN/CNN/GNN           & Deep Neural Network/Convolutional Neural Network/Graph Neural Network      \\ \hline
DPG/DDPG              & Deterministic policy gradient/Deep Deterministic Policy Gradient           \\ \hline
E2E                   & End-to-end                                                                 \\ \hline
HetNet                & Heterogeneous Network                                                      \\ \hline
IL                    & Independent Learning                                                       \\ \hline
IoT/IIoT              & Internet of Things/Industrial Internet of Things                           \\ \hline
LoS/NLoS              & Line-of-Sight/Non Line-of-Sight                                            \\ \hline
LSTM                  & Long Short-Term Memory                                                     \\ \hline
MAAC                  & Multi-Actor Attention-Critic                                               \\ \hline
MADDPG                & Multi-Agent Deep Deterministic Policy Gradient                             \\ \hline
MADRL                 & Multi-Agent Deep Reinforcement Learning                                    \\ \hline
MANET                 & Mobile Ad-hoc Network                                                      \\ \hline
MARL/single-agent RL             & Multi-Agent Reinforcement Learning/Single-Agent Reinforcement Learning     \\ \hline
MBS/FBS/PBS/SBS       & Macro Base Station/Femto Base Station/Pico Base Station/Small Base Station \\ \hline
MDP                   & Markov Decision Process                                                    \\ \hline
MEC                   & Mobile Edge Computing                                                      \\ \hline
MF-Q                  & Mean-Field $Q$-Learning                                                      \\ \hline
MG/SG                 & Markov Game/Stochastic Game                                                \\ \hline
MMDP                  & Multi-agent Markov Decision Process                                        \\ \hline
NB-IoT                 & Narrow Band Internet of Things                                             \\ \hline
NE                    & Nash equilibrium                                                           \\ \hline
PN/SN                 & Primary Network/Secondary Network                                          \\ \hline
POMDP                 & Partially Observable Markov Decision Process                               \\ \hline
PPO                   & Proximal Policy Optimization                                               \\ \hline
QoS                   & Quality of Service                                                         \\ \hline
RL                    & Reinforcement Learning                                                     \\ \hline
SAC                   & Soft Actor Critic                                                          \\ \hline
SINR                  & Interference Plus Noise Ratio                                              \\ \hline
TD/SD                 & Temporal Difference/Spatial Difference                                     \\ \hline
TDMA/NOMA             & Time Division Multiple Access/Non-Orthogonal Multiple Access               \\ \hline
TRPO                  & Trust Region Policy Optimization                                           \\ \hline
UAV                   & Unmanned Aerial Vehicle                                                    \\ \hline
UDN                   & Ultra-Dense Network                                                        \\ \hline
UE/GU/GT              & User/Ground User/Ground Terminal                                           \\ \hline
V2V/V2I               & Vehicle-to-Vehicle/Vehicle-to-infrastructure                               \\ \hline
\end{tabular}
\end{table}

\section{Single-Agent Reinforcement Learning}\label{single-agent RL}
Some single-agent RL algorithms can be generally used in multi-agent environments with a fully centralized learning setting or a fully decentralized learning setting. Thus, we first present the backgrounds of single-agent RL including MDPs and POMDPs. Then, we introduce some classical value-based and policy-based RL algorithms. For more details, we refer the readers to existing books and surveys, i.e.,~\cite{sutton2018reinforcement},~\cite{Kaelbling1996ReinforcementLA},~\cite{Arulkumaran2017ABS},~\cite{FranoisLavet2018AnIT}, related to the reinforcement learning.

\subsection{Markov Decision Process}
The sequential decision-making process of an agent in the environment is formulated as the Markov Decision Process (MDP)~\cite{Puterman1994MarkovDP}. An MDP is typically defined by a tuple $\langle\mathcal{S}, \mathcal{A}, R, T, \gamma\rangle$ where $\mathcal{S}$ represents a finite set of states, $\mathcal{A}$ is a finite set of actions. $T: \mathcal{S} \times \mathcal{A} \times \mathcal{S} \rightarrow[0,1]$ is the transition function that denotes the probability of a transition from state $s \in \mathcal{S}$ to next state $s^{\prime} \in \mathcal{S}$ after an action $a \in \mathcal{A}$ is executed. The reward function $R: \mathcal{S} \times \mathcal{A} \times \mathcal{S} \rightarrow \mathbb{R}$ determines the immediate reward that the agent receives after executing an action $a \in \mathcal{A}$ given state $s$. $\gamma \in [0,1]$ is a discount factor that can be used to trade-off the immediate rewards and the future rewards.

MDPs have been widely used to model the decision-making processes of agents with full observability of the system state. At the beginning of each time slot $t$, the agent takes an action $a_{t}$ by following a policy $\pi: \mathcal{S} \rightarrow \mathcal{A}$, which can map a state to an action. The objective of each agent under the classical infinite-horizon setting is to find an optimal policy $\pi^{*}$ that can maximize the expected discounted long-term reward defined by $\sum_{t=0}^{\infty} \gamma r_{t}\left(s_{t}, a_{t}\right)$ where $a_{t} \sim \pi(s_{t})$. Dynamic programming methods such as value iteration and policy iteration algorithms~\cite{Bertsekas1995DynamicPA} can be used to learn the optimal policy. However, they typically require knowledge of the dynamics of the MDP model, e.g., the transition function and the reward function. In contrast, RL algorithms can learn and find an optimal policy without a knowledge of the dynamics of the model. The RL agent can simply learn the policy from the experiences obtained by interacting with the environment.

\subsection{Partially Observable Markov Decision Process}
The MDP operates under the assumption that the agent can observe the global state of the system. However, this assumption may not be satisfied in most real-world communication networks. For example, the router can only observe the traffic flow status of its connected links and cannot observe the global network flow status. Thus, the agent only has a partial observation of the environment to make action decisions. This can be modeled by the POMDP~\cite{Monahan1982StateOT}, which is a generalization of the MDP to the environments with partial observations. A POMDP is represented by a 7-tuple $\langle\mathcal{S}, \mathcal{A}, R, T, \gamma, \mathcal{O}, Z\rangle$, where $\mathcal{S}, \mathcal{A}, R, T$, and $\gamma$ are defined similar to those in the MDP, $\mathcal{O}$ is denoted as the observation set, and $Z: \mathcal{S} \times \mathcal{A} \times \mathcal{O} \rightarrow[0,1]$ is the probability distribution over observations regarding to a state $s \in \mathcal{S}$ and an action $a \in \mathcal{A}$. This means that the agent will achieve a new observation $o \in \mathcal{O}$ with probability $Z\left(o \mid s, a, s^{\prime}\right)$ when the environment transits to the next state $s^{\prime} \in \mathcal{S}$ after an action $a \in \mathcal{A}$ is executed at the current state $s \in \mathcal{S}$.

\subsection{Value-based Algorithms}
The RL algorithms can be typically categorized into two types: the value-based algorithms and the policy-based algorithms. The value-based algorithms enable the agent to learn an optimal state-action value function estimate, also known as the optimal $Q$-function $Q^{*}$. As a result, the optimal policy can then be indirectly obtained by adopting the greedy action of the learned optimal $Q$-function $Q^{*}$. $Q$-learning~\cite{Watkins1989LearningFD} is one of the most popular value-based algorithms, where the agent maintains a $Q$-table and learns the $Q^{*}$ by constantly evaluating the $Q$-value $\hat{Q}(s, a)$ in the $Q$-table until convergence. In particular, when the agent executes an action $a$ at a state $s$, it receives a reward $r$ and updates the $\hat{Q}(s, a)$ as follows: 
\begin{equation}\label{qlearning}
\hat{Q}(s, a) \leftarrow \hat{Q}(s, a)+\alpha\left[\left(r+\gamma \max _{a^{\prime}} \hat{Q}\left(s^{\prime}, a^{\prime}\right)\right)-\hat{Q}(s, a)\right],
\end{equation} where the hyperparameter $\alpha \in \left[0,1\right]$ is the learning rate that controls the step size of the $Q$-value update. By using the $\alpha$ with certain conditions, the $Q$-learning algorithm with finite state and action spaces has been proved to converge to the $Q^{*}$~\cite{tsitsiklis1994asynchronous},~\cite{even-dar2004learning}. From Equation (\ref{qlearning}), we can see that $Q$-learning updates the $Q$-value per step via the Temporal Difference (TD) method~\cite{Tesauro1995TemporalDL}. TD method can be on-policy or off-policy. Particularly, $Q$-learning uses the off-policy TD that enables the agent to learn from the experiences of different policies. Thus, it can use the maximum $Q$-value of all the possible actions to update the current $Q$-value. In contrast, the on-policy value-based algorithm, e.g., SARSA, can only learn from the experiences of the current policy. Thus, it may suffer from poor sample efficiency since the experiences generated by other policies cannot be reused. The off-policy methods can also achieve a better exploration through visiting all the possible actions. In fact, the trade-off between exploration and exploitation is a challenge in RL algorithms. With $Q$-learning, the agent exploits the learned $Q$-function to choose the action with the maximum $Q$-value, but it also needs to explore other actions so as to improve its policy.

The $Q$-learning with the tabular setting may not efficiently address complicated problems with large state and action spaces since the $Q$-table can be difficult to maintain and slow to convergence under such conditions. To address the problem, recent works such as Deep $Q$-Network (DQN)~\cite{mnih2015human-level} apply the non-linear function approximators, e.g., DNNs, to estimate the $Q$-function. The DNN with weights $\phi$ used in DQN replaces the $Q$-table through using its excellent presentation characteristic, which can significantly improve the learning efficiency and has shown impressive results in communication networks. \textcolor{black}{Besides, an experience replay buffer $D$ is introduced to store the experience tuple $\left(s_{t}, a_{t}, r_{t}, s_{t+1}\right)$ for the purpose of efficient off-policy training.} To stabilize the agent learning process, a target $Q$-network $\bar{Q}$, which initiates the same weights with the primary $Q$-network, is leveraged to generate a stationarity learning target \begin{equation}
y=R(s, a)+\gamma \max _{a^{\prime}} \bar{Q}_{\phi}\left(s^{\prime}, a^{\prime}\right).
\end{equation} 

The parameter of the target $Q$ network, denoted by $\phi^{\prime}$, is periodically updated by the newest value of $\phi$. Although the utilization of non-linear function approximators improves the representation ability of $Q$-learning, it theoretically challenges the convergence analysis of DQN. Besides, the overestimation caused by the max operator for the greedy action selection may result in unstable learning processes. To overcome the overestimation bias, Double Deep $Q$ network (DDQN)~\cite{Hasselt2016DeepRL} is proposed that uses two deep $Q$ networks $\hat{Q}_{1}$ and $\hat{Q}_{2}$ with different parameters to decouple the action selection operation from the action evaluation, thus the learning target of the DDQN is defined as follows \begin{equation}
y_{\mathrm{DDQN}}=r(s, a)+\gamma \hat{Q}_{2}\left(s^{\prime}, \underset{a^{\prime}}{\arg \max } \hat{Q}_{1}\left(s^{\prime}, a^{\prime}\right)\right).
\end{equation} 

Apart from DDQN, there are some other DQN variants that have been developed to improve the performance of DQN. Prioritized experience replay~\cite{Schaul2016PrioritizedER} technique is adopted to train the $Q$-network to increase the probability of sampling rare and task-related samples from the replay buffer. Dueling DQN~\cite{Wang2016DuelingNA} utilizes a state value network $V$ as well as an advantage network $A(s, a)$ approximating the advantages of state-dependent action, to estimate the $Q$-network.

\subsection{Policy-based Algorithms}
The policy-based algorithms, in contrast to value-based algorithms, can directly find an optimal policy over a policy space instead of implicitly deriving it using the estimate of the action-value function. The policy can be parameterized by a function approximator, e.g., a DNN, with $\pi(\cdot \mid s) \approx \pi_{\theta}(\cdot \mid s)$. The agent learns the policy parameterized by $\theta$ through maximizing the expected long-term reward \begin{equation}
J(\theta)=\mathbb{E}_{\pi_{\theta}}\left[\sum_{t=0}^{\infty} \gamma^{t} R\left(s_{t}, a_{t}\right)\right].
\end{equation} 

This is known as the policy gradient method, and the optimal policy can be learned by performing gradient ascent on the expected return $J$. In particular, the policy gradient is given by~\cite{Sutton1999PolicyGM}\begin{equation}
\nabla J(\theta)=\mathbb{E}_{a \sim \pi_{\theta}}\left[\sum_{t=0}^{T}Q_{ \pi_{\theta}}(s_{t}, a_{t}) \nabla \log \pi_{\theta}(a_{t} \mid s_{t})\right],
\end{equation}where $Q_{\pi_{\theta}}(s, a)$ is the estimation function that can be represented by different forms in different policy gradient instances. The well known \emph{REINFORCE} algorithm~\cite{Williams2004SimpleSG} utilizes the cumulated reward during an episode defined as $\sum_{k=t}^{T} R\left(s_{k}, a_{k}\right)$ to estimate the $Q$-function. It can be inferred that \emph{REINFORCE} algorithm is the on-policy method that can only use the trajectories or samples collected by the current policy and cannot reuse the old policies' experiences, which suffers from poor sample efficiency. Moreover, the cumulated reward computed after one episode can be quite different due to the random trajectory sequences during one episode, which may lead to high variance. To solve the issue, the actor-critic is proposed to use an approximation of $Q_{\pi_{\theta}}(s, a)$ to estimate the expected return. The actor represents the policy for action selection, while the critic is used to estimate the $Q_{\pi_{\theta}}(s, a)$ to evaluate the actor's action. The use of the critic reduces the variance of the policy gradient since it can make use of different samples instead of only using one sample trajectory. Besides, the value function $V^{\pi_{\theta}}$ or the advantage function $A^{\pi_{\theta}}(s_{t},a_{t})=Q^{\pi_{\theta}}-V^{\pi_{\theta}}$ can also be used to estimate the $Q_{\pi_{\theta}}(s, a)$. However, these estimation methods may also introduce bias since the estimated result is inaccurate. To reduce the bias, the generalized advantage estimation~\cite{Schulman2016HighDimensionalCC} is proposed that uses $n$-step returns instead of one-step return to reduce the estimation bias.

In the area of future Internet, the actor-critic framework and some variants have recently been used in decentralized systems. To achieve a higher learning efficiency, multiple actors can learn policies for the same task in parallel, which is implemented in Asynchronous Advantage Actor-Critic (A3C) algorithm~\cite{Mnih2016AsynchronousMF}. In A3C, a global network is shared between an actor and a critic to output the action probabilities and an estimate of the advantage function. All the policies share the same copy of the global network and work in parallel to interact with the local environments, and collect experiences to estimate the gradients regarding the parameters of policies. Each actor can propagate its cumulated gradients to update the parameters of the global network. This enables the actor to learn from the experiences gathered by different actors in parallel environments. However, the asynchronous training may degrade the performance of the global network since some actors may use outdated policies to generate the gradients. To overcome the issue, a synchronous training mechanism, namely advantage actor-critic (A2C) can be used to aggregate and average all the actors' gradients to update the global network.

Learning rate is an important factor that can determine the difference between the old policy and the new policy, which can affect the performance of the learning algorithms dramatically. However, it can be hard to find a satisfactory learning rate since both a low or a high learning rate can lead to a poor policy. This motivates the Trust Region Policy Optimization (TRPO) method proposed in~\cite{Schulman2015TrustRP} that learns the optimal policy under the constraint of the policy difference measured by the Kullback-Leibler Divergence ($D_{KL}$). In addition to TRPO, another method Proximal Policy Optimization (PPO)~\cite{Schulman2017ProximalPO} uses a penalty scheme in which a clip function is imposed upon the loss function to penalize the new policy if it gets far from the old policy.

The policy of an agent can either output a probability distribution over possible actions or a deterministic action that the agent can execute directly, which is known as the stochastic policy and the deterministic policy, respectively. The stochastic policy $\pi$ that is widely used in the aforementioned policy-based algorithms, selects an action by sampling from a given action distribution, $a_{t} \sim \pi(s_{t})$. This can motivate the agent to explore more information about the environment. The deterministic policy, in contrast, outputs a deterministic action, $a_{t}= \mu(s_{t})$, which improves the efficiency of action selection since it does not need to sample the action over the whole action distribution. The typical deterministic policy-based RL method is Deterministic Policy Gradient (DPG)~\cite{Silver2014DeterministicPG} that learns a $Q$-function and a policy at the same time. The $Q$-function $Q_{\phi}$ is learned by minimizing the Bellman error, and the optimal deterministic policy $\mu^{*}_{\theta}$ is learned through maximizing the objective \begin{equation}
J(\theta)=\mathbb{E}_{s \sim D}\left[Q_{\phi}\left(s, \mu_{\theta}(s)\right)\right].
\end{equation}

Deep Deterministic Policy Gradient (DDPG)~\cite{Lillicrap2016ContinuousCW} as an extension of the DPG algorithm has been widely used to solve the issues in the area of future Internet. DDPG adopts the target network technique used in DQN to stabilize training, and the actor-critic framework to reduce the variance of the policy gradients. Accordingly, DDPG calculates the target values by leveraging the target actor network $\mu_{\theta^{\prime}}$ and the target critic network $Q_{\phi^{\prime}}$. These two networks are initialized and updated by the primary actor network $\mu_{\theta}$ and the primary critic network $Q_{\phi}$, respectively. Note that the target network is softly updated, e.g., $\theta \leftarrow \tau \theta+(1-\tau) \theta^{\prime}$ with $\tau \ll 1$, which allows the target network to change slowly and further improves the stability of the learning process.

\textbf{Summary: }In this section, we have reviewed the backgrounds and the classical algorithms of single-agent RL. It is worth noting that both MDPs and POMDPs only consider scenarios where a single agent interacts with the environment, and thus they are obviously not suitable for modeling multi-agent environments. In the next section, we review the backgrounds, challenges, learning schemes, and the algorithms of MARL.

\section{Multi-Agent Reinforcement Learning}\label{MARL}
In this section, we introduce the backgrounds of MARL, challenges and learning schemes in multi-agent environments, and some advanced MARL algorithms. Moreover, we introduce some commonly used reinforcement learning libraries and toolkits. 

\subsection{Markov Games}\label{MG}
The communication networks such as cellular networks or vehicular networks naturally involve multiple network entities, and the network entities such as mobile users or vehicles as agents usually need to interact not only with the environments but also with other entities. Thus, it is important to take the interactions among entities into account to develop efficient MRAL algorithms for future Internet issues. Markov Games (MGs)~\cite{Littman1994MarkovGA} or Stochastic Games (SGs)~\cite{shapley1953stochastic} as the extension of MDPs to multi-agent environments, can jointly consider all the agents in the environment as well as their interactions. In particular, the MG is defined by a 5-tuple $\left( \mathcal{S},\mathcal{N},\left\{\mathcal{A}^{i}\right\}_{i \in \mathcal{N}}, \mathcal{T},\left\{R^{i}\right\}_{i \in \mathcal{N}}, \gamma\right)$, where $\mathcal{S}$ is the state space observed by all the agents, $\mathcal{N}=\{1, \cdots, N\}$ represents a set of agent indices, $\mathcal{A}=A_{1} \times \ldots \times A_{\mathcal{N}}$ is the action space of all the agents. The main differences between the MG and the MDP are the transition function $\mathcal{T}: \mathcal{S} \times \mathcal{A} \times \mathcal{S} \rightarrow[0,1]$, and the reward function $R^{i}: \mathcal{S} \times \mathcal{A} \times \mathcal{S} \rightarrow \mathbb{R}$, which demonstrate the fact that the multi-agent environment is a entirety determined by all the agents. \textcolor{black}{In the area of the future Internet, the state of each agent can be the channel information, the power allocation status, and the transmission rate. The action of each agent can be the spectrum block, the power level, or the next-hop router. The reward is usually related to the learning objectives such as maximizing sum rate or energy efficiency and is defined as the currently achieved sum rate or energy efficiency.}

At time step \emph{t}, each agent chooses an action $a_{t}^{i} \in \mathcal{A}_{i}$ to execute depending on the system state $s_{t} \in \mathcal{S}$. The state $s_{t} $ then transits to the next state $s_{t+1} $, and each agent receives a reward $R^{i}\left(s_{t}, a_{t}, s_{t+1}\right)$. Similar to the MDP, the objective of each agent is to find the optimal policy $\pi_{i}^{*}: S \rightarrow A_{i}$ so as to maximize its own long-term reward. In multi-agent environments, the value-function ${V^{i}:\mathcal{S} \rightarrow \mathbb{R}}$ of agent $i$ is determined by the joint policy $\pi$ of all the agents, \begin{equation}\label{jointpolicy}
V_{\pi^{i}, \pi^{-i}}^{i}(s):=\mathbb{E}\left[\sum_{t=0}^{\infty} \gamma^{t} R^{i}\left(s_{t}, a_{t}, s_{t+1}\right)|_{a_{t}^{i} \sim \pi^{i}\left(\cdot \mid s_{t}\right), s_{0}=s}\right],
\end{equation} where $-i$ denotes the indices of all the agents except agent $i$. From Equation \ref{jointpolicy}, we can see that the performance of each agent is determined not only by its own policy but also by other agents' policies. Thus, the common solution of the MG is to find an optimal policy set $\pi^{*}=\left(\pi^{1, *}, \cdots, \pi^{N, *}\right)$, Nash equilibrium (NE)~\cite{Baar1982DynamicNG}, such that \begin{equation}
V_{\pi^{i, *}, \pi^{-i, *}}^{i}(s) \geq V_{\pi^{i}, \pi^{-i, *}}^{i}(s), \text { for any } s \in \mathcal{S}\text { and } \pi^{i}.
\end{equation}

The NE demonstrates that the optimal policy $\pi^{i, *}$ is the best response of $\pi^{-i, *}$, and it always exists in finite-space infinite-horizon discounted MGs~\cite{Filar1996CompetitiveMD}. Generally, the NE may not be unique. Most MARL algorithms aim to find and converge to such an NE.

The cooperation or competition relationships among agents render the multi-agent environment more complicated. \textcolor{black}{Both of them are very common in the future Internet. For example, cellular users may need to cooperate with other users to maximize the sum rate in the cellular system. However, in task offloading scenarios, the users may compete for limited edge computing resources to offload their tasks to reduce their task execution delays.} The MGs can typically deal with multiple MARL settings: fully cooperative setting, fully competitive setting, and mixed setting. In the fully cooperative setting, all the agents have the common reward function where $R^{1}=R^{2}=\cdots=R^{N}=R$. This setting is also referred to as a Multi-agent MDP (MMDP)~\cite{lauer2000algorithm}. Besides, all the agents may share one policy in such a cooperative setting, which enables the application of single-agent RL algorithms, e.g., $Q$-learning or SARSA. Hence, the global optimal solution learned by the common policy for cooperation becomes an NE of the MG. However, the common reward mechanism may make some agents lazy, and thus a different reward model used for the MARL cooperative setting is the team-average reward~\cite{kar2012qd}. The team-average reward allows agents to use heterogeneous reward functions, and the goal for cooperation is to maximize the long-term average reward $\bar{R}\left(s, a, s^{\prime}\right):=N^{-1} \cdot \sum_{i \in \mathcal{N}} R^{i}\left(s, a, s^{\prime}\right)$ of all the agents. When used for the fully competitive setting, the MGs are also referred to as zero-sum MGs, namely, $\sum_{i \in \mathcal{N}} R^{i}\left(s, a, s^{\prime}\right)=0$. In particular, in a two-player MGs, the reward of one agent is the loss of the competitor~\cite{littman1994markov}. Moreover, zero-sum games can model the uncertainty which may impede the agent's learning process, as a fictitious opponent~\cite{lagoudakis2002learning}, which leads to more robust learning. When used for the mixed setting, the MG is considered to be a general-sum game, that consists of the cooperative and the competitive agents~\cite{zhang2018finite}. In this case, each agent may have a common reward or the conflict reward with other agents. Some MRAL algorithms, such as the well-known MADDPG~\cite{lowe2017multi}, have been developed for mixed settings to search the NE of the MG.

\subsection{Dec-POMDP}\label{Dec-POMDP}
From the definition of MGs, we can see that the global state determines the joint policy of all the agents, which means that each agent needs to access the global state information to maximize its long-term reward. \textcolor{black}{However, in most distributed networking systems, e.g., multi-cell networks, the agents may not receive the global state information timely.} The uncertainty about other agents and environments can impede the agent from learning of a global optimum. In this case, the partially observed MG under cooperative settings is typically modeled as the Dec-POMDP~\cite{Bernstein2000TheCO},~\cite{Oliehoek2016ACI}. The Dec-POMDP almost shares all the elements of the MG such as the reward function and the transition function, except that each agent $i$ only has a partial observation $o^{i}$ of the system state $s$ to make local action decisions. The objective of each agent is to learn a local optimal policy $\pi_{i}$ that can maximize its cooperative long-term reward. The partial observation issue of the Dec-POMDP motivates the development of decentralized MARL methods, which have been widely used in the future internet. We will discuss these methods in Section \ref{MARLalgorithms}.

\subsection{Challenges and Learning Schemes in Multi-Agent Environments}\label{Challenges}
In this section, we discuss challenges in both theory and implementation when using MARL algorithms in multi-agent environments.

\textcolor{black}{\emph{\textbf{1) Non-stationarity issue: }}In multi-agent environments, the agents can learn their policies concurrently. As a result, the action of an agent may affect the rewards of other agents, and the state transition is decided by the joint actions of all the agents. This violates the stationarity assumption, also known as the stationary Markov property of the environment, that is, the current reward and the state depends only on the previous state and the action taken by the agent. Therefore, the convergence of the single-agent RL algorithm cannot be guaranteed. This is called the non-stationarity issue~\cite{hernandez2017survey}. To ensure the convergence of the learning algorithm, the learning agent should account for the actions of other agents when making action decisions.} \textcolor{black}{In the area of the future Internet, the network entities such as cellular users may suffer from the non-stationarity issue since other users' spectrum access policies or power control policies can significantly influence their own policies.}

\textcolor{black}{\emph{\textbf{2) Scalability issue: }}To overcome the non-stationarity issue, the joint actions of other agents need to be taken into account. As a consequence, the joint action space grows exponentially as the number of agents increases, which may raise scalability issues in multi-agent environments. To solve the issue, the approximators such as DNNs have been widely used in MARL algorithms to deal with such combinatorial information. However, this may challenge the convergence analysis of MARL because of the limited understanding of deep learning theory.} \textcolor{black}{In the networking area, the scalability issue is critical when the centralized methods that consider all the agents' actions are employed to solve the issues in communication networks such as IoTs with a large number of devices or cellular systems with dense users.}

\textcolor{black}{\emph{\textbf{3) Partial observability issue: }}The learning agents are typically unable to have a full observation on the environments such as large-scale communication networks in reality. They usually make autonomous action decisions based on their partial observations, e.g., the local CSI. Since the agents cannot access the actions of other agents, they may not solve the non-stationarity issue. Therefore, the convergence of the policy of the learning agent cannot be guaranteed.} \textcolor{black}{In the future Internet, the network entities typically only have partial observability since they may not observe the global system information, e.g., the global CSI.}

To solve the above issues, MARL enables the agents to use various state/observation structures, which result in different learning schemes. We discuss the commonly used learning schemes in communication networks as follows.

\begin{itemize}
\item \textcolor{black}{\emph{\textbf{Fully centralized learning: }}The fully centralized learning can leverage the instantaneous global state information, e.g., the global CSI, to learn a central policy for all the agents. In particular, such information can be collected by a central controller, e.g., the BS, through exchanging information with the agents.} \textcolor{black}{Though the fully centralized learning can be used to overcome the non-stationarity issue, it may occur high communication overhead for information collection, and the communication delay may render the collected state information to be outdated due to the time-varying network environments. As a result, it is not suitable to be employed in real-world networks.}

\item \textcolor{black}{\emph{\textbf{Centralized training with decentralized execution (CTDE): }}One compromise to solve the problem of fully centralized learning is to leverage CTDE~\cite{macua2014distributed}. CTED trains a unique policy for each agent through learning the experiences of all the agents, and it does not require the instantaneous global state.} After the offline training process, it distributes the learned policies to the corresponding agents for decentralized execution. The learned policy only needs its own observation to take action decisions. Thus, CTDE enables the agent to learn a decentralized policy through global optimization. \textcolor{black}{In the distributed network, CTDE enables the network entity to learn a decentralized policy by centralized training.}

\item \textcolor{black}{\emph{\textbf{Decentralized learning with networked agents: }}The centralized methods that directly leverage the global information may suffer from poor scalability. One solution is to use decentralized learning with networked agents~\cite{Zhang2018FullyDM} that enables the agents to exchange information with their neighbors to share their states, actions, and even policies for collaboration.} In this case, the agent only needs to use its local information and its neighbors' information to make action decisions, which can overcome the scalability issue and significantly reduce the computational complexity. \textcolor{black}{This learning scheme can be efficiently used in the area of the future Internet, as the network entities can naturally communicate with each other and share their local information for collaboration.}

\item \textcolor{black}{\emph{\textbf{Independent learning (IL): }}Different from decentralized learning with networked agents, the IL assumes that the agent cannot communicate with each other. Thus, the agent can only use its local observation to make action decisions without any collection of other agents' information.} \textcolor{black}{In some wireless networks with unreliable communications or high communication costs, e.g., UAV networks or IoTs, the IL is preferred. Each node or UAV only needs its own observation to make action decisions that can reduce signaling overhead dramatically. However, it may suffer from the non-stationarity issue, and the convergence of the learning policy cannot be guaranteed.}

\end{itemize}

\textcolor{black}{\subsection{CTDE-based MARL Algorithms}\label{MARLalgorithms}}

\textcolor{black}{In this section, we present and discuss some CTDE-based MARL algorithms that are commonly employed for solving issues in the area of the future Internet.}

\begin{figure}[h]
    \centering
    \subfigure[]{
    \includegraphics[width=0.22\textwidth]{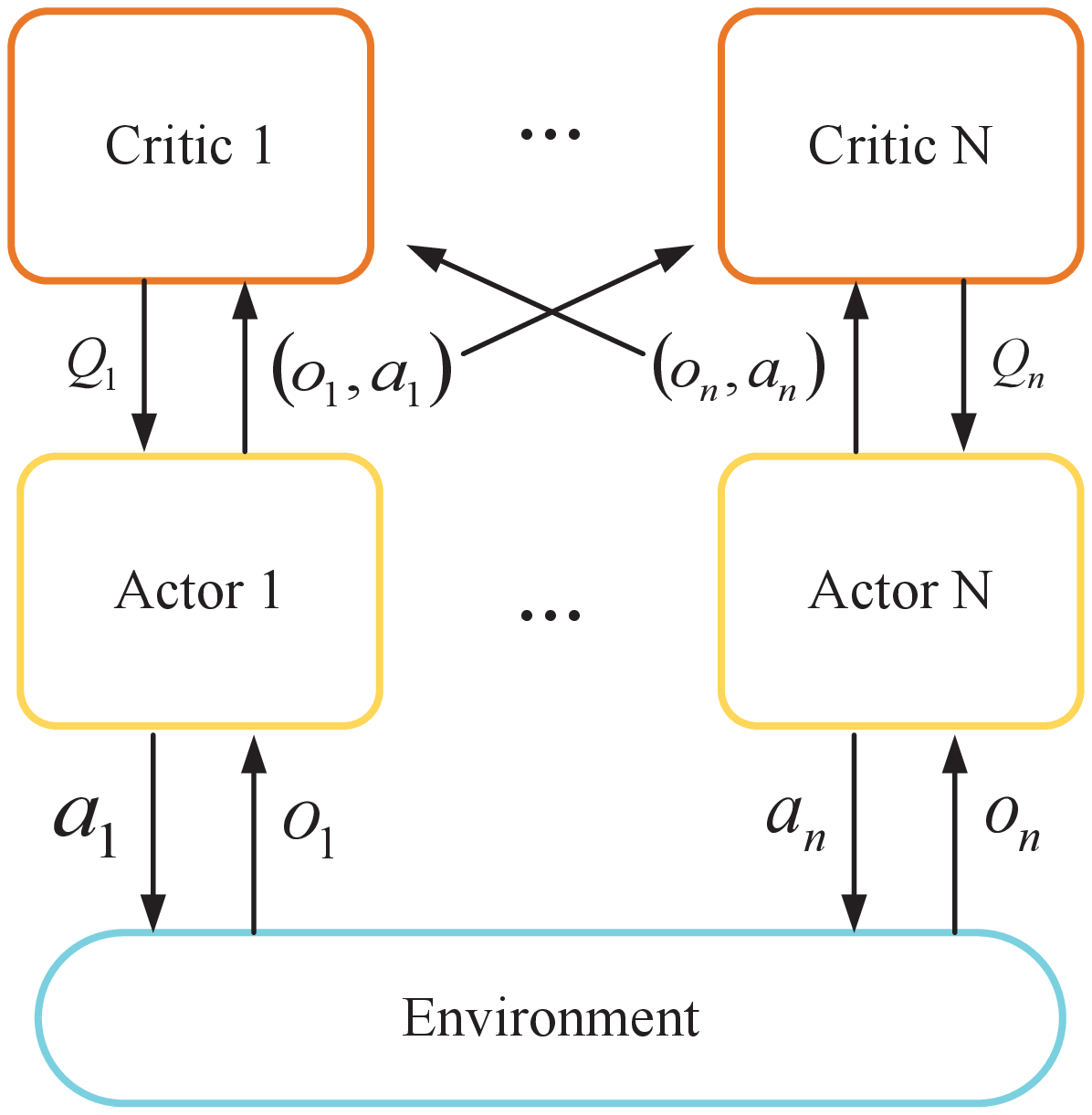}
}
    \subfigure[]{
    \includegraphics[width=0.22\textwidth]{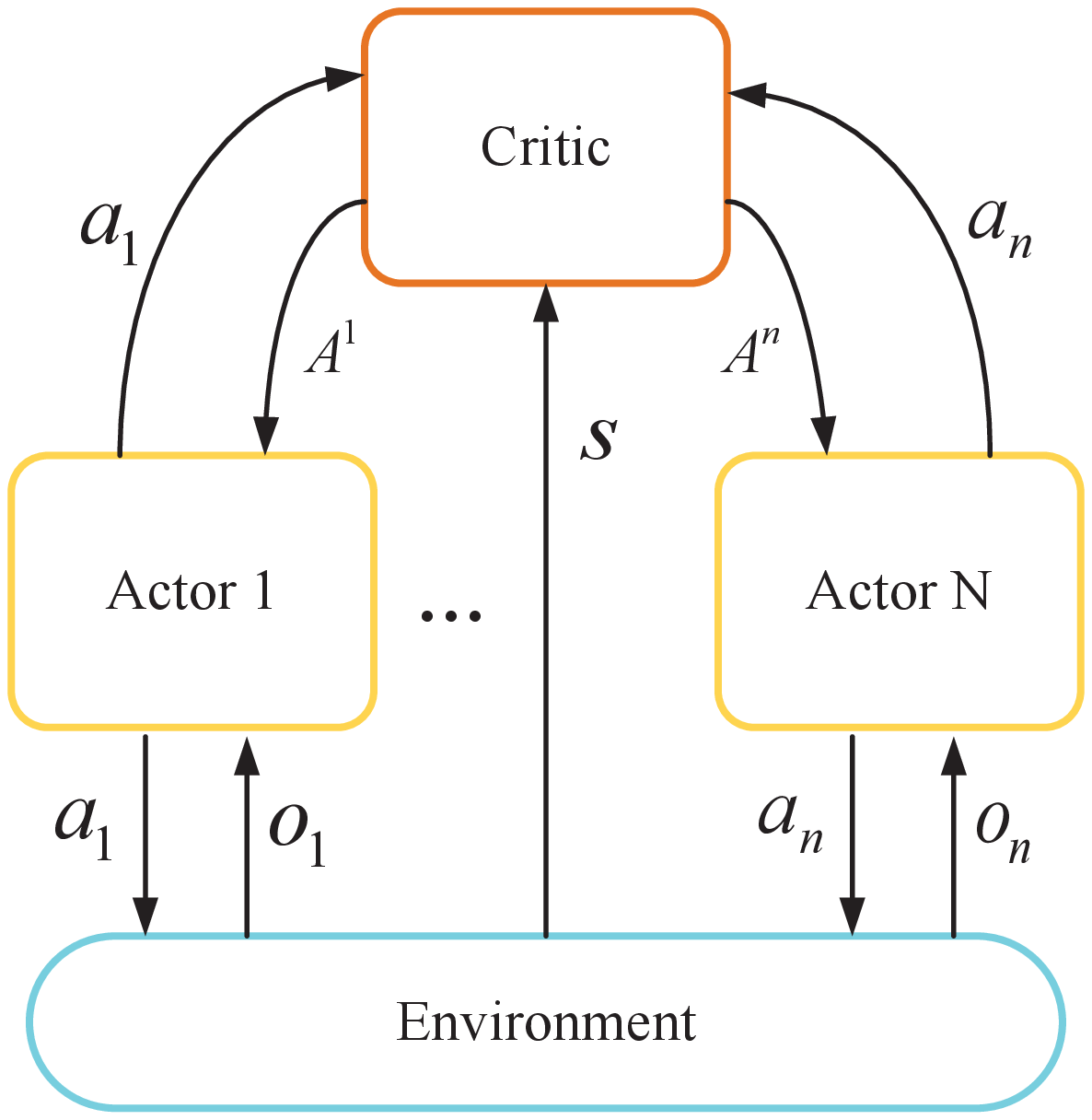}
}

\caption{Multi-agent learning model of (a) MADDPG and (b) COMA.}
\label{model}
\end{figure}

\textcolor{black}{\emph{1) Multi-Agent Deep Deterministic Policy Gradient (MADDPG): }MADDPG~\cite{lowe2017multi} is one of the most popular MARL algorithms using the CTDE framework, where each agent has a decentralized actor and a centralized critic. It is an extension of DDPG to multi-agent environments and can be applied to cooperation, competition, and mixed settings. MADDPG learns the optimal policy for each agent using CTDE.} Thus, as shown in Fig. \ref{model}(a), the critic of each agent is able to access the actions of all other agents enabling it to evaluate the impact of the joint policy on the expected reward of each agent. Moreover, the critic needs to access the policies of other agents to compute the corresponding target actions so as to minimize the bellman error of the action-value function. In this case, MADDPG allows the agents to learn approximate policies of other agents, and thus the assumption of knowing other agents' policies can be relaxed. To learn a policy that is robust to the changes of other agents' policies, each agent learns a policy ensemble instead of one single policy and selects one sub-policy to make action decisions. Moreover, ensemble training can also motivate the agent to achieve a better generalization when deployed with agents with unknown policies. \textcolor{black}{MADDPG can be used to solve various issues in the future Internet. For example, in the Mobile Edge Computing (MEC) network, MADDPG can be adopted to learn an optimal task offloading policy for each network user.}

\textcolor{black}{\emph{2) Counterfactual Multi-Agent Policy Gradients (COMA): }The second algorithm we discuss is COMA~\cite{foerster2018counterfactual}, as shown in Fig. \ref{model}(b). Similar to MADDPG, with COMA, each agent has a local actor to make action decisions. However, unlike the MADDPG, the COMA estimates the action-value function via a single centralized critic. The MADDPG uses the states and actions of all the agents to estimate the expected reward, and the agent may not figure out its action's contribution to the total reward, which is known as the credit assignment issue in multi-agent environments.} The COMA addresses the credit assignment issue using the counterfactual baseline method. In particular, COMA computes an advantage function for each agent $i$ by using the counterfactual baseline as follows,\begin{equation}
A^{i}(s, \bm{a})=Q(s, \bm{a})-\sum_{a^{\prime i}} \pi^{i}\left(a^{\prime i} \mid \tau^{i} \right) Q\left(s,\left(\bm{a}^{-i}, a^{\prime i}\right)\right),
\end{equation} where $Q(s, \bm{a})$ is the $Q$-value of the current joint action $\bm{a}$, the second term $\sum_{a^{\prime i}} \pi^{i}\left(a^{\prime i} \mid \tau^{i} \right) Q\left(s,\left(\bm{a}^{-i}, a^{\prime i}\right)\right)$ is the counterfactual baseline. Furthermore, COMA uses a single forward pass in both the critic network and the actor network to efficiently compute the counterfactual advantage function for each agent. It is noted that COMA can only be used in cooperative scenarios due to the use of a single centralized critic. Moreover, with COMA, the critic calculates the advantage function under the assumption of knowing the global state, which is impossible in large scale systems.

\textcolor{black}{\emph{3) Mean-Field $Q$-Learning (MF-Q): }Both MADDPG and COMA may suffer from poor scalability since the joint action space of all the agents grows exponentially with the number of agents. To address the issue, MF-Q~\cite{yang2018mean} can be used. The MF-Q adopts a mean-field approximation of other agents' actions, and the $Q$-function of agent $i$ can then be defined by\begin{equation}
Q_{i}(s, \bm{a})=\frac{1}{N_{i}} \sum_{k \in \mathcal{N}(i)} Q_{i}\left(s, a_{i}, a_{k}\right) \approx Q_{i}\left(s, a_{i}, \hat{a}_{i}\right),
\end{equation}where $\mathcal{N}(i)$ is the set of neighboring agents of agent $i$, $N_{i}=|\mathcal{N}(i)|$. The average value of $Q_{i}\left(s, a_{i}, a_{k}\right)$ can be approximated by $Q_{i}\left(s, a_{i}, \hat{a}_{i}\right)$. In this setting, the effect of actions of neighboring agents is converted to the mean-field value $\hat{a}_{i}$ and it can be treated as an action of a mean-field agent. As a consequence, the interactions between the agent and its neighbors become the interactions between the agent and the mean-field agent, which significantly reduce the action space caused by the increasing number of agents. Therefore, MF-Q can be used to solve the large scale multi-agent issues due to the mean-field benefit.} \textcolor{black}{For example, in wireless sensor networks with a large number of sensor nodes, MF-Q can be simply employed for the nodes to learn a decentralized policy.}

\textcolor{black}{\emph{4) Multi-Actor Attention-Critic (MAAC): }Another method to overcome the scalability issue is to use the attention mechanism~\cite{vaswani2017attention}, namely, MAAC~\cite{iqbal2019actor}. MAAC is an extension of SAC~\cite{Haarnoja2018SoftAO} to multi-agent environments, where each agent has a critic that shares a central attention mechanism with the critics of other agents. In particular, the $Q$-function, i.e., the critic, of agent $i$ is represented as\begin{equation}\label{maac1}
Q_{i}(\bm{o}, \bm{a})=f_{i}\left(g_{i}\left(o_{i}, a_{i}\right), x_{i}\right),
\end{equation}where $f_{i}$ is a local double layer dense net, $g_{i}$ is a local single-layer perception that is responsible for encoding the observation $o_{i}$ and the action $a_{i}$ of agent $i$, $x_{i}$ represents the contribution of other agents to the current agent $i$. To solve the non-stationarity issue, the calculation of $x_{i}$ is centralized at the shared central attention mechanism by using the embeddings of other agents,\begin{equation}\label{maac2}
x_{i}=\sum_{j \neq i} \alpha_{j} v_{j}=\sum_{j \neq i} \alpha_{j} h\left(V g_{j}\left(o_{j}, a_{j}\right)\right),
\end{equation}where $\alpha_{j}$ is the attention weight of agent $j$ to represent its contribution to the agent $i$, and $v_{j}$ is the linear transformation of agent $j$'s embedding.} By using Equation \ref{maac2}, the embeddings of other agents can be converted to the contribution $x_{i}$ via the attention mechanism, and each local agent only needs to use its own embedding and the contribution $x_{i}$ to make action decisions. Therefore, the scalability issue can be solved. Moreover, the attention mechanism can selectively scale different agents' embeddings, and thus the learning efficiency can be improved.

\textcolor{black}{\subsection{Value-Decomposition MARL Algorithms}\label{Value_Decomposition}}
\textcolor{black}{In this section, we review some value-decomposition MARL algorithms which can efficiently solve the credit assignment issue in multi-agent environments, and significantly improve the performance of CTDE.}

\textcolor{black}{\emph{1) Value-Decomposition Networks (VDN): }The shared team reward in cooperative multi-agent environments may be the spurious rewards for some lazy agents since the team reward sometimes is the contribution of other teammates. This means that the lazy agent can also receive a decent team reward when some other agents learned a better policy and completed the task. The credit assignment issue is solved in COMA based on the counterfactual baseline. Another value-based method, namely, VDN~\cite{sunehag2017value}, is to decompose the total $Q$-function to individual $Q$-functions under the assumption \begin{equation}
Q(\boldsymbol{s}, \boldsymbol{a})=\sum_{i=1}^{d} Q_{i}(\boldsymbol{s}, \boldsymbol{a}),
\end{equation} where $d$ is the number of agents. Given that the global state $\boldsymbol{s}$ may not be accessible by the individual agent, VDN adopts the historical observation sequences $h^{i}$ to approximate the global state as follows:\begin{equation}
Q\left(\left(h^{1}, h^{2}, \cdots, h^{d}\right),\left(a^{1}, a^{2}, \cdots, a^{d}\right)\right) \approx \sum_{i=1}^{d} \tilde{Q}_{i}\left(h^{i}, a^{i}\right).
\end{equation}}

Each agent learns the optimal value decomposition from the global value function depending only on its local observation. In particular, during centralized training, VDN only needs to calculate the TD-error of the global value function and then back-propagates the TD-error to each sub-value function, which greatly reduces the amount of computation. Similar to COMA, VDN also adopts the parameter sharing mechanism to reduce the training parameters.

\textcolor{black}{\emph{2) QMIX: }The decomposition of VDN depends on the liner splits of global value function that can satisfy the monotonicity, i.e.,\begin{equation}
\frac{\partial Q_{total}}{\partial Q_{i}}=1, \forall i=1, \cdots, n.
\end{equation}}

\textcolor{black}{This means that the result of the decentralized $Q$-function is consistent with the global $Q$-function in terms of the optimal performance. However, the monotonicity does not need to strictly follow such tight rule defined in VDN that makes the individual $Q$-functions not fit the optimal global $Q$-function very well. QMIX~\cite{rashid2018qmix}, on the other hand, sets a loose monotonicity rule as\begin{equation}
\frac{\partial Q_{total}}{\partial Q_{i}} \geq 0, \forall i=1,\cdots, n,
\end{equation}which demonstrates that the global $Q$-function does not need to be the sum of sub-$Q$-functions. As a consequence, QMIX estimates the global $Q$-function by inputting the individual $Q$-functions of each agent to a designed mixing network under the defined monotonicity constraint in order to make the learning robust and stable. Besides, the global state is utilized as the input of the mixing network during training, which is not used in VDN. The mixing network improves the fit ability of QMIX compared with the simple summation utilized in VDN.}

\textcolor{black}{\emph{3) Weighted QMIX (WQMIX): }The monotonicity constraint in QMIX limits the joint action representation of $Q$-function that causes it to fail in tasks where agents may interact with each other and the agent needs to consider other agents' actions to make action decisions. To overcome the issue, WQMIX~\cite{rashid2020weighted}, an extension of QMIX, uses the weight function $w: S \times \boldsymbol{U} \rightarrow(0,1]$ to assign different weights to the joint actions $\boldsymbol{u}$ of all the agents. Besides, QMIX adopts a $\hat{Q}^{*}$ function to evaluate the joint action to maximize the Monotonic function $Q_{total}$ instead of searching directly the $Q^{*}$ over the joint action all the agents, which can significantly improve the learning efficiency.}

We summarize the referred MARL algorithms in Table \ref{MARLalg} in terms of setting, learning type, algorithm type, shared information, and whether to consider the credit assignment issue. As observed in Table \ref{MARLalg}, only MADDPG can be used in a mixed environment that consists of both cooperative and competitive agents since each agent has a local critic and a local actor that can deal with various reward structures. The parameter sharing mechanism-enabled algorithms such as COMA and VDN can only be used in cooperative settings. Moreover, most of the algorithms consider the credit assignment issue except for MADDPG and MF-Q.

\textcolor{black}{\subsection{Reinforcement Learning Libraries and Toolkits}\label{Reinforcement Learning Libraries and Toolkits}
Building reinforcement learning algorithms especially MARL from scratch can be difficult and complicated for researchers. In this section, we aim to introduce some Reinforcement Learning libraries and toolkits that can be easily used to simulate different networking scenarios and evaluate their algorithms.}

\textcolor{black}{\emph{1) Tensorflow: }Tensorflow~\cite{abadi2016tensorflow} is a well known machine learning library developed by Google. It allows the researchers to construct their computation graphs on different devices, e.g., CPUs and GPUs. It can be employed not only for single architecture but also for distributed architecture that is very common in networking scenarios such as UAV networks, vehicular networks, and D2D networks. Furthermore, Tensorflow provides a visualization tool, TensorBoard. The researchers can see their constructed model structures and data flows from the web page generated by TensorBoard. Many reinforcement learning researchers also use it to understand the real-time reward of the agent and learning loss from the TensorBoard page. Tensorflow not only supports Python not also some commonly used programming languages, e.g., C and Java. With the development of deep reinforcement learning, Tensorflow has become one of the most popular libraries for researchers to implement their own reinforcement learning models.}

\textcolor{black}{\emph{2) Pytorch: }Another popular reinforcement learning library is pytorch~\cite{collobert2011torch7}. It provides the Tensor computation with strong GPU acceleration. Therefore, it can replace Numpy~\cite{oliphant2006guide}, a scientific computation package, and significantly accelerate the computation speed with the power of GPUs. Moreover, different from the way that Tensorflow uses to build the neural network structure, Pytorch provides a unique way of constructing neural networks with the reverse-mode auto-differentiation technique~\cite{baker2021efficient}. It allows the researchers to transform the way that the neural network behaves arbitrarily. Pytorch also supplies a straightforward style for the researchers to build their learning models. The above reasons make Pytorch more and more popular in the area of academia. }

\textcolor{black}{\emph{3) Gym: }Gym~\cite{brockman2016openai} is an open-source reinforcement learning toolkit developed by OpenAI. It can be simply employed to develop and compare different reinforcement learning algorithms. Gym provides standard APIs to connect the learning algorithms and simulation environments. Thus, it allows the researchers to easily evaluate their algorithms with different simulation environments. Moreover, Gym allows researchers to develop their own simulation environments under the standard of Gym. Thus, it can be used to simulate various environments according to the research demands.}

\begin{table*}[]\scriptsize\centering
\renewcommand\arraystretch{1.5}
\caption{A summary of MARL algorithms based on information sharing}
\label{MARLalg}
\begin{tabular}{|c|c|c|c|c|c|}
\hline
\rowcolor[HTML]{9B9B9B}
\textbf{Algorithms} & \textbf{Settings}     & \textbf{Learning types}   & \textbf{Algorithm types} & \textbf{Shared information}  & \textbf{Credit assignment}           \\ \hline
MADDPG~\cite{lowe2017multi}     & Mixed       & CTDE                                 & Actor-Critic   & Joint observations and actions & \XSolidBrush \\ \hline
COMA~\cite{foerster2018counterfactual}       & Cooperative & CTDE                                 & Actor-Critic   & Joint observations and actions &\CheckmarkBold \\ \hline
MF-Q~\cite{yang2018mean}       & Cooperative & CTDE                                 & Value-based    & Global state and joint actions &\XSolidBrush     \\ \hline
MAAC~\cite{iqbal2019actor}       & Cooperative & CTDE                                 & Actor-Critic   & Joint observations and actions &\CheckmarkBold  \\ \hline
VDN~\cite{sunehag2017value}        & Cooperative & Value Decomposition & Value-based    & Joint $Q$-functions    &  \CheckmarkBold        \\ \hline
QMIX~\cite{rashid2018qmix}       & Cooperative & Value Decomposition & Value-based    & Joint $Q$-functions    & \CheckmarkBold         \\ \hline
WQMIX~\cite{rashid2020weighted}      & Cooperative & Value Decomposition & Value-based    & Joint $Q$-functions   &  \CheckmarkBold         \\ \hline
\end{tabular}
\end{table*}

\textbf{Summary: }In this section, we have reviewed the backgrounds, challenges, learning schemes, and algorithms in MARL as well as some commonly used reinforcement learning libraries and toolkits. Note that the challenges of MARL such as non-stationarity, scalability, and partial observability result in various learning schemes and multiple MARL algorithms. In the rest of the paper, we discuss the applications of MARL for solving challenging issues in the future internet.

\section{Network Access and Power Control}\label{Network Access and Power Control}
In future wireless networks, the high user density may result in poor QoS due to the high competition on the available spectrum resources and serious interference. Although the emergence of new technologies such as 5G and Narrow Band Internet of Things (NBIoT) expands the available spectrum resource and improves the spectrum efficiency, they may not be able to meet the growing QoS demands of the network users~\cite{popli2018survey}. Thus, network resource allocation becomes an emerging issue that enables the network entity, e.g., a mobile user or a base station (BS), to autonomously select network resources such as spectrum or power to achieve their goals, e.g., maximizing their throughput. However, the dynamics and uncertainties of wireless networks challenge the conventional methods such as geometric programming~\cite{gjendemsjo2008binary} and game theory~\cite{song2014game} to realize efficient and practical solutions. MARL enables network entities to autonomously learn the knowledge about the environment and other entities to make optimal decisions on network resources. In this section, we review the applications of MARL for spectrum access and power control in wireless networks.

\begin{itemize}
\item \emph{\textbf{Dynamic spectrum access: }}Dynamic spectrum access enables the network entity to select available spectrum resources to maximize their throughput. Some conventional optimization methods such as RICE~\cite{wu2018high} and ETRP~\cite{kuang2018energy} need to use the collected instantaneous global CSI to allocate spectrum resources to the network entities, which may result in a large amount of signaling overhead for exchanging CSI and high computational complexity for global optimization. On the contrary, MARL can be efficiently used to provide decentralized solutions that enable the network entity to only use its local observations, e.g., the local CSI, to make action decisions.

\item \emph{\textbf{Transmit power control: }}To improve the transmission rate, the network entity, e.g., a mobile user or a base station (BS), may need to increase its transmit power. However, this can cause serious interference to other network entities. Therefore, each network entity needs to control its transmit power to maximize the overall network throughput or energy efficiency. This problem is typically NP-hard~\cite{luo2008dynamic} and non-convex and may need complete network information. Thus, MARL can be used as an effective tool to solve the power control problem.

\end{itemize}

\subsection{Dynamic Spectrum Access}
In this section, we review the applications of MARL for dynamic spectrum access in emerging networks such as D2D-enabled cellular systems, UAV networks, cognitive radio networks, and vehicular networks. Furthermore, we review the applications of MARL for joint spectrum access and user association.

\emph{\textbf{1) Dynamic spectrum access: }}

\emph{Spectrum access for D2D communications: }In a D2D-enabled cellular network, the D2D users may work in the underlay mode and share the same spectrum resources with cellular users to improve the spectrum efficiency. Thus, they may incur interferences to the cellular users. The problem of each D2D user is to select a spectrum resource so as to maximize its own throughput under the constraints of the Interference-Plus-Noise-Ratio (SINR) of the cellular users. Directly solving such a combinatorial optimization problem with nonlinear constraints can be hard, thus the authors in~\cite{Zia2019ADM} propose to use MARL algorithms to solve the problem. The agent is the D2D user. Each D2D user learns its spectrum access policy independently by adopting the tabular $Q$-learning algorithm under a non-cooperative scenario in which the D2D user cannot communicate with each other for cooperation. The state of each D2D user is represented by the selected resource block in the previous time slot. The D2D user receives a positive reward equivalent to its throughput achieved by the joint actions of all the agents if the cellular users’ QoS requirements are satisfied, otherwise, a penalty will be imposed. The simulation results show that the proposed non-cooperative tabular $Q$-learning performs well in dense multi-tier HetNets compared with other non-RL methods such as distance-based resource criterion~\cite{Wang2012DistanceconstrainedRC} algorithm and Joint-RALA~\cite{Rostami2015AJR} algorithm.

Different from the non-cooperative scenario considered in~\cite{Zia2019ADM}, the authors in~\cite{Li2020MultiAgentDR} consider a cooperative scenario, where the D2D pair as an agent can collaborate with each other to maximize the sum rate of D2D links subject to the outage probability of cellular links. Given that the centralized methods require the global information to make action decisions that lead to a large amount of signaling overhead, while the decentralized methods need to share information with other neighboring agents that may not achieve a global optimization, consequently, \textcolor{black}{the authors in~\cite{Li2020MultiAgentDR} adopt a multi-agent actor-critic (MAAC) algorithm for spectrum allocation by exploiting the framework of CTDE. How the CTDE framework enables the collaboration between agents is explained in Section \ref{Challenges}.} Both the actor and the critic are approximated by DNNs. The training process is centralized at the BS, where a centralized critic learns the shared historical states, actions, and policies of all the agents to train an actor for each agent. Thus, the agent needs to upload their historical information to the BS periodically. During the execution period, the agent downloads the weights of its actor from the BS to choose the spectrum resource depending only on its local state. The local state of each D2D pair is represented by (i) the instant channel information of the current D2D link, (ii) the channel information of the corresponding D2D transmitter's cellular link, (iii) the previous interference caused to the D2D link, and (iv) the spectrum resource selection of the D2D link in the previous time slot. The reward function is similar to~\cite{Zia2019ADM}.

MAAC requires all the agents' information to train their policies, which may lead to high computational complexity especially in large-scale networks. To address this issue, neighbor-agent actor-critic (NAAC) is further proposed in~\cite{Li2020MultiAgentDR} which only uses the states and actions of neighboring agents as the input of the critic network to approximate the states and actions of all the agents, thereby reducing the computational complexity at the BS. This is reasonable due to the fact that the main interferences to the current agent come from the neighboring agents. In particular, the neighbors of an agent are defined as a set of a fixed number of agents closest to it, and thus the input dimension of the critic is fixed. This means that when the user densities change, NAAC does not need to change the structure of the critic network, which allows the NAAC to scale well to networks with different user densities. The simulation results show that both MAAC and NAAC outperform the independent $Q$-learning used in~\cite{Zia2019ADM} in terms of sum rate of D2D links and outage probability of cellular links, which demonstrates the significance of cooperation. Moreover, though NAAC uses less information to make spectrum decisions compared with MAAC, it can achieve similar performance with MAAC.

The idea of leveraging the neighbors' information for centralized training is also considered in~\cite{Zhang2020DeepMR}. However, different from~\cite{Li2020MultiAgentDR} that uses the BSs to collect information, the authors in~\cite{Zhang2020DeepMR} consider that each D2D user is able to collect information from neighbors, including interfering neighbors' states and interfered neighbors' states. This information can then be fed to a DDQN-based agent model to learn an optimal spectrum selection policy. Note that the number of neighbors is not fixed like~\cite{Li2020MultiAgentDR}, and thus it may not scale well when the user density changes. 

\emph{Spectrum access for UAVs: }The centralized training used in the above works requires reliable communication links for information exchange. However, in some scenarios with unreliable and dynamic links, e.g., UAV networks, a fully decentralized learning scheme that does not need to collect global information for offline training may be more suitable. 

\begin{figure}[h]
  \centering
  \includegraphics[scale=0.8]{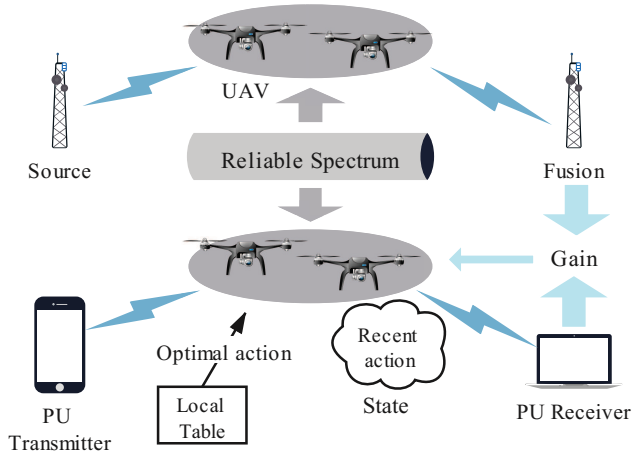}
  \caption{UAV relay model in~\cite{Shamsoshoara2019DistributedCS} where two UAVs relay packets between the Source and the Fusion Center and the other two UAVs relay packets for the PU transmitter. Each UAV takes action depending on its recent action. The gain of the fusion and the PU receiver can be used to update the local table of each UAV. The optimal action in the local table is equal to that in the global table under the deterministic multi-agent MDP setting.}
  \label{Shamsoshoara}
\end{figure}

The authors in~\cite{Shamsoshoara2019DistributedCS} adopt a fully decentralized MARL framework for spectrum allocation in UAV networks. The system model involves a remote sensing mission as shown in Fig. \ref{Shamsoshoara}, where the UAVs are categorized into two groups according to their tasks: relaying and sensing UAVs. The relaying UAVs provide relay service for a primary user (PU) to obtain the licensed spectrum resources of the PU for the sensing UAVs to transmit sensing packets. At each time slot, the UAVs independently decide to relay for the PU or transmit their own packets to a fusion center so as to maximize the total utility of the system, i.e., the gain achieved from the action selection. The authors model the spectrum decision-making problem of the UAV as a deterministic MMDP. The agent is the UAV. The observation of each UAV is the action executed in the previous time slot. The common reward for all the UAVs is the gain achieved from the action selection that is a function of the difference between the rate in the current time slot and the average of previous rates for the fusion center and the PU, and fairness of task partitioning.

Similar to~\cite{Zia2019ADM}, the tabular $Q$-learning is then employed by each UAV to solve the deterministic MMDP. To obtain the global optimal solutions, the authors utilize the projection characteristic of deterministic MMDP proposed in~\cite{lauer2000algorithm}. In particular, for the same sequence of states and actions, if each UAV selects the optimal actions from its local table, the result is equivalent to choosing the global optimal action from the global table of all the UAVs. This is reasonable since only one task partition yields the maximum utility under the deterministic assumption. This method can significantly reduce the computational complexity of the algorithm and eliminate the need of sharing information among the UAVs. To achieve enough exploration to guarantee optimal performance, $\epsilon$-greedy with a decaying exploration is adopted during the training process to select actions for the UAVs with the probability defined by a decaying function. The simulation results show that the proposed tabular $Q$-learning enables the UAVs to converge to an NE in terms of sum-rate in both 2-UAV scenarios and 6-UAV scenarios. However, the 6-UAV scenario takes more episodes to converge compared with the 2-UAV scenario since it has larger state space, and the tabular $Q$-learning can be slow to deal with such a state.

\emph{Spectrum access for cognitive radios: }Cognitive Radio (CR) is able to improve the spectrum efficiency~\cite{kumar2016spectrum}. However, the co-existence of the Primary Network (PN) and the Secondary Network (SN) in the Cognitive Radio Network (CRN) leads to aggregated interferences that impose a critical challenge on radio resource allocation. Various approaches such as heuristic algorithms and auction-based algorithms have been investigated to solve the resource allocation issue in CRNs~\cite{el2016resource}. However, these works typically assume that the model of network dynamics is perfectly characterized, which is impractical due to the stochastic behavior of the network user. MARL is capable of learning network dynamics and allows the CRs to cooperate with each other to maximize the network capacity or the system energy efficiency. Thus, it has been recently used for resource allocation in the CRNs.

The authors in~\cite{Kaur2020EnergyEfficientRA} propose to adopt a cooperative $Q$-learning and a cooperative SARSA to allocate spectrum resources for cognitive radios, respectively. The system model consists of a PU network and a CRN. The CRN shares the same spectrum with the PU network. The CR's problem is to select the spectrum resource as well as the transmission power so as to maximize the overall system energy efficiency. \textcolor{black}{The sequential decision-making problems of CRs are formulated as an MG, where each CR is an agent. The definition of MG can be found in Section \ref{MG}.} The state of each CR is the number of PUs and CRs in the network and a binary indicator that represents whether the SINR for the CR is higher than the defined threshold in the previous time slot. If the joint actions of all the CRs satisfy the spectrum usage constraint, the QoS requirements of CRs, the constraint on the interference to the PU, and the power limit constraint, all the CRs receive the energy efficiency as their common rewards. Otherwise, they receive rewards equal to "$0$". After receiving a reward, the CR uses the update rules of $Q$-learning and SARSA to update its $Q$-value with respect to stationarity policies of other CRs under weakly acyclicity to compute its best response policy. This allows the CRs to form a near-stationarity MDP and learn an optimal policy under the strict best response with inertia. The simulation results show that the proposed collaborative algorithms outperform the independent algorithms in terms of average capacity and average energy efficiency. In particular, the cooperative SARSA is able to achieve better energy efficiency compared with the cooperative $Q$-learning since SARSA does not suffer from the overestimation issue~\cite{hasselt2010double}.

\textcolor{black}{The tabular $Q$-learning and SARSA proposed in~\cite{Kaur2020EnergyEfficientRA} may suffer from the scalability issue. The reasons can be found in Section \ref{Challenges}.} In order to achieve better scalability, the authors in~\cite{Yang2020PartiallyOM} employ DQN and DDQN to jointly optimize the channel access and mode selection, e.g., the relay mode or the overlay mode, so as to maximize the sum rate in CRNs, respectively. The system model is similar to that in~\cite{Kaur2020EnergyEfficientRA}. \textcolor{black}{The authors formulate this problem as a Dec-POMDP, where each CR as an agent selects a channel and a mode depending on its local observation, i.e., the action acknowledgments (ACKs) and the rewards of all the agents in the previous time slot. The details of Dec-POMDP is introduced in Section \ref{Dec-POMDP}.} If the agent successfully accesses its selected channel and mode, it will receive a reward equal to "$1$". Otherwise, if the collision happens, it will receive a penalty equal to "$-1$". To collaborate with other agents, all the agents share the same copy of DQN for online execution, and the sum of each agent’s mean-square error is utilized to compute the multi-agent loss for updating the parameters of the DQN. To reduce the overestimation issue of DQN, DDQN is further adopted for modeling agents. The simulation results demonstrate that the proposed algorithm can scale well to overcome the issue caused by the increase in the number of CRs, and the shared DDQN-based policy obtains a higher success rate of packets transmission than that of the DQN-based policy. The shared DDQN can significantly reduce the number of training parameters and improve the collaboration performance of agents~\cite{terry2020parameter}. However, the mis-synchronization of agents' policies degrades the performance of the shared policy since the outdated policies executed by some agents may generate harmful experiences for training the current policy. Besides, it cannot be applied to networks with heterogeneous network entities.

\emph{Spectrum access for vehicular networks: }Efficient spectrum allocation is particularly important for some applications that require ultra-reliable communication, such as Vehicle-to-Vehicle (V2V) communication that provides security service for vehicles. The 3rd Generation Partnership Project (3GPP) standardized a semi-persistent scheduling (SPS)-based decentralized resource allocation method that each vehicle independently senses the channel state and chooses the resource block. However, this may cause serious interference among vehicles in mutual vicinity, e.g, undecoded cooperative awareness messages are repeatedly sent by the vehicles.

To solve this issue, the authors in~\cite{Gndoan2020DistributedRA} adopt an MADRL-based method to obtain the global optimal policy of spectrum allocation. Since the existence of a BS cannot always be guaranteed, the system model involves a cellular-free scenario of a vehicular network that consists of multiple vehicles, and none of them are connected to a BS. Each vehicle can independently select a spectrum resource according to its own state to exchange messages via V2V links. The common objective of all the vehicles is to learn an optimal policy that can maximize the Packet Reception Ratio (PRR)~\cite{3gpp2016technical}, i.e., the ratio of the successful receivers among the total number of the transmitter vehicle's neighbors. To achieve this goal, the optimization problem is modeled as a Dec-POMDP. The vehicle becomes an agent. The state of each vehicle is defined by its action taken in the previous time slot and the positional distributions of other vehicles. If the agent successfully accesses the selected spectrum without congestion, it receives a reward equal to "$1$". If the agent shares the same spectrum with a distant agent, the reward is set to "$0$" so as to improve the spectrum efficiency. To avoid interference and local congestion, if the agent shares the same spectrum with nearby agents, it receives a penalty equal to the number of the nearby agents. The DDQN algorithm is then adopted to learn the policy of each agent. The Long Short-Term Memory (LSTM)~\cite{hochreiter1997long} constructs the first layer of the DDQN to forecast the mobile pattern of vehicles and a fully connected layer is followed to estimate the $Q$-functions. Similar to ~\cite{Yang2020PartiallyOM}, the parameters of DDQN are shared among all the vehicles in the network due to the homogeneous property of the vehicles, which can significantly accelerate the convergence speed~\cite{gupta2017cooperative}. The simulation results reveal that the proposed algorithm improves the PRR in a congested scenario by up to $20$\% compared with SPS. However, the PRR drops quickly when the number of vehicles beyond the available resources.

To support the communications between the vehicles and infrastructures, e.g., a roadside unit based on BS, for high data rate service, the vehicular networks can consist of a number of Vehicle-to-Infrastructure (V2I) links. The spectrum sharing between V2V and V2I links is investigated in~\cite{Liang2019SpectrumSI}. The system model consists of multiple V2I and V2V links to provide high data rate entertainment service and reliable safety message exchange service. The V2I links use the cellular interfaces to connect the vehicles, while the V2V links transmit safety messages via the sidelink interfaces. The vehicle has a spectrum resources pool that can be used for V2V communication. To improve the spectrum efficiency, the V2I links can also reuse the spectrum resources in the pool to communicate with the BS. Each V2V link aims to learn an optimal spectrum access policy that can maximize the capacity of V2I links while improving the payload delivery reliability of the V2V links. The problem is modeled as a Dec-POMDP, where the V2V link is treated as an agent. The observation of each agent is its local channel information including its own signal channel as well as the interference channels from other V2V transmitters, its own transmitter to the BS, and all V2I transmitters. To stabilize the multi-agent learning environment, fingerprints that are able to track the changes of other agents' policies are added to the agent's observation. The common reward for all the agents is defined as the sum capacity of all the V2I links. Besides, to improve the probability of successful V2V payload delivery, the reward for each agent is defined as the transmission rate that the V2V link can achieve. 

Different from~\cite{Gndoan2020DistributedRA} that uses the shared DQN, in~\cite{Liang2019SpectrumSI}, each agent uses a dedicated DQN that takes the local channel information as input and output the corresponding $Q$-values of actions. This can significantly reduce the signaling overhead caused by parameter sharing. The simulation results demonstrate that the proposed dedicated DQN is able to achieve better results than that of the shared DQN used in~\cite{Gndoan2020DistributedRA} and the random baseline in terms of sum capacity and the successful transmission probability of the V2V links.

The agent information collecting mechanism utilized in the above works for centralized training, e.g.,~\cite{Li2020MultiAgentDR} and~\cite{Zhang2020DeepMR}, do not consider a preprocess of state information, which may lead to extra signaling overhead and increase the computational complexity at the BS. The state information preprocess is investigated in~\cite{Wang2020LearnTC}. To this end, each V2V link adopts a DNN to compress its local observation of currently available CSI and then feedback the compressed local CSI to the BS, which reduces the signaling overhead and mitigates the damaging influence of CSI feedback. \textcolor{black}{A fully centralized DQN is deployed at the BS to jointly optimize the actions of all V2V links by utilizing the feedback information from them. The details of the fully centralized learning scheme can be found in Section \ref{Challenges}.} Then, the BS broadcasts the decisions of actions to all the V2V links. Different from~\cite{Wang2020LearnTC}, the authors in~\cite{He2020ResourceAB} propose to use the Graph Neural Network (GNN)~\cite{zhou2020graph} for the state information preprocessing. For this, the V2V network is first represented by a graph where each V2V link, i.e., the agent, is treated as a node and the corresponding interference links are the edges of the graph. The GNN is then used to learn a feature embedding of each node in the defined graph, including the information of adjacent nodes and edges. After extracting features, the DQN is employed for each V2V link to select the spectrum resource via feeding the local observation of each agent, including the channel gain of V2V links and their power level, and the extracted features of nodes. 

The GNN used in~\cite{He2020ResourceAB} provides a novel decentralized way for compressing neighboring agents' state information without the demand for the global CSI in high-mobility environments compared with the fully centralized approach proposed in~\cite{Wang2020LearnTC}. The GNN is thus suitable for real-world applications since the high mobility of vehicles may lead to unreliable cellular links. However, similar to~\cite{Gndoan2020DistributedRA}, when the traffic jam of the neighboring nodes increases, the algorithm may suffer from poor scalability due to the high-dimensional input of neighbors' information. For future research, Graph Attention Networks (GAT)~\cite{velivckovic2017graph} can be directly used to improve the GNN since it can allocate different attention weights to the neighboring nodes embeddings according to their features via a masked self-attention mechanism, and output a single attention embedding.

\emph{\textbf{2) Joint user association and spectrum access: }}In HetNets with multiple BSs, each user needs to associate with a BS and select the spectrum from the BS. Thus, the joint spectrum access and user association should be considered. This problem is typically non-convex and needs the complete information of the network. MARL can be used to efficiently solve this problem.

The authors in~\cite{Zhao2019DeepRL} consider applying MARL to address the joint user association and spectrum allocation in HetNets. The system model consists of multiple Macro BSs (MBSs), Femto BSs (FBSs), Pico BSs (PBSs), and users (UEs). Different kinds of BSs have different coverage ranges and transmission power. To achieve a higher spectrum efficiency, all the BSs can share the same spectrum resources to serve UEs. Each UE's problem is to select one BS to associate with as well as a transmission channel so as to maximize its transmission capacity while satisfying a minimum QoS requirement. The sequential decision-making problems of UEs can be modeled as an MG. The state of each UE is represented as a boolean value that indicates whether the UE meets its QoS requirement in the previous time slot. The reward of each UE is defined as its utility minus the action selection cost in the current time slot. In particular, its utility is the difference between the transmission cost and the obtained profit, i.e., the sum transmission capacity achieved in the current time slot. 

At the beginning of each time slot, the UE takes an action on the transmission channel selection and the BS selection to meet its QoS requirement. Then the action decisions will be sent to the selected BS, and the UE will get an immediate reward if it receives the feedback signal from the BS. The decentralized Multi-Agent Dueling Double DQN (D$3$QN)~\cite{Wang2016DuelingNA} is applied for each UE to solve the MG, where the dueling neural network is used to obtain the estimation of a value function and an advantage function, respectively. The input of the D3QN is all the UEs’ states, which can be obtained through exchanging the uploaded state information of the associated UEs among BSs via the backhaul communication link. The simulation results show that the decentralized D3QN can obtain a higher system capacity than those of DQN, $Q$-learning, and genetic algorithm. Besides, it also has a faster convergence speed and better scalability compared with other RL algorithms. However, with the proposed learning algorithm, the UE may need to consume more computational and network resources to learn a policy compared with the centralized scheme.

As we discussed in~\cite{Shamsoshoara2019DistributedCS}, the communication links among UAVs cannot reliably exchange state information. The joint user association and spectrum allocation in multi-UAV downlink communication networks is investigated in~\cite{Cui2020MultiAgentRL}, where each UAV independently selects one user to associate with as well as the corresponding sub-channel and power level so as to maximize its throughput. \textcolor{black}{Considering the uncertainties of multi-UAV environments, the authors propose to use the IL-based MARL to solve the problem. The reasons for choosing IL and the explanation of IL can be found in Section \ref{Challenges}.}

Each UAV as an agent runs a standard tabular $Q$-learning and makes action decisions depending on its own partial observation on the SINR level. If the SINR provided by the UAV exceeds the predefined threshold, the UAV receives an immediate reward defined as the difference between its throughput achieved by the joint actions of all the UAVs, and its cost of power consumption. Otherwise, the reward is "0". Note that though no coordination issues are solved in IL-based MARL, IL-based MARL takes the effect of other UAVs into consideration compared with the single-agent RL. Moreover, the proposed algorithm is not scalable due to the utilization of the tabular $Q$-learning.

\subsection{Power Control}
This section reviews the applications of MARL for transmitting power control and user association in cellular systems, IoTs, and vehicular networks.

\emph{\textbf{1) Transmit power control: }}

\begin{figure}[h]
  \centering
  \includegraphics[scale=0.7]{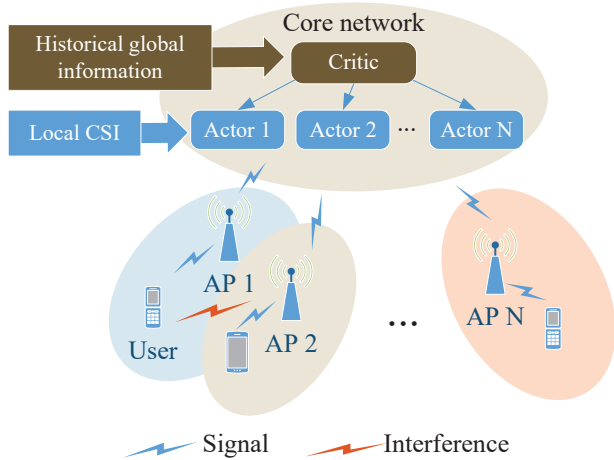}
  \caption{A HetNet with multiple APs where each AP cannot communicate with other APs to exchange information. A central critic using historical global information as input is utilized to train actors for all APs. Each actor only needs its local CSI to select the transmit power.}
  \label{Zhang2021}
\end{figure}

\emph{Power control for cellular systems: }In future networks, the ultra-dense deployment of BSs and Access Points (APs) in cellular systems can cause high interference and reduce the system capacity as well as QoS. To reduce the interference, transmit power control in such cellular systems has been investigated in~\cite{Zhang2021DeepRL} and~\cite{Khan2020CentralizedAD}. The authors in~\cite{Zhang2021DeepRL} consider the power allocation issue in a HetNet. The system model as shown in Fig. \ref{Zhang2021} consists of multiple APs that share the same spectrum band to serve users. Each AP's problem is to allocate the transmit power to its serving user so as to maximize the overall downlink sum rate. The APs cannot communicate with each other, and thus they cannot exchange information with neighbors for cooperation. To collaboratively learn an optimal power allocation policy, a CTDE-based Multiple-Actor-Shared-Critic (MASC) algorithm is adopted for power allocation, where each AP as an agent has a unique actor and shares a public critic with other APs. The actor of each AP is trained in the core network by the public critic through exploiting the experiences of all the APs, and then each local AP downloads its actor from the core network periodically to select the power level depending on its local state. The local state of each AP includes (i) the channel gain to the serving UE in the current time slot and the previous time slot, (ii) the interference to the serving UE in the current time slot and the previous time slot, (iii) the transmit power in the previous time slot, (iv) the downlink transmission rate from the AP to the UE in the previous time slot, and (v) the SINR at the serving UE in the previous time slot. After the action is executed, the reward the AP receives is equal to its transmission rate. The simulation results reveal the superiority of MASC in both average sum-rate and computational complexity compared with conventional power control methods like WMMSE~\cite{Shi2011AnIW} and FP~\cite{Shen2018FractionalPF}.  

Different from the CTDE-based MASC used in~\cite{Zhang2021DeepRL}, a partially decentralized power allocation method based on TRPO~\cite{Schulman2015TrustRP} is proposed in~\cite{Khan2020CentralizedAD} for power control. The system model is a time-division duplexed network that consists of multiple single-antenna BSs and multiple uniformly distributed single-antenna users. The goal of each BS is to allocate transmit power to its serving users so as to maximize the sum rate of the network. To achieve the goal, the power control decision problem is formulated as an MDP. The agent is the BS. The state of each BS is the local CSI of its own downlink and the previous BSs’ power allocation in the current round. All the agents learn the optimal policy under the common reward, which is defined as network spectral efficiency. Different from the MASC used in~\cite{Zhang2021DeepRL}, the TRPO method learns a shared policy for each BS, which reduces the training parameters and accelerates the convergence speed~\cite{gupta2017cooperative}. To control the learning step and achieve a more robust power control algorithm, the trust-region method alters the ascent direction of policy while exploiting the decaying step sizes, which can avoid large policy updates dramatically~\cite{Schulman2016HighDimensionalCC}. The simulation results show that the TRPO-based partially decentralized algorithm generally achieves a higher network sum rate than those obtained by the conventional centralized methods like WMMSE~\cite{Shi2011AnIW}, FP~\cite{Shen2018FractionalPF}, without global CSI requirement. Moreover, the shared policy only needs to compute an action for a single BS depending on its local state, which circumvents the scalability issue of the centralized method caused by the increasing state and action spaces as the number of BSs increases.

Learning directly the observed local CSI through the MADRL model such as~\cite{Khan2020CentralizedAD} and~\cite{Zhang2021DeepRL} may increase the computational complexity and even incur performance degradation since the observed local CSI typically consists of a large amount of information of weak-correlated interference signals. Thus, the selection of local CSI is significant, which is investigated in~\cite{Meng2020PowerAI}. The authors consider a single-input single-output (SISO) system with multiple cells, and each cell has one BS with a single antenna to serve the users. The problem of the transmitter of each cellular link is to select the transmit power so as to maximize its data rate. MARL methods are then adopted to solve the problem, and the transmitter of each cellular link is regarded as an agent. The current local CSI, i.e., the local state, of each transmitter is pre-processed by a logarithmic normalized process, and then the normalized results are sorted to find the main interference signals to approximate the interferences in the network. This can significantly reduce the state space due to the fact that the number of the main interference signals is typically far less than the total interference signals. Therefore, the computational complexity can be reduced. Similar to the policy sharing scheme used in~\cite{Khan2020CentralizedAD}, all the transmitters share a single policy and learn with the collected experiences from all the agents. The authors then adopt three MARL algorithms to implement the shared single policy: \emph{REINFORCE}, DQL, and DDPG, respectively. Both \emph{REINFORCE} and DQL can only deal with discrete power levels which may result in quantization error. DDPG, in contrast, can use a deterministic policy to output a continuous power level for each agent. The simulation results show that DDPG outperforms \emph{REINFORCE} and DQL in terms of the sum rate and robustness. 

The interference sorting technique can also be found in~\cite{Nasir2019MultiAgentDR}. Instead of sorting the interference signals used in~\cite{Meng2020PowerAI}, the authors in~\cite{Nasir2019MultiAgentDR} sort the transmitter's interferers according to the received power from the interferers at the corresponding receiver. This sorting process can determine interferers that have high impacts on the current transmitter and put them into the interferer set with a fixed size. This enables the power control algorithm to scale to various user densities. Besides, the authors also defined an interfered neighbor set following the same procedure. However, this time the sorting standard is the proportion of the current transmitter's interferences at the interfered receivers. 

The system model in~\cite{Nasir2019MultiAgentDR} involves a HetNet with multiple APs and users. Each AP is connected to a single user via a transmission link. The transmitter of each link can automatically select a transmit power for transmission. The objective is to find an optimal transmit power decision policy for each transmitter that maximizes the weighted sum rate under the transmit power constraint to ensure proportional fairness~\cite{zhang2011weighted}. This problem is total non-convex and has been proved to be NP-hard~\cite{luo2008dynamic}. Thus, the MARL method has been used to solve the problem. The agent is the transmitter of each link. The state of each transmitter is defined as the downlink channel gain to its receiver, the total received interference-plus-noise power at its receiver, and the received power at its receiver from every interferer in the defined interferers set. In addition to the local information, the transmitter can also utilize the interference information from its interferer set, the feedback, e.g., the channel gain and the share of current transmitter on the interference at the interfered receivers, from its interfered neighbor set, as well as their contribution to the objective of weighted sum-rate, i.e., the weight of the link and the spectral efficiency. Similar to the policy sharing scheme as proposed in~\cite{Khan2020CentralizedAD} and~\cite{Meng2020PowerAI}, the authors propose a shared DQN that is trained by using all the transmitters' experiences. This can improve the learning efficiency of DQN dramatically. The simulation results show that the proposed algorithm can achieve a higher average network rate than WMMSE~\cite{Shi2011AnIW}, FP~\cite{Shen2018FractionalPF}, and the central power allocation algorithm. However, the experience sharing mechanism may produce a huge signaling overhead for experience exchange.

\emph{Power control for IoTs: }IoTs such as Wireless Sensor Networks (WSNs) typically involve a large number of sensor nodes. For cooperation, the state information, e.g., the battery level and the local CSI, exchange among sensor nodes typically leads to high energy consumption, which is infeasible for the IoT nodes with energy constraints. To address this issue, the authors in~\cite{Sharma2019MultiD} propose a decentralized Mean-Field MARL (MF-MARL) algorithm for the power allocation in large energy harvesting (EH) networks, where multiple identical EH nodes aim to transmit their data to an AP. The problem of each node is to select a transmit power level so as to maximize the time-averaged sum throughput of the network, which is modeled as a finite state discrete time mean-field game~\cite{Gomes2010DiscreteTF}. The agent is the EH node. The state of the node includes (i) its battery level, (ii) its harvested energy, and (iii) the fading channel gain between it and the AP. As such, the EH node does not need to exchange state information with other nodes, thereby reducing the communication overhead of EH nodes. The reward is defined as a function of the channel gain between the nodes and the AP as well as the energy level of nodes utilized for transmission. Given the ultra-dense EH nodes, MF-MARL is utilized to learn a stationarity NE of the mean-field game via a fictitious play procedure. In particular, for a given belief distribution, the MF-MARL leverages the $Q$-learning algorithm to learn a Nash maximizer at each iteration, and then the belief distribution is updated correspondingly. The simulation results demonstrate that the proposed decentralized algorithm can achieve near-optimal performance in terms of sum throughput compared with the state-of-the-art centralized methods, which utilizes all the nodes' state in the network as input. However, the MF-MARL requires discrete actions, and thus the primitive action space, e.g., the transmit power range, has to be quantized which incurs the loss in the throughput.

\emph{Power control for vehicular networks: }The rapid proliferation of vehicles may lead to serious interference among V2V links, which leads to the need for an efficient power control approach to improve the network performance while guaranteeing the QoS of V2V links. 

The power allocation in D2D-enabled V2V communications is investigated in~\cite{Nguyen2019DistributedDD}. The system model consists of multiple V2V links that are randomly distributed within the coverage of one BS. The vehicles can be directly connected to each other via D2D-enabled V2V links, instead of the BS. The goal of each vehicle is to select a transmit power for data transmission so as to maximize the total energy efficiency of the network while meeting the QoS constraints of V2V links, which is formulated as an MDP. Each V2V transmitter acts as an agent to select transmit power depending on its state, which is defined by its index and interference level. If the agent's selected power level satisfies its QoS constraint, it will receive a reward that is equal to the energy efficiency. Otherwise, the reward will be zero. In the multi-agent environment, to solve the power selections of the V2V transmitters, the SDDPG algorithm is adopted, rather than the standard DDPG and distributed DDPG, since it allows the agents to share a single DDPG policy with each other, that can significantly reduce the computational complexity and accelerate the convergence speed~\cite{gupta2017cooperative}. The simulation results show that the energy efficiency obtained by the SDDPG is higher than that obtained by the standard DDPG and equal to the distributed DDPG. However, the SDDPG has a faster convergence speed compared with the distributed DDPG.

In cellular-based vehicular networks, each V2V transmitter is allowed to select different communication modes for its data transmission, which makes the power allocation non-convex. The authors in~\cite{Zhao2020ARL} propose to use MARL to jointly optimize mode selection and power control in cellular-based vehicular networks. The system model consists of one BS and multiple VUEs covered by the BS. Three different communication modes are available for the V2V links: cellular mode for long-distance communication, dedicated mode, and reuse mode for D2D links. In particular, the dedicated mode can use the unoccupied sub-band for communication, while the reuse mode is capable of reusing the sub-band occupied by V2I links. 

The problem of each V2V link is to learn an optimal transmit power and mode selection policy so as to maximize the total capacities of the V2I links while guaranteeing the ultra-reliable and low latency requirements of V2V links. The problem is modeled as a Dec-POMDP in which each V2V link is an agent. The common reward for all the agents is defined as a function of the total capacities of the V2I and V2V links, the connectivity probability, and the transmission delay. The V2V link selects the power level and the transmission mode in a decentralized way based on its local state. This local state includes the fading parameters and statistical information of the channel state rather than the instantaneous CSI, which is more practical in real-world applications. Besides, the mode selection states of neighboring agents in the previous time slot are also taken into consideration to overcome the stationarity issue. Based on the collected state information, the agent uses the standard DDQN to learn its policy. Numerical results demonstrate that the proposed decentralized DDQN outperforms the centralized DQN and the random search method in terms of the sum-throughput of V2I links and the probability of satisfying V2V links under various vehicle speed values. Moreover, the decentralized DDQN enables the agent to achieve a near-optimal solution compared with the optimal solution with the full knowledge of the environment information.

\emph{\textbf{2) Joint user association and power control: }}In a downlink wireless network, an AP can select an UE it serves to transmit data according to the SINR at the UE. Thus, the joint user association and power control should be considered as investigated in~\cite{Naderializadeh2020ResourceMI}. The system model involves multiple APs and UEs. Each AP serves a number of users and aims to transmit data to the UEs associated with it. Thus, at the beginning of each time slot, the AP can independently select one UE from its local user pool as well as the corresponding transmit power in a distributed manner so as to maximize the long-term weighted sum rate. To achieve the goal, the MARL method is adopted for each AP to learn an optimal policy. The agent is the AP. The observation of each AP is the weight and the SINR of its associated UE. In particular, the weight of each UE is defined as the inverse of its long-term average rate. To collaborate with neighboring APs, the AP can exchange observation information with them. To design a scalable algorithm, the number of neighboring APs is fixed and selected by a distance-based method, which enables the observation dimension of the AP to not vary with the AP density. The centralized reward based on all the UEs' actions is defined as a weighted sum rate, which is computed by the weights and the transmit rates of the UEs. This renders the reward function gives more attention to the local transmit rate of each UE instead of only focusing on the average reward that may result in unfair resource allocation. The authors adopt the DDQN and the A2C to learn an optimal policy for the AP, respectively. The simulation results show that although each AP takes actions in a decentralized manner, it can still achieve a similar sum rate to a centralized information-theoretic method~\cite{naderializadeh2014itlinq}. Besides, the A2C achieves a better sum rate than that of DDQN. However, DDQN has a faster convergence speed since it has a better sample efficiency thanks to the replay buffer.

Due to the limited ranges of BSs, the moving UEs need to switch their serving BS for meeting its QoS requirement. The authors in~\cite{Guo2020JointOO} consider adopting MADRL to solve the handover and power allocation problem in a two-tier HetNet. The system model consists of an MBS, multiple small base stations (SBSs), and UEs. Each UE can be either served by one SBS or one MBS. The problem of each UE is to make decisions on the BS selection and the transmit power selection so as to maximize the system throughput while decreasing the handover frequency. To this end, the sequential decision-making problem of UE is modeled as Dec-POMDP, where each UE is treated as an agent. The observation of each UE includes its BS connection, SINR measurements from the BSs to the current UE, and the number of UEs served by each BS that can be obtained from the MBS. The common reward for all the UEs is defined as a function of the system throughput and a handover penalty achieved by the action selections of all the UEs. 

A PPO~\cite{Schulman2017ProximalPO}-based cooperative scheme MAPPO is proposed to help the UE learn an optimal decision-making policy. \textcolor{black}{The MAPPO uses the CTDE framework to train the UEs, and the sampled joint trajectories of all the UEs collected from the SBSs by a central processor (CP) are used to train decentralized policies for each UE to overcome the non-stationarity issue that is defined in Section \ref{Challenges}.} Moreover, the MAPPO solves the credit assignment problem through counterfactual baseline~\cite{foerster2018counterfactual} proposed in COMA which marginalizes out the current UE’s action while keeping other UEs’ actions fixed. This can help the UE figure out its contribution to the centralized reward, thereby improving its performance. The simulation results reveal that the proposed MAPPO outperforms the IL-based PPO and some other existing methods like MADDPG, A$3$C in terms of both the individual throughput of the UE and the total throughput of the network. Moreover, the proposed MAPPO significantly reduces the handover frequency due to the introduction of a penalty in the reward function. 

\textbf{Summary: }This section reviews the application of MARL in spectrum access and power control. These works are briefly summarized in Table \ref{tab:networkaccess} and Table \ref{tab:powercontrol}, from which we can see that the problems are mostly modeled as Dec-POMDP where each agent, e.g., D2D pair, UAV, UE, or AP, can only observe its local states such as the local CSI, previous channel or power selection, and IoT node's battery level. To solve the non-stationarity issue caused by Dec-POMDP, most works adopt CTDE or IL scheme. In particular, the CTDE scheme used in these works is capable of using the global state information collected by a local agent, a central unit, or a cloud platform for offline training. Moreover, CTDE is usually used in the cooperative scenario, while IL can be used not only in the competitive scenario but also in the cooperative scenario to achieve the global optimal solutions via the projection characteristic of deterministic MMDP or a common reward based on all the agents' actions, etc. However, few works consider the credit assignment issue for collaborative agents, which can be addressed by using the counterfactual baseline methods or the value deposition methods as mentioned in Section \ref{MARLalgorithms}.

%table IV 

\begin{table*}[]\scriptsize\centering
\renewcommand\arraystretch{1.5}
\caption{A summary of approaches applying MARL for network access.}
\label{tab:networkaccess}
\begin{tabular}{|m{0.5cm}<{\centering}|m{0.6cm}<{\centering}|m{3cm}<{\centering}|m{1.5cm}<{\centering}|m{1.5cm}<{\centering}|m{1.5cm}<{\centering}|m{1cm}<{\centering}|m{1.5cm}<{\centering}|m{2cm}<{\centering}|m{1cm}<{\centering}|}
\hline
\rowcolor[HTML]{9B9B9B}
\textbf{Issues}            & \textbf{Works}      & \textbf{Main Findings}                    & \textbf{Model}     & \textbf{Learning Schemes}      & \textbf{Algorithms}                & \textbf{Private or Shared Policy} & \textbf{Cooperation or Competition}  & \textbf{Learning Objectives}   & \textbf{Network Models}                \\ \hline
\multirow{12}{*}{{\rotatebox{90}{Network access\quad \quad \quad \quad \quad \quad \quad \quad \quad \quad \quad \quad \quad \quad \quad \quad \quad \quad \quad \quad \quad \quad \quad \quad \quad}}} & ~\cite{Zia2019ADM}      & \textcolor{black}{Proposes an independent spectrum sharing algorithm for D2D-enabled cellular networks to allocate the spectrum resources}              & POMDP             & IL                   & $Q$-learning                   & Private                  & Competition                & Network throughput maximization & HetNet               \\ \cline{2-10}
                                 & ~\cite{Li2020MultiAgentDR}      & \textcolor{black}{Considers a cooperative spectrum allocation algorithm for HetNets}       & Dec-POMDP             & CTDE               & MAAC \& NAAC                    & Private                  & Cooperation                & Network throughput maximization & HetNet               \\ \cline{2-10}
                                  & ~\cite{Zhang2020DeepMR}      &  \textcolor{black}{Learns joint spectrum assignment and power allocation policies in D2D-enabled cellular systems with neighboring agents' states}   & Dec-POMDP             & CTDE               & DDQN                    & Private                  & Cooperation                & Spectrum efficiency maximization & Cellular system              \\ \cline{2-10}
                                 & ~\cite{Shamsoshoara2019DistributedCS} & \textcolor{black}{Proposes a distributed cooperative spectrum sharing algorithm in UAV networks with MARL} & Deterministic MMDP & IL                   & $Q$-learning                   & Private                  & Cooperation                &System utility maximization & UAV                   \\ \cline{2-10}
                                 & ~\cite{Kaur2020EnergyEfficientRA} & \textcolor{black}{Adopts a cooperative MARL algorithm to allocate spectrum resource for cognitive radios}  & Dec-POMDP & CTDE                   & $Q$-learning \& SARSA                  & Private                  & Cooperation                & System energy efficiency maximization&CRN                   \\ \cline{2-10}
                                 & ~\cite{Yang2020PartiallyOM} & \textcolor{black}{Studies the resource management for Cognitive Networks with partially observable setting}  & Dec-POMDP & CTDE                   & DQN \& DDQN                   & Shared                  & Cooperation                & Network throughput maximization & CRN                   \\ \cline{2-10}
                                 & ~\cite{Gndoan2020DistributedRA} & \textcolor{black}{Solves the spectrum allocation problem in cellular free vehicular networks with MADRL}      & Dec-POMDP             & CTDE                 & DDQN                         & Shared                   & Cooperation                &PRR maximization & Vehicular network     \\ \cline{2-10}
                                 & ~\cite{Liang2019SpectrumSI}   & \textcolor{black}{Investigates the spectrum sharing between V2V and V2I links with MARL}    & Dec-POMDP             & IL                   & DQN                          & Private                  & Cooperation                & Capacity of V2I links maximization&Vehicular network     \\ \cline{2-10}
                                 & ~\cite{Wang2020LearnTC}     &   \textcolor{black}{Develops the CSI compression and proposes a resource allocation algorithm for vehicular networks}     & MDP               & Fully centralized          & DQN                          & Shared                   & Cooperation                &Network throughput maximization& Vehicular network     \\ \cline{2-10}
                                 & ~\cite{He2020ResourceAB}    & \textcolor{black}{Adopts a GNN to preprocess state information and allocate spectrum resource in vehicular networks}       & Dec-POMDP             & CTDE                 & DQN                          & Private                  & Cooperation                & Network throughput maximization&Vehicular network     \\ \cline{2-10}
                                 & ~\cite{Zhao2019DeepRL}     & \textcolor{black}{Uses MARL to solve the joint user association and spectrum access problem in HetNets}     & Dec-POMDP                & CTDE                 & D3QN                         & Private                  & Cooperation                & UE's transmission capacity maximization &HetNet               \\ \cline{2-10}
                                 & ~\cite{Cui2020MultiAgentRL}   & \textcolor{black}{Studies the joint user association and spectrum allocation with MARL in multi-UAV downlink communication networks}        & POMDP             & IL                   & $Q$-learning                   & Private                  & Competition                & UAV's throughput maximization& UAV         \\ \hline
\end{tabular}
\end{table*}

%table V

\begin{table*}[]\scriptsize\centering
\renewcommand\arraystretch{1.5}
\caption{A summary of approaches applying MARL for power control.}
\label{tab:powercontrol}
\begin{tabular}{|m{0.5cm}<{\centering}|m{0.6cm}<{\centering}|m{3cm}<{\centering}|m{1.5cm}<{\centering}|m{1.5cm}<{\centering}|m{1.5cm}<{\centering}|m{1cm}<{\centering}|m{1.5cm}<{\centering}|m{2cm}<{\centering}|m{1cm}<{\centering}|}
\hline
\rowcolor[HTML]{9B9B9B}
\textbf{Issues}            & \textbf{Works}      & \textbf{Main Findings}                    & \textbf{Model}     & \textbf{Learning Schemes}      & \textbf{Algorithms}                & \textbf{Private or Shared Policy} & \textbf{Cooperation or Competition}  & \textbf{Learning Objectives}   & \textbf{Network Models}                \\ \hline
\multirow{9}{*}{{\rotatebox{90}{Power control\quad \quad \quad \quad \quad \quad \quad \quad \quad \quad \quad \quad \quad \quad \quad \quad \quad}}}  & ~\cite{Nasir2019MultiAgentDR}  & \textcolor{black}{Proposes a interference sorting technique to deal with the interference information for dynamic power allocation in wireless networks}       & Dec-POMDP             & Decentralized with Networked Agents                & DQN                          & Shared                   & Cooperation                &Weighted sum-rate maximization& Mobile ad-hoc network \\ \cline{2-10}
                                 & ~\cite{Zhang2021DeepRL}     & \textcolor{black}{Applies MARL to solve the power control issue in HetNets}         & Dec-POMDP             & CTDE                 & MASC                         & Private                  & Cooperation                & Overall downlink sum-rate maximization&HetNet               \\ \cline{2-10}
                                 & ~\cite{Meng2020PowerAI}   &  \textcolor{black}{Uses a logarithmic normalized process to select local CSI for power control in SISO systems}          & Dec-POMDP             & CTDE                 & \emph{REINFORCE} \& DQN \& DDPG           & Shared                   & Cooperation                &Transmitter's data rate maximization& HetNet               \\ \cline{2-10}
                                 & ~\cite{Khan2020CentralizedAD} & \textcolor{black}{Considers using a single shared DQN for all agents to allocate power in cellular systems}       & Dec-POMDP             & CTDE                 & TRPO & Shared                  & Cooperation                &Network throughput maximization & Cellular system       \\ \cline{2-10}
                                 & ~\cite{Sharma2019MultiD}   &  \textcolor{black}{Proposes a decentralized MARL algorithm for power allocation in large scale energy harvesting networks}           & Mean-field game             & IL & MF-MARL                      & Private                  & Cooperation                &Time-averaged sum throughput maximization& IoT                   \\ \cline{2-10}
                                 
                                 & ~\cite{Nguyen2019DistributedDD}  & \textcolor{black}{Investigates the power allocation with MARL in D2D-enabled V2V communications}     & Dec-POMDP             & IL & SDDPG                        & Shared                   & Cooperation                & Network energy efficiency maximization& Vehicular network    \\ \cline{2-10}
                                 & ~\cite{Zhao2020ARL}   & \textcolor{black}{Jointly optimizes mode selection and power control using MARL in cellular-based vehicular networks}               & Dec-POMDP             & CTDE                 & DDQN                         & Private                  & Cooperation                &Total capacities of the V2I links maximization& Vehicular network     \\ \cline{2-10}
                                 & ~\cite{Naderializadeh2020ResourceMI} & \textcolor{black}{Studies the joint user association and power control using a cooperative algorithm in a downlink wireless network} & Dec-POMDP             & Decentralized with Networked Agents                 & DDQN \& A2C                     & Private                  & Cooperation                &Weighted sum-rate maximization& Cellular system       \\ \cline{2-10}
                                 & ~\cite{Guo2020JointOO}  & \textcolor{black}{Solves the handover control and power allocation problem using MADRL in HetNets}              & Dec-POMDP             & CTDE                 & MAPPO                        & Private                  & Cooperation                & Network throughput maximization&HetNet      \\ \hline
\end{tabular}
\end{table*}

\section{Offloading and Caching}\label{Offloading and Caching}
Task offloading and content caching techniques are emerging solutions typically used in MEC networks that can significantly improve the task processing ability and reduce data access latency of edge devices. This is promising for meeting the increasing QoS demands of emerging mobile applications that require low latency and intensive computations. Given the decentralized and dynamic nature of MEC networks, MARL can provide collaborative and decentralized solutions for offloading and caching decisions. In this section, we review the applications of MARL for computation offloading and content caching.

\begin{itemize}
\item \emph{\textbf{Computation offloading: }}Mobile devices (MDs), such as smartphones, inherently have limited computation ability, battery capacity, and memory size. The MEC/edge server is able to provide a more efficient computation service for nearby MDs with computation-demanding and latency-critical tasks compared with conventional cloud computing services, which can reduce the task execution delay dramatically. Each MD needs to decide whether to offload its tasks to the MEC servers and which MEC server to offload. However, this is challenging since multiple MDs may compete for the limited computational resource of one MEC server, which may incur longer task execution delays or even task failures. To address the issue, some recent works such as~\cite{flores2017evidence} and~\cite{liu2017latency} are dedicated to directly solving the optimization problem using techniques such as convex optimization and Lyapunov optimization, to optimize the task failure rate, the execution delay, the energy consumption, or a combination of them. However, they typically need the full knowledge of environment dynamics and may suffer from the slow convergence speed in MEC networks with dynamic computation tasks. MARL as an effective tool can be used to learn an optimal offloading policy for the MD without the knowledge of the environment.

 \item \emph{\textbf{Content caching: }}Apart from the offloading, the MEC servers can provide content caching service through pre-caching the popular contents from the content server, to reduce the content access latency of the nearby users and the core network traffic congestion. Due to the limited caching capacity, the problem of each MEC server is to decide which contents to cache so as to achieve the objectives of different networks including, e.g., minimizing the communication cost and content access latency. The existing works focus mostly on investigating the content caching with a single MEC server or a single user~\cite{zhong2018deep},~\cite{he2017cache},~\cite{he2017optimization}. However, the real-world network environment may typically involve multiple MEC servers and multiple users, and thus these results are insufficient to be used. MARL enables multiple MEC servers to collaborate with each other by sharing caching status and requested contents to serve the users. This can enlarge the caching capacity of MEC servers and further improve the QoS of users.

\end{itemize}

\subsection{Computation Offloading}
The computation offloading problem in a non-cooperative MEC system is investigated in~\cite{Heydari2019DynamicTO}, where each MD learns its offloading policy independently. The system model consists of multiple MDs and MEC servers. The MDs have limited computation and battery capability, and they can offload their tasks in their task queue to the MEC servers. All the tasks of the MD in its task queue should be executed in the current time slot, otherwise, they will be dropped. At the beginning of each time slot, each MD decides whether to execute its tasks locally or offload them to one of the MEC servers for execution, and the amount of energy to be consumed to execute or offload its tasks. The objective is to balance the task drop rate and the execution delay so as to minimize the MD's long-term expected cost. In particular, the cost for each MD is defined as the sum of the execution delay cost and the task drop cost. The task offloading problem is modeled as a Dec-POMDP since the offloading actions of other MDs are unknown to the current MD. The MD is the agent. The state of each MD includes the task queue length, the battery level, the channel gains, and the waiting time of MEC servers. Moreover, to overcome the non-stationarity issue caused by the changes of other MDs' policies, the authors add the execution delay of all MEC servers in the previous time slot to the current state of MDs. Directly solving this task offloading problem may result in sub-optimal policies since the users may spend more energy to finish the existing tasks to reduce the task execution delay, which may drain the battery and incur a high task drop rate in the future. Thus, with the consideration of long-term reward, the A2C~\cite{Mnih2016AsynchronousMF} algorithm is adopted for each MD to learn the optimal offloading policy. Compared to the DQN algorithm, it enables the agent to learn a more stable policy with low-variance gradient estimates. The simulation results show that the A2C-based non-cooperative algorithm can efficiently reduce the execution delay, the task drop rate, and the mobile CPU usage compared with the naive only-local and only offloading policies, and the DQN algorithm. 

 \begin{figure}[h]
  \centering
  \subfigure[]{
  \includegraphics[width=0.22\textwidth]{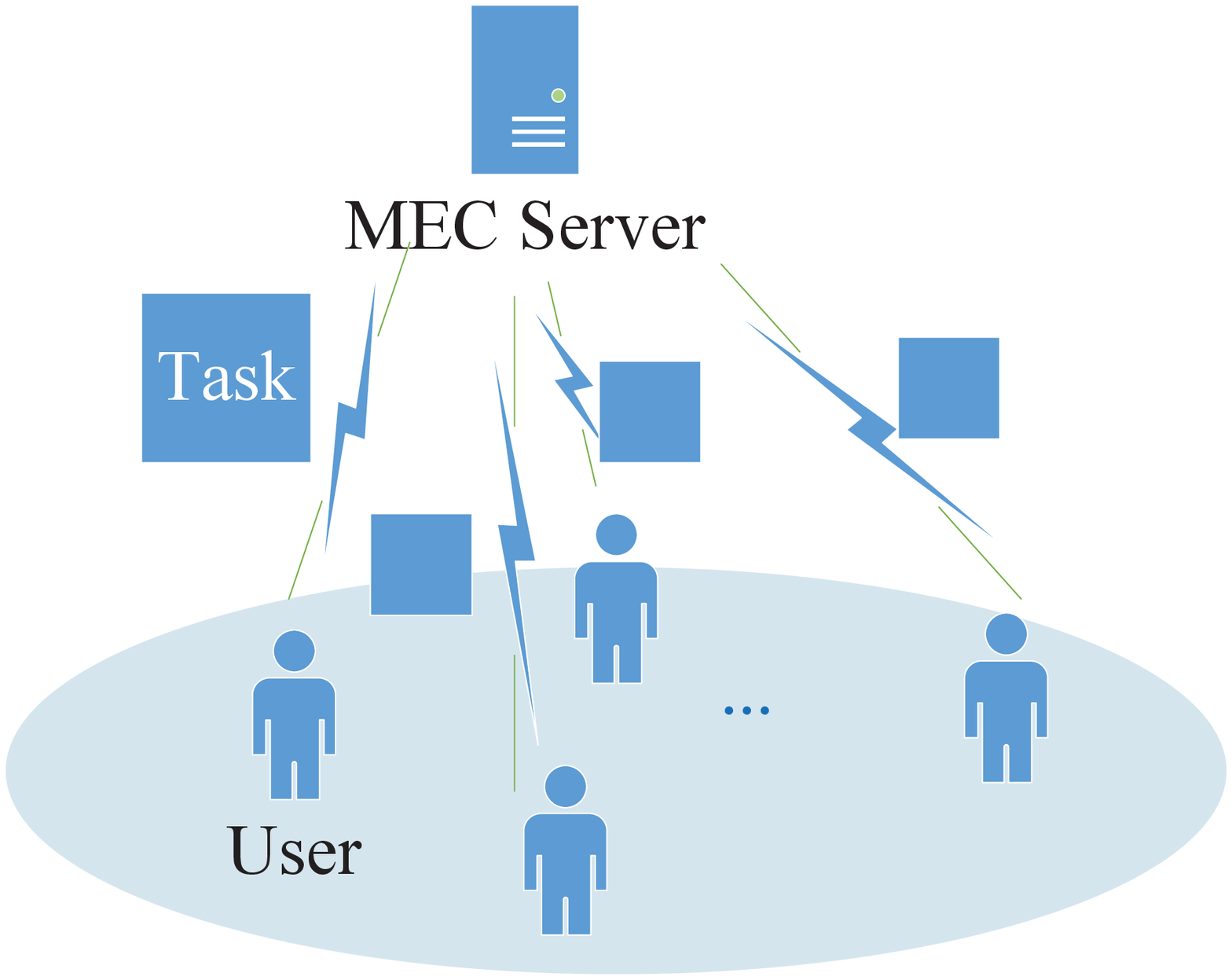}
}
  \subfigure[]{
  \includegraphics[width=0.22\textwidth]{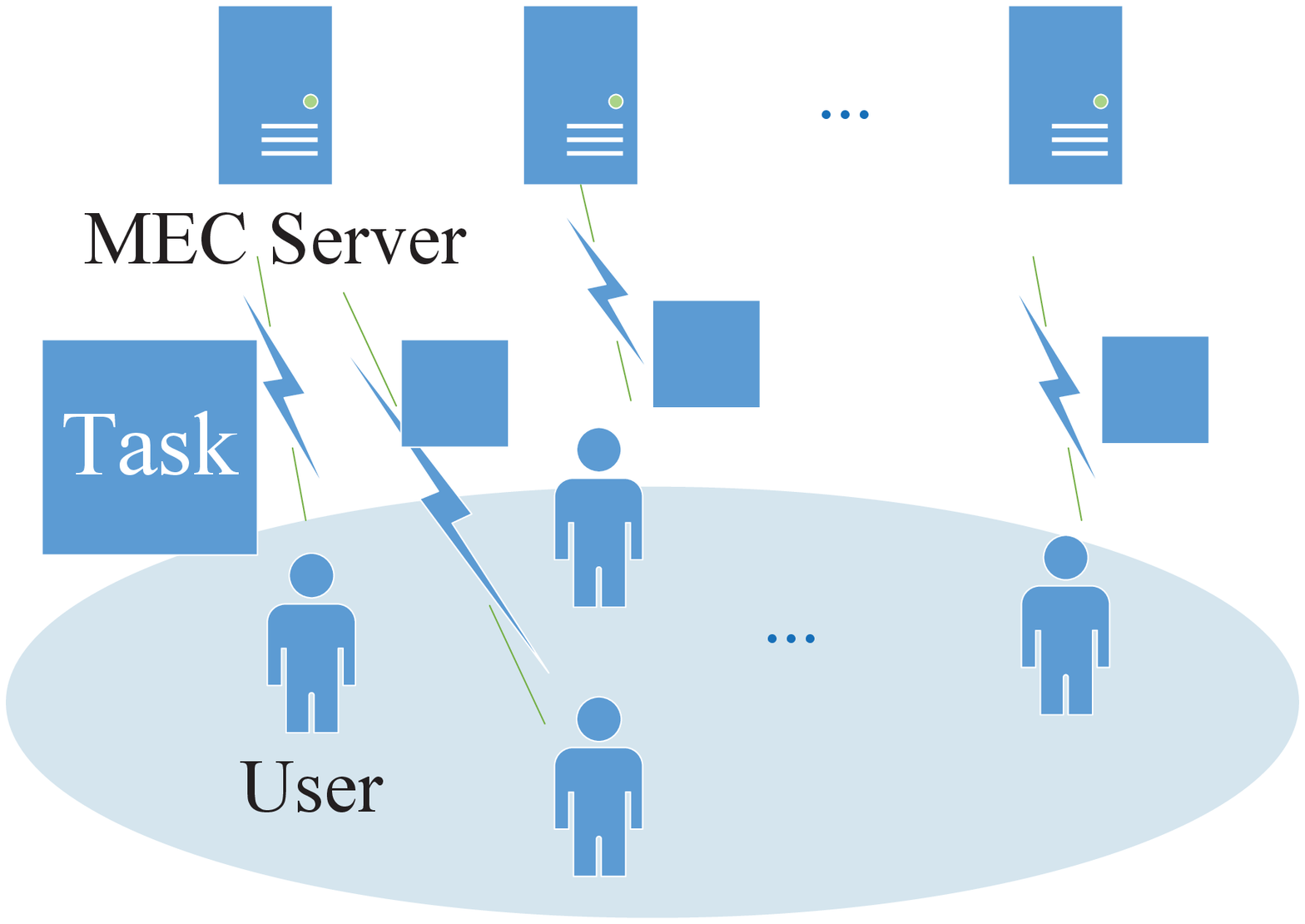}
}
  \subfigure[]{
  \includegraphics[width=0.22\textwidth]{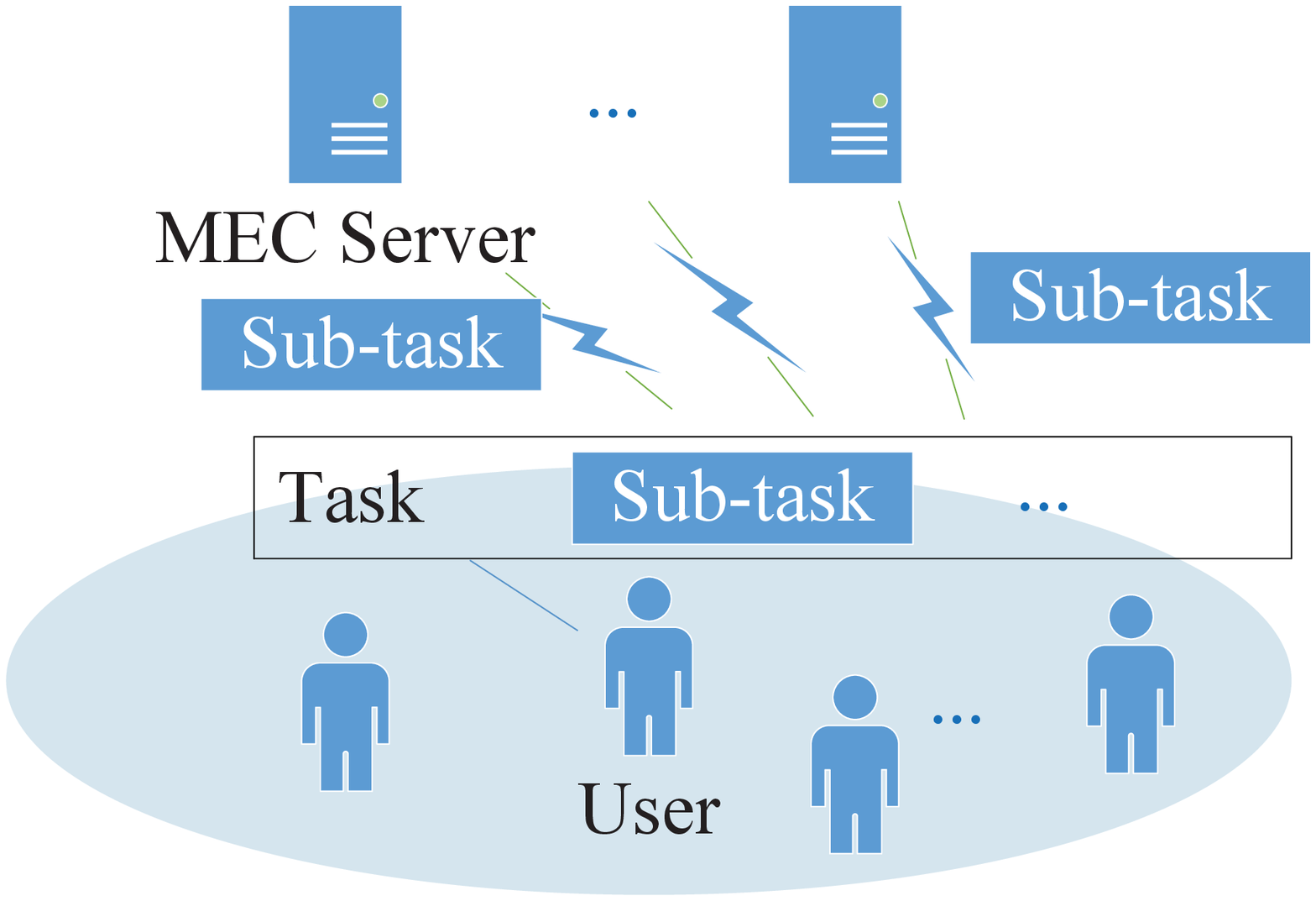}

}
\subfigure[]{
  \includegraphics[width=0.22\textwidth]{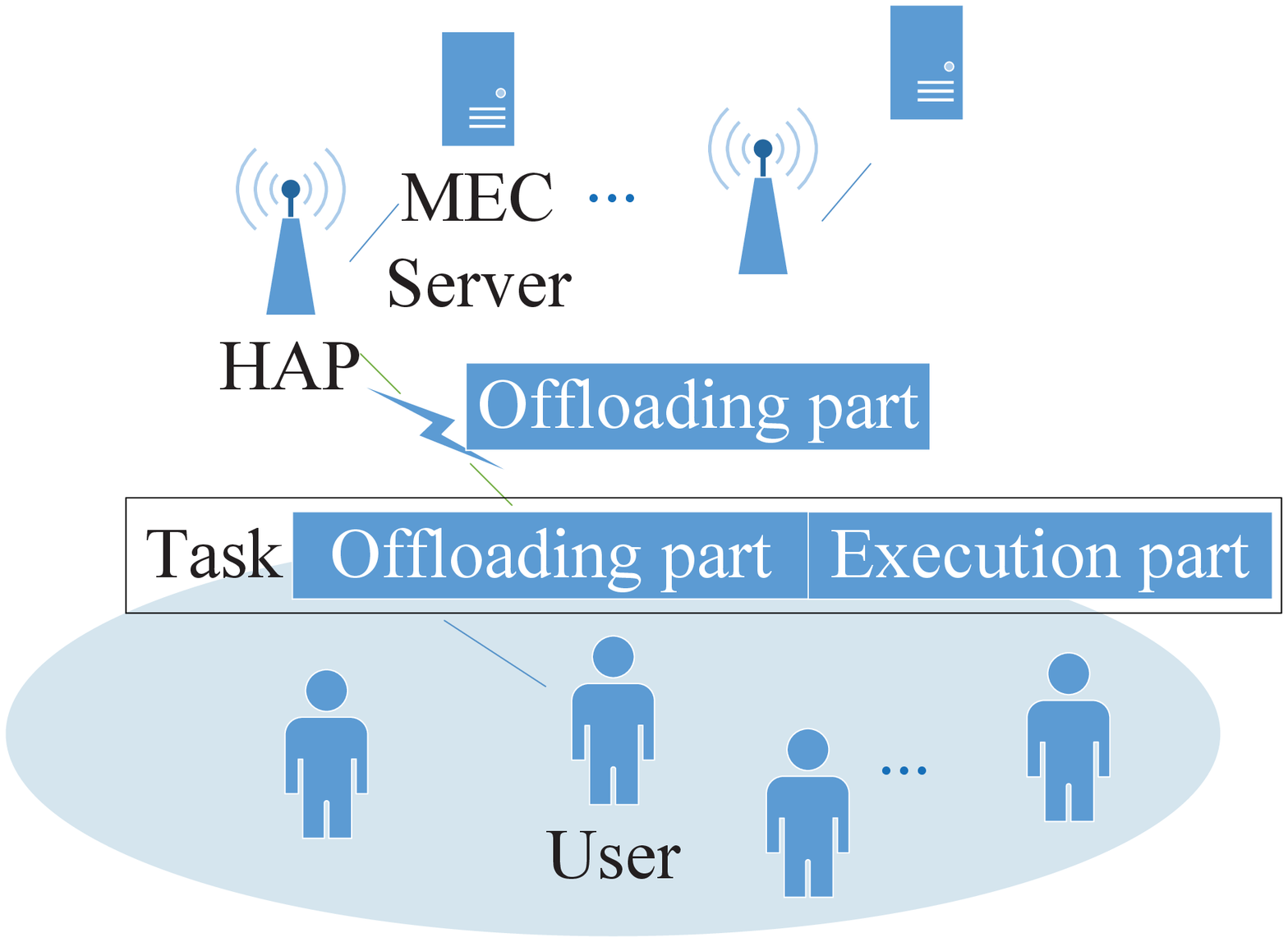}

}

\caption{Task offloading models: (a) offloading tasks to one MEC server~\cite{Cao2020MultiagentDR}, (b) offloading tasks to multiple MEC servers~\cite{Heydari2019DynamicTO},~\cite{Qi2019KnowledgeDrivenSO},~\cite{Zhu2021MultiagentDR}, (c) offloading divided sub-tasks to multiple MEC servers~\cite{Cheng2020JointTO}, (d) offloading specified size of tasks to multiple MEC servers~\cite{Lu2020OptimizationOT}.}
\label{offload}
\end{figure}

In reality, the user's task can usually be partitioned into serval sub-tasks to offload to the MEC servers instead of offloading a single task like~\cite{Heydari2019DynamicTO}. The joint optimization of task partition-enabled task offloading, transmit power control, and communication and computational resource allocation in an MEC-enabled Ultra-Dense Network (UDN) is investigated in~\cite{Cheng2020JointTO}. The system model consists of multiple randomly distributed MEC-enabled APs with various computation and transmission resources and a number of users with limited battery capacity and computational resources. Each user can divide its task into multiple sub-tasks, and then distributes them to the MEC-enabled APs, as shown in Fig. \ref{offload}(c). The AP allocates the communication and computational resources to the users in proportion to their sub-task workloads. The problem of each user is to determine the task partitions so as to maximize the utility of each user, which is a function of each user's task size, energy consumption, and task execution delay. \textcolor{black}{To solve the joint task partition and transmit power control issue, MADDPG~\cite{lowe2017multi} is adopted to learn an optimal task partition policy for each user. The details of MADDPG can be found in Section \ref{MARLalgorithms}.} The state of each user is represented as the available communication and computational resources in the network as well as its local status including the local computational resources, the battery level, and the task requirements. The action of each user includes its task partitions and the transmit power selection. After performing an action, each user receives a reward that is equal to the value of utility. The simulation results show that the proposed MADDPG scheme outperforms the DDPG scheme in terms of average energy consumption, total utility, and average task delay. Given the advantages of MADDPG, this algorithm is also used for the task offloading decisions in vehicular networks~\cite{Zhu2021MultiagentDR} and IoT systems~\cite{Cao2020MultiagentDR}. However, the centralized training of MADDPG may lead to the scalability issue especially in UDNs as we discussed in section \ref{Challenges}, and generate a large amount of signaling overhead for information sharing.

To encourage the agents to explore the environments, the authors in~\cite{Lu2020OptimizationOT} combine MADDPG with SAC to solve the task partition-based task offloading in an MEC network. The system model involves multiple mobile users (MUs) and MEC-enabled hybrid access points (HAPs). At the beginning of each time slot, the MU selects one HAP to offload a specified amount of data, as shown in Fig. \ref{offload}(d), which contains two different actions: the discrete HAP selection and the continuous data size for offloading. The objective of each MU is to reduce the execution delay, the task failure rate, and the energy consumption of the whole system. The optimization problem is modeled as an MDP in which each MU as an agent can observe the environment state that includes (i) the transmission rates between the MU and HAPs, (ii) the CPU utilization of HAPs, (iii) the MU's remaining data to be processed, (iv) the remaining power of the MU, and (v) the power generated by the MU. The common reward for all the MUs is a function of the total delay and total energy consumption of all MUs, as well as the total cost of all HAPs. In order to enlarge the exploration of agents, the MADDPG leverages the maximum entropy term of SAC to explore optimal solutions, thereby enhancing the stability of the algorithm. The simulation results show that compared with DDPG, MADDPG, and SAC, the combination of SAC and MADDPG achieves significant performance gain in low task latency, task drop rate, and energy consumption.

The service task offloading in heterogeneous vehicular networks is investigated in~\cite{Qi2019KnowledgeDrivenSO}, where the MEC servers are deployed at different infrastructures, such as BSs, APs, or resource-rich vehicles. Different MEC servers can provide different capabilities and are located in different regions to satisfy the requirements of vehicles. Each vehicle has different services, and each service has various numbers of tasks. Each service's problem is to decide whether to offload its task to an MEC server or execute it locally so as to minimize the task execution delay. To solve the problem, the authors propose a knowledge-driven service offloading decision framework that keeps one decision model for each agent, i.e., a service. The A$3$C algorithm~\cite{Mnih2016AsynchronousMF} is adopted for each service and pre-trained at a powerful MEC server through utilizing the generated service data. When a vehicle starts a service, the corresponding decision model is fetched. Then, the vehicle can learn the distributed decision model online using its local state. The local state of each service consists of its task profile, the current state of MEC servers, and the velocity of the vehicle. The vehicles running the same service can update the corresponding A$3$C model asynchronously, and the new model is periodically updated to the MEC server. The reward for each vehicle is determined by its task execution delay. The simulation results show that the proposed algorithm can reduce task execution delay compared with the greedy algorithm. However, the proposed algorithm has a similar average delay to the greedy algorithm when the number of tasks exceeds $25$, since the limited computational resources in the network may not be enough to meet the need for increased tasks. Besides, the asynchronous update of parameters may occur the high signaling overhead, and degrade the service's performance since the parameters of the resource-constraint training vehicles may be outdated~\cite{Mnih2016AsynchronousMF}.

Cloud centers (CCs) can also offload their tasks to the edge devices (EDs) and MEC servers deployed at APs to reduce the overall computing latency. The joint optimization of task offloading and resource allocation of CCs in a multi-cloud Het-MEC network is investigated in~\cite{Zhang2019JointDO}. The system model consists of multiple CCs, APs, and EDs. Each AP is connected to multiple CCs via wired communication, while each ED connects with only one AP through wireless links. The task is published at the CC and can be divided into multiple sub-tasks to allocate to different nodes, i.e., the CC, the connected AP, and the ED. For each task, the ED is responsible for generating the data of the task, processing its part, and transmitting the processed data and the remaining raw data to the connected AP. The AP and the CC process the remaining part and integrate the results at the CC. The problem of each CC is to allocate the computing and transmitting capacity, and the raw data volume to the connected EDs, APs, and itself so as to minimize the overall system latency. The offloading policy of the CC is modeled as an MG, where each CC is considered as an agent and takes actions independently. An MARL-based joint action learners $Q$-learning~\cite{claus1998dynamics},~\cite{1997Adversarial} algorithm is then adopted to solve the MG. The environment state is observable to all the CCs, and it consists of the data generation speed of all EDs for each task. The actions of each CC include the computing and transmitting capacity, and the raw data volume allocation. The reward is defined as the negative value of the overall system latency. To depict the competition among CCs, each CC makes decisions based on the learned explicit models of other CCs and the learned joint $Q$-values from the standard TD method. The simulation results show that the proposed MARL algorithm can achieve better system latency of each CC and of the overall system compared with the local computing and the cloud computing cases. 

In a dynamic MEC network, the MUs can travel across multiple MEC servers. Considering the QoS requirements of the MUs, the authors in~\cite{Liu2021DistributedTM} address the task migration issue caused by the mobility of MUs. The system model consists of multiple MEC servers and multiple MUs. Each MU has one corresponding task executed on the MEC servers. At each time slot, each MU's task needs to decide whether to migrate to other MEC servers so as to minimize the average completion time of all the tasks. This collaboration problem is formulated as an MMDP. Here, the MU's task is the agent. The state of each MU's task includes the task's remaining computation requirement, the task's serving MEC server in the previous time slot, and the connected MEC server of the task’s corresponding MU. The global reward for all the tasks is the difference in the average estimated completion time of all tasks between the previous time slot and the current time slot. To maximize the global reward, COMA is adopted to learn an optimal policy for each task through using a centralized critic, which can calculate a counterfactual baseline for the actor of each agent to solve the credit assignment issue, thereby improving the performance of agents. The actor of each agent only leverages its state to make migration decisions. However, the centralized critic needs the states and actions of all tasks as the input to calculate baselines, which may lead to the scalability issue and is not suitable for large-scale networks. The simulation results show that the proposed COMA-based decentralized task migration algorithm can obtain at most $50$\% reduction in average completion time compared with the Not Migrating strategy and the single-agent actor-critic algorithm.

\subsection{Content Caching}
\begin{figure}[h]
  \centering
  \includegraphics[scale=0.4]{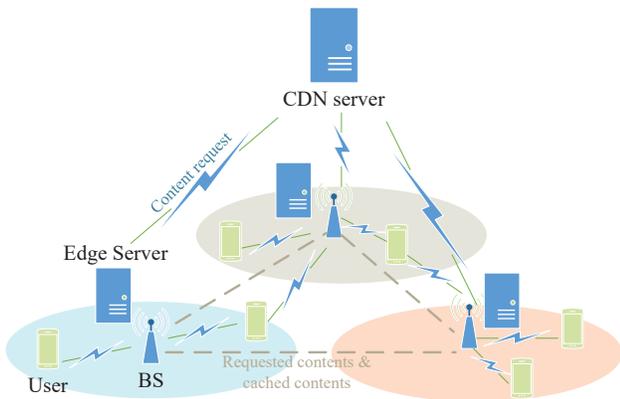}
  \caption{Cooperative edge caching with multiple MEC-enabled BSs.}
  \label{caching}
\end{figure}

\emph{\textbf{1) Cooperative edge caching: }}Cooperative edge caching enables multiple edge servers to cooperatively cache contents to serve the nearby users, as shown in Fig. \ref{caching}. This means that the serving users can request the content not only from the local edge servers but also from neighboring edge servers.

The cooperative edge caching for mobile video steaming is investigated in~\cite{Wang2020IntelligentVC}. The system model consists of a remote Content Delivery Network (CDN) server with all the requested video contents, multiple BSs equipped with edge servers, and users. The BSs are connected to the remote CDN server through the backbone network. Each BS has limited storage capacity for content caching to serve the local video requests from the users within its coverage range. To solve the problem, cooperative edge caching enables the BS to fetch the requested content from neighboring BSs if its local cache does not cache the content, which is more faster and cost-effective than fetching them from the CDN. Thus, the user's requested contents can be fetched from the local BS, the neighboring BSs, or the remote CDN. The BS's problem is to select a set of contents to cache so as to minimize the communication cost and the content access latency of users. 

To reveal the potentials and the challenges of applying cooperative edge caching, the authors first analyze a real-world video watching dataset. The analysis results demonstrate that neighboring areas have a high content similarity that is extremely beneficial to cooperative caching. However, the content similarity is also time-varying and heterogeneous across multiple edges. Thus, to learn the dynamics of the content similarity, multi-agent advantage actor-critic (MAA2C) is adopted to build an adaptive caching policy. The state of each BS is the status of the current requested contents and the cached contents. The reward for each BS is defined as the negative value of the weighted sum of the transmission latency and traffic cost. The input of MAA2C refers to the current BS’s and its neighboring BSs' states, as well as the fingerprints of neighboring BSs, i.e., the probability simplex of neighboring BSs, to learn the changes of other BSs' policies. The real trace-driven evaluation results show that the proposed MAA2C reduces the latency around $21$\% and the cost around $26$\% compared with the state-of-art caching methods such as DRL~\cite{zhu2018caching}, joint action learners~\cite{Jiang2019MultiAgentRL}. The actor-critic framework for cooperative edge caching can also be found in~\cite{Zhong2019DeepMR}. However, unlike the unique model for each BS used in~\cite{Wang2020IntelligentVC}, the agent model of each BS in~\cite{Zhong2019DeepMR} contains a unique actor and a shared critic. The shared critic network computes a TD error to update the actor by using all the BSs’ observations, which may increase the computational complexity at the training unit and has poor scalability.

Unlike~\cite{Wang2020IntelligentVC} and~\cite{Zhong2019DeepMR}, the authors in~\cite{Xu2018CollaborativeMR} consider adopting a collaborative multi-agent multi-armed bandit (MAMAB) to address the cooperative caching problem. The system model consists of multiple small base stations (SBSs) and multiple users. The SBSs can serve the users within their communication ranges. The users can request contents from the neighboring SBSs, and they will be served by the nearest SBSs that cached the requested contents. If the users do not fetch the content from the neighboring SBSs, it will fetch it from the core network directly. Thus, the problem of each SBSs is to learn an optimal caching policy so as to minimize the expected transmission delay of the network, which is modeled in an MAMAB framework with stateless setting~\cite{kapetanakis2002reinforcement}. In this case, the SBS is an agent.

To achieve this global goal, the collaborative MAMAB is proposed for SBSs to learn the optimal caching policies. To this end, the authors construct a coordination graph~\cite{kok2006collaborative}, in which each SBS represents a vertex, and if two SBSs cover the same users there will be an edge to connect them. Besides, each vertex has an edge to connect itself. A reward assignment scheme is proposed to help the SBSs learn to collaborate with each other by assigning the reward of the SBS to its edges. Specifically, if a user is served by a neighboring SBS which is the nearest SBS of the user, the reward will be totally assigned to the SBS's edge that connects itself. If the serving SBS is not the nearest, then the reward will be divided proportionally to the edges between the serving SBS and all nearer neighboring SBSs of the user. Note that the reward on each edge only relates to the caching actions of the connected two SBSs which reduces the joint action space of the reward function and the computation complexity. The optimal caching policies can be learned by maximizing the total reward using coordinate ascent algorithm~\cite{vlassis2004anytime} in which each SBS optimizes its own action in sequence while keeping the action of other SBSs fixed. The simulation results show that the proposed collaborative MAMAB can achieve a significant gain in low average delay compared with the proposed fully centralized and decentralized baselines as well as the greedy algorithm, the Least Recently Used (LRU)~\cite{abrams1996removal} caching algorithm, and the Least Frequently Used (LFU)~\cite{arlitt2000evaluating} caching algorithm.

The content caching can be implemented in D2D networks as investigated in~\cite{Jiang2019MultiAgentRL}. In the network, the UEs are connected to a BS, and each UE has a local cache with limited storage capacity. The cached content in each UE can be shared with others via D2D links. Each UE can request content from the local cache, other UEs, or the BS. In this case, if the BS cache does not hit, the UE retrieves the requested content from the core network. Each UE tries to learn an optimal cache policy that can minimize the total latency of all the UEs. Similar to~\cite{Xu2018CollaborativeMR}, the D2D caching problem is modeled as an MAMAB in which each UE is an agent, and the content items are the arms. However, different from the coordination graph solution used in~\cite{Xu2018CollaborativeMR}, joint actions learners-based modified combinatorial upper confidence bound (JAL MCUCB) method is proposed to select an optimal super action for each UE. In particular, the super action of each UE is a vector that represents its cache actions on the whole content items, whose elements are underlying actions. The reward for each UE is defined as the reduction of downloading latency.

To enable the neighboring UEs for cooperative caching, the joint super actions of neighboring UEs are taken into consideration as the state of the current UE to calculate the expected rewards to update $Q$-values. However, this may incur huge action space. To address the problem, the JAL MCUCB adopts the idea of CUCB~\cite{chen2013combinatorial} algorithm that uses the joint underlying actions of neighboring UEs to calculate the expected rewards, thus, the joint action space can be reduced to the number of underlying actions. Moreover, the JAL MCUCB enhances the CUCB with the Zipf-like distribution~\cite{chen2013combinatorial} of content popularity to update $Q$-values, which can improve the exploration and exploitation. The UE updates the $Q$-values of its joint underlying actions according to the JAL MCUCB algorithm, and input the updated $Q$-values into the $(\alpha, \beta)$-approximation oracle~\cite{chen2017caching} to find the optimal super action for its caching decision. The simulation results show that compared with LRU~\cite{abrams1996removal}, LFU~\cite{arlitt2000evaluating}, and Informed Upper Bound (IUB)~\cite{blasco2014learning}, the proposed JAL-based CUCB can achieve gains equal to $8$\%, $11$\%, and $19$\%, respectively, in terms of transmission latency. Besides, the proposed JAL-based CUCB can increase the cache hit rate of the IL-based CUCB and LRU by $40$\% and $25$\%, respectively. The same D2D cooperative edge caching problem is also investigated in~\cite{Yan2020CooperativeCA}, where the problem is formulated as a Dec-POMDP to minimize the total content fetching latency. The authors extend the soft actor-critic (SAC)~\cite{Haarnoja2018SoftAO} to a multi-agent environment as multi-agent SAC (MASAC) to solve the problem. The MASAC is further enhanced by utilizing the GAT~\cite{velivckovic2017graph}-based self-attention. It provides complex coordination among agents through characterizing the game state as a graph and computing the node-embedding feature of each agent by attending its neighboring nodes. Then, the node-embedding feature of each agent is used as the input of its MASAC model to compute the action and the corresponding $Q$-value.

\emph{\textbf{2) Cooperative coded caching: }}Considering the mobility of users, the user may not be able to completely download a whole content from a single small base station (SBS). To address the problem, a caching technique, namely cooperative coded caching, has been proposed. The cooperative coded caching enables the content to be divided into many segments by leveraging a coded caching system~\cite{Shokrollahi2007RaptorC}, and the SBSs can cooperatively cache these content segments so that the users distributed across the SBSs can flexibly request these segments. 

The cooperative coded caching problem in ultra-dense cellular networks is investigated in~\cite{Chen2021MultiAgentDR}. The system model consists of multiple SBSs, multiple users, and a content server. The SBSs can provide a high channel capacity within a small coverage range. All the SBSs are connected to the content server through backhaul links. The user can either fetch the content from the SBSs or fetch it from the remote content server. The problem of each SBS is to decide how many segments to cache for each content so as to minimize the content fetching cost. The cooperative multi-agent actor-critic framework is then adopted to develop a cooperative caching algorithm. The state of each SBS is defined as the current local cache status, the user request status, and the user access status. In particular, the current request status and access status are estimated by an LSTM network through inputting the historical state sequences of them. The reward for each SBS is defined as the cost savings obtained by its cache action. The actor network makes content caching decisions depending on its local state and a shared global state of the previous time slot output by a communication module for cooperation. In particular, the shared global state represents the time-dependent variability of the actions and observations of all the agents. This can reduce the interaction data overhead compared with sharing directly the actions and observations of all the agents. The centralized critic network is deployed at the central core network to estimate an action-value function to train an actor for each SBS network through using all the SBSs’ experiences. The evaluation experiments conducted within an ultra-dense network with a large number of BSs reveal that the proposed algorithm can achieve a better fetching cost and cache hit rate compared with non-cooperative multi-agent actor-critic~\cite{zhong2020deep}, (LRU)~\cite{abrams1996removal}, and DRL~\cite{kapetanakis2002reinforcement}.

Unlike~\cite{Chen2021MultiAgentDR}, the authors in~\cite{Pedersen2021DynamicCC} propose to use an MARL-based DDPG to solve the cooperative coded caching in small cell networks. The system model is similar to that in~\cite{Chen2021MultiAgentDR}. The file to be cached can be divided into a number of packets and encoded through a maximum distance separable (MDS) code to improve the caching efficiency. Different from~\cite{Chen2021MultiAgentDR}, the authors apply the soft time-to-live (STTL)~\cite{goseling2018soft} caching policy to manage the cached encoded packets of the requested file, and the packets can be moved out of the SBS at periodic times after the most recent request for the file. Thus, each SBS learns an STTL policy so as to minimize the network loads. The problem is formulated as an MG. The agent is the SBS. To solve the MG, an MARL-based DDPG is adopted to learn the optimal caching policy for each SBS. The state of each SBS is represented as the file requested at the current time-step, the caching status of each file at the SBSs, the average cache usage of each file, and the amount of file requested in the previous time slot cached by the other SBSs. The reward for each SBS is defined as the amount of data downloaded from its cache. Under the heterogeneous request process, the simulation results show that the MARL-based algorithm can achieve a lower network load than the proposed single-agent RL baseline algorithm. The cooperative coded caching using MDS code in small cell networks is also investigated in~\cite{Wu2020MultiAgentRL}. Instead of using the STTL caching policy like~\cite{Pedersen2021DynamicCC}, the SBSs need to decide the proportion of MDS encoding bits of each content to cache so as to minimize the fronthaul traffic loads for acquiring contents from the content server. Given the DDPG's unreasonable mapping between a proto-action and a feasible one, the proposed HDDPG enhanced DDPG by introducing a reasonable cumulative penalty term in the objective function of RL. Here, the penalty term is determined by the remaining storage of all caching units after taking action. The simulation results show that the proposed HDDPG outperforms plain DDPG in terms of fronthaul traffic loads.

\textbf{Summary: }This section reviews the application of MARL for computation offloading and content caching. From Table \ref{tab:offloading}, we may observe that CTDE is popular for training edge users for computation offloading, which enables the edge user to learn a decentralized policy that only uses its observation, e.g., the task queue length, to make offloading decisions. On the contrary, for content caching, the edge servers, e.g., BSs, may need to exchange content caching status or caching actions with neighboring edge servers to implement the cooperative edge caching, and \textcolor{black}{thus decentralized learning with networked agents is commonly used to learn an optimal decentralized policy for the edge server. How the decentralized learning with networked agents scheme works can be found in Section \ref{Challenges}.} The decentralized offloading or caching policy can reduce the computational complexity for action decision-making compared with the centralized policy using all the agent's observations. Actually, the offloading and caching usually involve data transmission, thus, the spectrum access or power control needs to be jointly considered with the offloading and caching, which deserves further investigation.

%table VI

\begin{table*}[]\scriptsize\centering
\renewcommand\arraystretch{1.5}
\caption{A summary of approaches applying MARL for computation offloading and content caching.}
\label{tab:offloading}
\begin{tabular}{|m{0.5cm}<{\centering}|m{0.6cm}<{\centering}|m{3cm}<{\centering}|m{1.5cm}<{\centering}|m{1.5cm}<{\centering}|m{1.5cm}<{\centering}|m{1cm}<{\centering}|m{1.5cm}<{\centering}|m{2cm}<{\centering}|m{1cm}<{\centering}|}
\rowcolor[HTML]{9B9B9B}
\textbf{Issues}            & \textbf{Works}      & \textbf{Main Findings}                    & \textbf{Model}     & \textbf{Learning Schemes}      & \textbf{Algorithms}                & \textbf{Private or Shared Policy} & \textbf{Cooperation or Competition}  & \textbf{Learning Objectives}   & \textbf{Network Models}                \\ \hline
\multirow{8}{*}{{\rotatebox{90}{Computation Offloading\quad \quad \quad \quad \quad \quad \quad \quad \quad \quad \quad \quad \quad}}} & ~\cite{Heydari2019DynamicTO}   & \textcolor{black}{Investigates the computation offloading problem in a non-cooperative MEC system where each MD learns its policy independently} & POMDP     & IL & A2C                      & Private                  & Competition    &   MD's expected cost minimization     & MEC system                    \\ \cline{2-10}
                                       & ~\cite{Cheng2020JointTO}   & \textcolor{black}{Solves joint task partition and power control problem with MARL for an MEC enabled-UDN}    & POMDP     & CTDE                 & MADDPG                    & Private                  & Competition           &User's utility maximization  & UDN               \\ \cline{2-10}
                                       & ~\cite{Lu2020OptimizationOT} & \textcolor{black}{Considers the joint task partition and computation offloading in an MEC network where each MD can split its tasks and offload them to multiple HAPs}  & POMDP       & CTDE                 & MADDPG combined with SAC         & Private                  & Competition      & MU's task execution delay, task failure rate, and energy consumption minimization         & MEC system                    \\ \cline{2-10}
                                       & ~\cite{Zhu2021MultiagentDR} & \textcolor{black}{Studies the computation offloading using MARL in vehicular networks}    & Dec-POMDP     & CTDE                 & MADDPG                   & Private                  & Cooperation           & Total task processing delay minimization     & Vehicular network \\ \cline{2-10}
                                       & ~\cite{Qi2019KnowledgeDrivenSO} & \textcolor{black}{Solves the service task offloading problem in heterogeneous vehicular networks}  & POMDP       & CTDE                 & A3C                      & Private                  & Competition           & Task execution delay minimization    & Vehicular network \\ \cline{2-10}
                                       
                                       & ~\cite{Cao2020MultiagentDR}   & \textcolor{black}{Propose a cooperative MARL algorithm for task offloading in IIoTs}  & Dec-POMDP     & CTDE                 & MADDPG                   & Private                  & Cooperation               &Network delay minimization & IIoT              \\ \cline{2-10}
                                       & ~\cite{Zhang2019JointDO}   & \textcolor{black}{Jointly optimizes the resource allocation and task offloading for CCs via MARL}      & MG       & IL & $Q$-learning               & Private                  & Cooperation              &Overall system latency minimization & Multi-cloud Het-MEC network   \\ \cline{2-10}
                                       & ~\cite{Liu2021DistributedTM}  & \textcolor{black}{Solves the task migration problem with MARL in a dynamic MEC network}   & MMDP       & CTDE                 & COMA                     & Private                  & Cooperation             & Task average completion time minimization  & MEC system                    \\ \hline
\multirow{7}{*}{{\rotatebox{90}{Content Caching\quad \quad \quad \quad \quad \quad \quad \quad \quad \quad \quad \quad}}}       & ~\cite{Wang2020IntelligentVC} & \textcolor{black}{Studies the cooperative edge caching for mobile video steaming}   & Dec-POMDP     & Decentralized with Networked Agents                 & MAA2C                    & Private                  & Cooperation            & Communication cost and content access latency minimization   & Cellular system   \\ \cline{2-10}
                                       & ~\cite{Zhong2019DeepMR}  &  \textcolor{black}{Solves the cooperative edge caching in cellular systems where each BS has a unique actor and a shared critic}     & Dec-POMDP     & IL & Multi-agent actor-critic & Private                  & Cooperation               & Network transmission delay minimization & Cellular system   \\ \cline{2-10}
                                       & ~\cite{Jiang2019MultiAgentRL}  &  \textcolor{black}{Formulates the content caching problem as an MAMAB for D2D-enabled cellular systems}  & MAMAB      & Decentralized with Networked Agents & JAL MCUCB               & Private                  & Cooperation                & Total content fetching latency minimization & Cellular system   \\ \cline{2-10}
                                       & ~\cite{Yan2020CooperativeCA} & \textcolor{black}{Investigates the cooperative edge caching problem in D2D-enabled cellular systems}   & Dec-POMDP & Decentralized with Networked Agents                 & MASAC                    & Private                  & Cooperation               & Total content fetching latency minimization & Cellular system   \\ \cline{2-10}
                                       & ~\cite{Xu2018CollaborativeMR} & \textcolor{black}{Addresses the cooperative edge caching problem with a collaborative MAMAB for small cell networks}  & Cooperative MAMAB       & Decentralized with Networked Agents                 & MAMAB                    & Private                  & Cooperation               &Network transmission delay minimization & Cellular system   \\ \cline{2-10}
                                        & ~\cite{Chen2021MultiAgentDR} &  \textcolor{black}{Proposes a cooperative MARL algorithm for cooperative coded caching in ultra-dense cellular networks}   & Dec-POMDP     & CTDE                 & Cooperative multi-agent actor-critic & Private                  & Cooperation                &Content fetching cost minimization& UDN               \\ \cline{2-10}
                                       & ~\cite{Pedersen2021DynamicCC} & \textcolor{black}{Investigates the cooperative coded caching problem in small cell networks and applies an STTL to manage the cached encoded packets}   & MG     & IL                 & DDPG                     & Private                  & Cooperation             &Network loads minimization  & Small cell network           \\ \hline
\end{tabular}
\end{table*}

\section{Packet Routing}\label{packet routing}
Packet routing is one of the challenging issues in decentralized and autonomous networks. Given the dynamics of network environments, each router should learn an adaptive routing policy that can dynamically change its routing rules to reduce network congestion. However, conventional packet routing protocols such as Border Gateway Protocol (BGP) are almost rule-based and cannot make performance-based routing decisions, i.e., they cannot change the routing rules adaptively according to the network performance such as throughput and latency. To address the problem, the standard DRL approaches, e.g., ~\cite{Stampa2017ADL},~\cite{Wang2017AutonomousNO},~\cite{Challita2018DeepRL}, are recently used to learn the performance-based routing methods. \textcolor{black}{However, such an approach typically treats each router as an independent agent without considering the impacts of other routers' routing policies. As explained in Section \ref{Challenges}, this may result in the non-stationarity issue.} MARL allows the routers to cooperate with each other, and thus the global optimal routing policy can be obtained to balance the traffic flow distribution and improve the performance of the whole network system. As a result, MARL has been recently used as an effective solution for the packet routing.

The first work can be found in~\cite{Zhao2020ImprovingIR} that proposes an MARL framework for a decentralized performance-based inter-domain routing. The system model consists of multiple Autonomous Systems (ASs) that interact with each other for traffic management and business negotiation. The ASs' business relationships are modeled as customer-provider and peer-peer. The routing policy of each AS follows the rules of business relationship: (i) the customer AS as the next-hop has a higher priority than the peer AS; (ii) the peer AS as the next-hop has a higher priority than the provider AS. Each AS aims to learn to select the best next-hop to maximize its own average throughput.

For that, each AS as an agent observes its state that includes the source and destination of the flow to be routed with the current inference, the flows that the AS needs to transfer, the traffic loads of the links connecting to neighboring ASs, and the observed throughput of routed flows from the last inference. The AS can further learn flow status from neighbors within different ranges, e.g., the neighbors within $n$-hop range, so as to explore the information sharing ranges' influences on learning effectiveness. The AS's action is to select one of the neighboring ASs as the next-hop for the flow to be routed. The reward for each AS is defined as its average throughput of by-passing flows. The routing policy of each AS is trained locally by adopting the A2C algorithm, which can efficiently solve the credit assignment issue and improve the AS's performance. The simulation results show that the proposed algorithm can improve the overall throughput by $20$\% compared with the BGP routing. Moreover, the performance increases as the scope of information-sharing increases.

 \begin{figure}[h]
  \centering
  \includegraphics[scale=0.35]{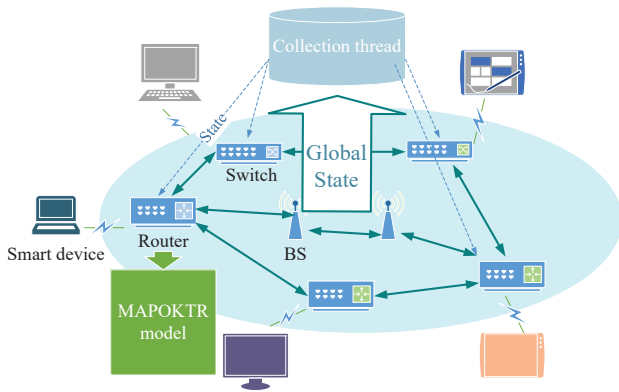}
  \caption{MAPOKTR model for packet routing with global state information.}
  \label{MAPOKTR}
\end{figure}

Edge networks can provide computation offloading and content caching services for edge users' computation-intensive and ultra-low-latency applications. However, this may occur heavy traffic loads to edge networks, thus an efficient packet routing method for edge networks is investigated as in~\cite{He2020DeepHopOE}. The edge network model consists of multiple infrastructures such as BSs distributed in the center of a circle, multiple edge nodes such as routers and switches distributed on the edge of the circle, and multiple smart devices. The smart device in the edge network can only exchange traffic data with its connected edge nodes. After receiving a packet from a smart device or other edge nodes, the edge node aims to select a next-hop node for the packet forwarding so as to minimize the average slowdown and the number of dropped packets in edge networks.

The sequential decision-making of multiple edge nodes is modeled as an MG since each edge node as an agent can access the global network information that includes all the edge nodes' processing capability and the bandwidths of links, which can be obtained from a collection thread as shown in Fig. \ref{MAPOKTR}. Besides, each edge node can also observe the target node of the current packet to be routed and the packet's Maximum Transmission Time (MTT) and Transmission Priority (TP). A decentralized MADRL method, namely Multi-Agent Policy Optimization using Kronecker-Factored Trust Region (MAPOKTR), is proposed to learn an optimal routing policy for each agent. The MAPOKTR is actually an extension of the Actor-Critic using the Kronecker-Factored Trust Region (ACKTR) algorithm~\cite{wu2017scalable} to multi-agent environments. To deal with the complicated state information of each agent, the MAPOKTR uses an attention mechanism~\cite{vaswani2017attention} that learns the information of interest such as the destination of the packet from the huge state information by giving such interesting information more attention weights. The simulation results show that the proposed MAPOKTR can achieve a better network throughput with the huge state information compared with the state-of-the-art MADRL algorithms. Note that the collection thread is actually challenging to be deployed in realistic networks due to the decentralized and dynamic nature of modern networks.

Given the challenge of adopting the collection thread to collect instantaneous global network information, the authors in~\cite{You2020TowardPR} adopt a fully decentralized MARL method for packet routing. The network model with multiple routers is formulated as a directed graph, where the node is the router and the edge is the transmission link. Each node's problem is to select its adjacent nodes as the next-hops to forward the received packets so as to minimize average delivery time. To this end, a deep $Q$-routing algorithm that is the combination of the $Q$-learning based $Q$-routing~\cite{Boyan1993PacketRI} and MADRL is proposed to learn a fully decentralized routing policy for each router node, i.e., the agent. Deep $Q$-routing improves the $Q$-routing by using a DNN to replace the Q-table used in the $Q$-routing. The input of the DNN refers to a one-hot encoding of the current packet’s destination, and the output is the estimated $Q$-values of the routing decisions. Given the non-stationarity multi-agent environments, the expected delivery time, i.e., the estimated $Q$-value, for a packet from the selected next node to the destination will be recalculated before the training process. To reduce the computational complexity of calculating the expected delivery time, a tag that denotes whether the next-hop is the destination is added to the replay buffer tuple. In particular, if the tag is set to $1$, the delivery time is directly set to $0$. Similar to~\cite{Zhao2020ImprovingIR}, to improve the learning efficiency, the deep $Q$-routing algorithm can use extra state information including the historical action decisions of the past packets and the destinations of packets to be routed as the input of the DNN. The simulation results show that the deep $Q$-routing using extra information can yield lower delivery delays compared with the deep $Q$-routing that only uses its local state. 

The deep $Q$-routing algorithm proposed in~\cite{You2020TowardPR} uses the regular TD prediction to calculate the expected long-term return, which may lead to high bias since the TD prediction only considers the impact of the next-hop on the expected return. To overcome this issue, the authors in~\cite{Pinyoanuntapong2019DelayOptimalTE} propose a spatial difference (SD) prediction approach for the packet routing problem. The SD prediction approach leverages the number of hops from the current router node as the standard to outline different $Q$-value estimation methods, e.g., $1$-hop or $n$-hop action-value estimation. The $1$-hop action-value estimation uses the $1$-hop delay, i.e., the packet arriving time, and the $Q$-value estimation of the next-hop to update $Q$-values based on the exponential weighted average rule, while the $n$-hop action-value estimation uses the corresponding $n$-hop delay and the corresponding $Q$-value estimation. Thus, the $n$-hop method can lead to better policy due to the smaller bias in the $Q$-value estimates compared with the $1$-hop method. However, the stochastic environment may lead to a high variance in $Q$-value estimates. To address this issue, an expected $n$-hop action value estimation is employed that leverages the expected $Q$-value of the next-hop to update the current $Q$-values.

Due to the dynamic network topology caused by the moving nodes in highly dynamic Mobile Ad-hoc Networks (MANETs), conventional $Q$-routing~\cite{Boyan1993PacketRI} used in~\cite{You2020TowardPR} is inefficient in such dynamic environments since the $Q$-values quickly become stale as links break. Thus, the design of packet routing with dynamic nodes becomes more challenging than that with stationarity nodes, which is investigated in~\cite{Kaviani2021RobustAS}. The authors aim to combine MADRL with the CQ+ routing~\cite{Johnston2018ARL} algorithm to design the routing protocol so as to minimize the normalized overhead under the goodput rate constraint. With the CQ+ routing, each node has a \emph{H}-factor that represents an estimation of the least number of hops from the current node to the destination of the packet through one potential next-hop, and a corresponding confidence level that represents the probability of the packet reaching the destination. The CQ+ routing then chooses to unicast or broadcast its packets depending on the best path confidence level. However, it may fall into a local optimal solution due to its limited perspective of the overall network. Furthermore, CQ+ routing ignores the change rate of \emph{H}-factors and the corresponding confidence levels.

To solve the problem, the DEEPCQ+ is proposed that improves the CQ+ routing in routing policy by adopting the PPO algorithm to train a shared policy for all the nodes, i.e., agents, since PPO inherits TRPO's stability and reliability but works in a simpler way. In order to make the policy scalable in various network sizes and dynamic network topologies, a preprocessing program is implemented to select a fixed number of top neighbors with smaller products of the number of hops and the uncertainty. The agent can then use the network parameters (\emph{H}-factors and corresponding confidence levels) of these neighbors as well as these parameters' changes compared to the previous ones as the observations. Besides, the action taken by the agent in the previous time slot is also included in its observation. Based on its observation, the node can select the best neighbor to forward its packet if the selected neighbor has low uncertainty. Then, the node will receive a reward that is equal to the success probability of the path via the selected next-hop. Otherwise, if the selected neighbor has high uncertainty, the node will broadcast its packet to explore the network so as to achieve a more reliable transmission. Then, the reward for each agent is defined as the uncertainty of the path via the selected next-hop. The simulation results show that the proposed DEEPCQ+ has a lower transmission overhead than that of the conventional CQ+ routing. Moreover, the policy sharing mechanism enables the nodes to reuse the trained policy in different network dynamics and topologies.

Apart from the MANETs, packet routing is an emerging issue in wireless sensor networks (WSNs). In WSNs, due to the limited energy recourses of sensor nodes, an efficient routing policy is required to reduce the energy consumption of sensor nodes and prolong the lifetime of sensor nodes. To this end, the authors in~\cite{Li2020RoutingPD} propose to adopt the MARL framework to design the routing policy for sensor nodes in the underwater optical WSNs. The system model consists of multiple sensor nodes and a sink node. The data collected by the sensor nodes will be sent to the sink by one or multiple hops. The sensor node's problem is to select the next-hop from the neighboring nodes so as to maximize the network life cycle. The problem is modeled as an MG in which the agent is the sensor node. Two reward schemes are proposed to evaluate the agent’s action under different perspectives: the local reward and the global reward (GR). The local reward that is based on the residual energy and the link quality is used to evaluate the current agent’s action from a local perspective, while the global reward (GR) defined by the transmitting direction (i.e., an indicator that represents whether the packet forwarded by the action is farther from or closer to the sink node compared with the current location), evaluates the agent's action from a global perspective. A distributed value function (DVF)~\cite{liang2008multi} is adopted by each agent to learn the routing policy by iteratively updating the $Q$-values, and the GR is introduced in the DVF as the extra reward for the agent to evaluate how good the action is from a global perspective. To deal with the non-stationarity issue and balance the energy distribution of WSNs, the $Q$-values of the current agent’s neighbors are also taken into account in the DVF to update the current agent's $Q$-values so as to make stable routing decisions, and the achieved GR can be allocated to neighbors to award their information gain. The simulations results show that the proposed routing protocol can obtain a significant performance gain in terms of packet delivery ratio, energy consumption, and network load balance compared with Double $Q$-learning and AODV.

The DVF method for packet routing can also be found in~\cite{Li2020AMM}. The authors propose a Modified-QLAODV method based on the DVF to deal with the packet routing in dynamic Unmanned Aeronautical Ad hoc Networks (UAANETs). The conventional QLAODV~\cite{wu2010distributed} method can only obtain two-hop neighbors' information through exchanging Hello. Thus, it is not suitable for solving multi-hop routing. To solve the problem, the Modified-QLAODV uses the DVF to employ all the neighbors' $Q$-tables to update the current node's $Q$-value. Different from~\cite{Li2020RoutingPD}, considering the dynamic topology of UAANETs, a weight value for each neighboring node is used in the $Q$-value update process, which is decided by the link duration between the node and its neighboring node. In particular, the link duration is the duration obtained by the node after it receives Hello and extracts its neighbors' information. This enables the node to learn a reliable routing protocol. Moreover, to balance the energy distribution and prolong the lifetime of UAV nodes, the ratio of the average queue delay, i.e., the load, of the selected next-hop in the sum of average queue delay of all the neighbors are introduced in the reward function of the current node.

The large-scale networks can be divided into a few regions, and each region can make local routing decisions independently depending on the regional network status such as the communication link capacity utilization of the region. However, this makes it difficult for the regional network to obtain global objectives. To solve the problem, the authors in~\cite{Geng2020AMR} propose to adopt the MARL framework for packet routing in multi-region networks. The entire network is modeled as a directed graph, where the node is the router and the edge is the corresponding link capacity. The network traffic is quantized to a set of traffic demands, which represent the amount of traffic to be delivered from the source node to the destination nodes. The traffic demands enter a region through ingress nodes, i.e., the source nodes or the border nodes of the region, and leave the region via egress nodes, i.e., the destination nodes or the border nodes of the region. Thus, each region contains two kinds of traffic demands: terminal demands and outgoing demands. The terminal demands' the destination nodes are in the local region, while those of outgoing traffic are in the other regions. The objective is to optimize each region's local routings so as to minimize the maximum edge cost, i.e., the edge's congestion status, of the network. To achieve this goal, each kind of traffic demands learns a unique RL agent model for packet routing in each region. Specifically, the T-agent controls the terminal demands, while the O-agent controls the outgoing demands. 

A centralized server deployed at each region maintains the T-agent and the O-agent. The agents make actions decisions, i.e., the traffic splitting ratios over the constructed paths between each ingress-egress pair, depending on the collected local state information, i.e., the edge utilizations of the region. The T-agents receive rewards equal to the local objective, i.e., minimizing the local region's maximum edge cost, while O-agents receive rewards regarding the cooperative objective by combining the T-agents' reward values of the local and neighboring regions. Each agent is independently trained by the DDPG algorithm to learn an optimal routing policy. Numerical results show that the proposed MARL-based solution can achieve nearly-optimal policy and significantly reduce congestion.

\textbf{Summary: }This section reviews the applications of MARL for packet routing. These works are summarized in Table \ref{tab:traffic}. We can observe from the table that MARL can be used to solve the packet routing problems in various scenarios such as edge networks, UAANETs, MANETs, and WSNs. In the reviewed approaches, the network entities such as router nodes, IoT devices, and UAVs can learn their routing policies from the information of neighbors, i.e., next hops, which can improve the network performance such as average delivery time, end-to-end delay, and network life cycle. Some works focus on improving the existing routing policies such as CQ+ routing and QLAODV with MARL to make the routing policies more stable and scalable. Furthermore, we can note that the MARL-based packet routing methods can balance the energy distribution in UAV networks or WSNs, which can prolong the lifetime of such networks dramatically. Though MARL-based packet routing is promising for improving the network performance of future networks, designing efficient and practical packet routing methods still faces many challenges such as instantaneous state information collection and slow convergence of MARL learning models in dynamic network environments.

%table VII

\begin{table*}[]\scriptsize\centering
\renewcommand\arraystretch{1.5}
\caption{A summary of approaches applying MARL for packet routing.}
\label{tab:traffic}
\begin{tabular}{|m{0.5cm}<{\centering}|m{0.6cm}<{\centering}|m{3cm}<{\centering}|m{1.5cm}<{\centering}|m{1.5cm}<{\centering}|m{1.5cm}<{\centering}|m{1cm}<{\centering}|m{1.5cm}<{\centering}|m{2cm}<{\centering}|m{1.2cm}<{\centering}|}
\hline
\rowcolor[HTML]{9B9B9B}
\textbf{Issues}            & \textbf{Works}      & \textbf{Main Findings}                    & \textbf{Model}     & \textbf{Learning Schemes}      & \textbf{Algorithms}                & \textbf{Private or Shared Policy} & \textbf{Cooperation or Competition}  & \textbf{Learning Objectives}   & \textbf{Network Models}                \\ \hline
\multirow{8}{*}{{\rotatebox{90}{Packet routing\quad \quad \quad \quad \quad \quad \quad \quad \quad \quad  \quad \quad \quad \quad \quad}}}& ~\cite{Zhao2020ImprovingIR}       &  \textcolor{black}{Proposes a decentralized performance-based routing algorithm for multiple autonomous systems}       & POMDP     & Decentralized with Networked Agents                 & A2C             & Private & Competition & AS's average throughput maximization & Autonomous system             \\ \cline{2-10}
                                      & ~\cite{He2020DeepHopOE}   &  \textcolor{black}{Investigates the packet routing for edge networks and proposes a decentralized packet routing algorithm}            & MG       & CTDE                 & MAPOKTR            & Private & Cooperation & The average slowdown and the number of dropped packets minimization & Edge network                             \\ \cline{2-10}
                                      & ~\cite{You2020TowardPR}     &      \textcolor{black}{Applies a fully decentralized MARL algorithm to solve the packet routing problem in autonomous systems}        & Dec-POMDP     & CTDE                 & Deep $Q$-routing  & Shared  & Cooperation &Average delivery time minimization &Autonomous system \\ \cline{2-10}
                                      & ~\cite{Pinyoanuntapong2019DelayOptimalTE} & \textcolor{black}{Develops a spatial difference prediction method to address the packet routing problem} & POMDP     & IL & Actor-critic    & Private & Competition & End-to-end delay minimization   & Autonomous system \\ \cline{2-10}
                                      & ~\cite{Kaviani2021RobustAS}       & \textcolor{black}{Investigates the packet routing algorithm using MARL for Mobile Ad-hoc Networks with dynamic nodes}         & Dec-POMDP & Decentralized with Networked Agents                 & DEEPCQ+ routing & Shared  & Cooperation &  Normalized overhead minimization & MANET  \\ \cline{2-10}
                                      & ~\cite{Li2020RoutingPD}  & \textcolor{black}{Adopts a DVF-based MARL algorithm to solve the packet routing problem in WSNs so as to maximize the network life cycle}             & Dec-POMDP     & Decentralized with Networked Agents                 & DVF      & Private & Cooperation & Network life cycle maximization & WSNs                                     \\ \cline{2-10}
                                      & ~\cite{Li2020AMM}     & \textcolor{black}{Improves the QLAODV algorithm with MARL to solve the multi-hop routing problem for dynamic UAANETs}  & Dec-POMDP     & Decentralized with Networked Agents                 & Modified-QLAODV & Private & Cooperation & Balance the energy distribution and prolong the lifetime of UAV nodes &UAANET     \\ \cline{2-10}
                                      & ~\cite{Geng2020AMR} &   \textcolor{black}{Combines the intra-region and the inter-region packet routing algorithm with MARL to solve the packet routing problem for multi-region networks}                   & Dec-POMDP     & IL & DDPG            & Private & Cooperation & Maximum edge cost minimization & Multi-region network                            \\ \hline
\end{tabular}
\end{table*}

\section{Trajectory Design for UAV-Aided Networks}\label{Trajectory Design for UAVs}

\begin{figure}[h]
  \centering
  \includegraphics[scale=0.8]{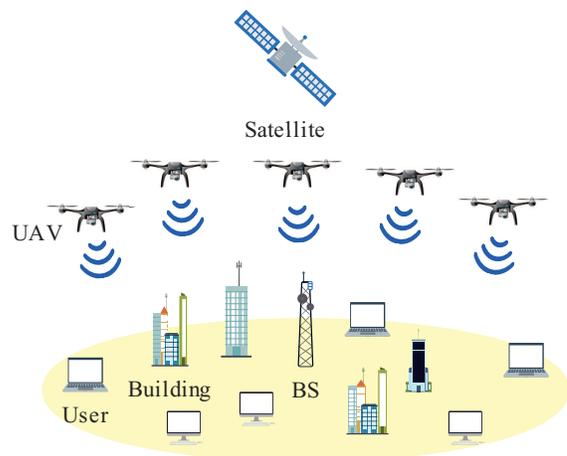}
  \caption{An UAV-aided communication network, where the satellite can provide location information for UAVs, and each UAV provides communication service to ground users and aims to maximize its own throughput~\cite{Ding2021TrajectoryDA}, or overall system throughput~\cite{Zhong2020MultiAgentRL}.}
  \label{UAV}
\end{figure}

In UAV-aided communication networks as shown in Fig. \ref{UAV}, the mobility of UAVs is capable of providing additional degrees of freedom (DoF) for wireless networks that may lead to a better QoS for ground users~\cite{liu2019comp}. However, this requires well-designed trajectories of the UAVs. To be specific, the trajectory design enables the UAV to dynamically move around within the service range so as to improve the network performance, e.g., network throughput. This problem is non-convex and typically combined with network resource allocation problem~\cite{shi2019multi}, e.g., the joint optimization of UAVs trajectory design and power allocation. Some centralized methods such as~\cite{cui2019multiple} and~\cite{shi2019multi} aim to divide the source trajectory design problem into sub-problems for simplification. However, these centralized methods usually require the complete instantaneous network state information which causes a large amount of communication overhead for state information collection. Furthermore, the environment of the UAV network is highly dynamic that requires an adaptive trajectory design algorithm. MARL can totally be used to learn collaborative, energy-efficient, and decentralized trajectory design policies for UAVs that can adapt to the dynamics of the UAV environment.

The authors in~\cite{Ding2021TrajectoryDA} investigate the joint access control of ground users (GUs) and trajectory design of UAVs. The system model consists of multiple BS-enabled UAVs flying horizontally to provide communication service to GUs. The GUs compete with each other on the limited frequency resource to maximize their own throughput, while the UAVs can move and cooperate with each other to maximize their fair throughput subject to a safe distance. Thus, a mixed cooperative-competitive game is used to model the coordination among the GUs and the UAVs. \textcolor{black}{To deal with such a mixed game, the authors proposed an MADDPG based AG-PAMADDPG algorithm to coordinate the UAVs and GUs since each MADDPG agent has a unique actor and critic, and can deal with different reward structures. How MADDPG works is explained in Section \ref{MARLalgorithms}.} The UAVs can obtain the location information of all the UAVs and GUs by exchanging messages with other UAVs, while the GU can only observe its local state, e.g., the signal power, the connected UAV, and the throughput in the previous time slot. The reward functions of GUs and UAVs are defined as their throughput and the network fair throughput, respectively. The network fair throughput enables the UAVs to achieve a better balance between the total throughput and the fairness among the GUs. The action of each UAV is to select a moving direction, and the action of each GU is to select a frequency resource. Thus, the action space of the UAV is continuous, while the action space of the GU is discrete. To deal with the hybrid action space, the AG-PAMADDPG transforms the discrete action of GUs to a continuous probability distribution and samples the action from the distribution. Numerical results show that AG-PAMADDPG outperforms the distributed DQN algorithm~\cite{cao2019deep}, and the proposed AC-PAMADDPG that only optimizes the access control of GUs, in terms of network throughput and GUs' fairness.

In a user-intensive network with an overloaded BS, the UAV-aided cellular offloading technique can leverage aerial base stations (ABSs) carried UAVs to improve the network connectivity. The authors in~\cite{Zhong2020MultiAgentRL} investigate the joint optimization of UAV trajectory design and power allocation in such a UAV-aided cellular offloading scenario. Each UAV has a single antenna and shares the same frequency band. Different from~\cite{Ding2021TrajectoryDA}, in~\cite{Zhong2020MultiAgentRL}, the roaming users are grouped into different clusters, and each UAV is associated with one cluster.

Each UAV's problem is to learn a joint trajectory design and power allocation policy to serve the users in its cluster for maximizing the overall network throughput under the user fairness constraint. The mobility of both UAVs and users makes the problem difficult to be solved by convex optimization. To solve the problem, the authors propose a multi-agent MDQN algorithm to learn such a policy. The agent is the UAV. The state of each agent is an array that includes its 3D coordinates and those of other agents as well as the corresponding channel gains of their associated users. The common reward for all the UAVs is defined as the overall data rate of UAVs. Besides, a penalty coefficient is introduced to the reward function to guarantee user fairness. Due to the mobility of users, the upper-bounded K-means method~\cite{cui2018unsupervised} is used to periodically partition the users into multiple clusters. Based on these clusters, all the UAVs share a common DQN to design trajectory and allocate power to users in their serving clusters, and the common DQN uses all the UAV's experiences for centralized training, which can accelerate the convergence speed and stabilize multi-agent environments. To simplify the action selection, the action space is represented by a combination of discrete trajectory directions and power levels. The simulation results show that the proposed MDQN has a faster convergence speed than the benchmark conventional DQN in the multi-agent scenario. Also, by leveraging the MDQN to design the 3D trajectory, the sum rate of the overall network is capable of achieving $142$\% and $56$\% gains over those of employing the circular trajectory and the 2D trajectory, respectively. However, the centralized training of MDQN may occur a large amount of communication overhead for collecting UAVs' experiences.

Recently, UAVs can be used as edge servers that are able to provide computation offloading services to the GUs. However, due to the dynamic change of GUs’ locations and task loads, the UAVs have to learn trajectory management to adapt to the dynamics of the environment. To the end, the authors in~\cite{Wang2021MultiAgentDR} propose MADRL-based trajectory management for each UAV to jointly maximize regional fairness of GUs and the GU-load fairness of UAVs, while minimizing the total consumption of GUs in multi-UAV aided MEC networks. The trajectory design problem of the UAV is modeled as a Dec-POMDP, and the agent is the UAV. The UAV can make decisions on the flying direction and distance depending on its observation, which includes (i) the coordinates of the current UAV, (ii) the relative distance between the current UAV and other UAVs, (iii) the cumulative service times for GUs, and (iv) the historical GU-load of current UAV. Then, the UAVs learn the trajectory design by adopting the MADDPG algorithm with the CTDE framework. The environment can integrate all the UAVs’ observations and actions as the system state, and then feedback the system state to all the UAVs for local off-line training, which can improve the collaboration performance of UAVs. However, this occurs high computational overhead in each UAV. After the movement of UAVs, the UE selects a UAV that consumes the least energy to offload its task. The simulation results show that the proposed MADDPG algorithm has a significant performance gain compared with the circular trajectory and random trajectory. However, the proposed MADDPG is not scalable since it depends on the number of UAVs in the network. 

The authors in~\cite{Wu2020CellularUC}, ~\cite{Wu2020AoIMF} and~\cite{Hu2020CooperativeIO} consider using MARL for UAV trajectory design in UAV sensing networks. The authors in~\cite{Wu2020CellularUC} consider a UAV sensing scenario. The system model consists of a single BS and multiple UAVs performing sensing tasks. Each UAV needs to sense its target at the sense stage and transmit the sensory data to the corresponding terrestrial mobile device directly (the D2D mode) or through the relay of the BS (the Cellular mode) at the transmission stage. To avoid mutual interference among UAVs, the BS is responsible for allocating available orthogonal sub-channels to the UAVs for transmitting data to the terrestrial mobile device. The UAVs with the allocated sub-channel aim to transmit data to the terrestrial mobile device using the transmission mode that can satisfy the QoS requirement. When a UAV successfully transmitted its data to the mobile device in the previous time slots, the BS will not allocate the sub-channel to the UAV in the current and future time slots.

Due to the limited sub-channel resources, the authors propose a sub-channel allocation method to select a number of UAVs with the highest Successful Transmission Probabilities (STPs) that have not successfully transmitted data to the mobile device, to allocate the sub-channel resources. The STP is measured by the signal-to-noise ratio (SNR) of the transmission links, and it is a function of the locations of the UAV and the BS (for the Cellular mode) or the mobile device (for the D2D mode). The probability that the UAV can complete the sensing task is evaluated by the Successful Sensing Probability (SSP), which is a function of the locations of the UAV and the sensing target. Note that both the STP and the SSP are determined by the locations of the UAVs. Thus, each UAV's problem is to learn an optimal trajectory design policy so as to maximize its utility that is defined as the total number of valid transmissions, i.e., both the sensing task and the transmission task are successful. To solve the problem, the authors adopt the DQN algorithm to design a trajectory for each UAV. The state of each UAV includes the locations of all the UAVs obtained from the BS at the beginning of each time slot. To reduce the action space of the UAV in a 3D environment, a lattice model is introduced to discretize the continuous action space of UAVs. The reward for each UAV is defined as the sum of the valid transmission probability during the transmission. After the UAVs get the valuable spectrum, they can choose the transmission mode with higher STP to transmit sensory data for higher throughput. The simulation results show that the proposed algorithm can achieve a higher total utility than the single-agent $Q$-learning algorithm~\cite{hu2018reinforcement} and policy gradient algorithm~\cite{Sutton1999PolicyGM}.

In the above approaches, e.g.,~\cite{Wu2020CellularUC}, the sub-channel resource constraint can result in the poor Age-of-Information (AoI)~\cite{kaul2012real}, which is an important metric to represent the freshness of the data of the sensing target. The authors in~\cite{Wu2020AoIMF} investigate the AoI minimization problem by optimizing the UAVs' trajectories for both the sensing and transmitting stage. Here, AoI is defined as the time steps since the last valid data transmission of the UAV is completed. Each UAV senses its target and then transmit their sensory data to the BS or terrestrial mobile devices. Particularly, some UAVs transmit data to the BS over U2N links, while the others transmit data to the mobile device over U2D links. To reduce the interference, the BS allocates an exclusive sub-channel to each U2N link for data transmission. To improve the sub-channel efficiency, the U2D links work in the underlay mode and share the same sub-channels with the U2N links. The BS allocates the sub-channel resources to the UAVs by calculating a sub-channel allocation matrix that can maximize the sum of expected data size to be transmitted. Specifically, such a sub-channel allocation matrix is only determined by the locations of the UAVs. Thus, each UAV has to compete with others for sub-channel resources to minimize its accumulated AoI over future time slots by designing its trajectory. The problem is modeled as a Dec-POMDP, where the UAV is an agent. 

The state of each UAV is defined as the current time slot index, the sensing and transmitting locations, the remained data size, a stage indicator (sensing stage or transmission stage), and the current AoI. The UAV's action includes the sensing and transmission locations. When the action is taken, the UAV flies to the sensing location to finish the sensing task and then moves to its transmission location for data transmission. Instead of using the DQN algorithm like~\cite{Wu2020CellularUC}, the authors propose to adopt the DDPG algorithm to solve the multi-UAV trajectory design problem. Both the actor network and critic network of each UAV can use the states of all the UAVs which can be obtained from the BS as the input to improve algorithm stability. The actor learns a trajectory policy towards maximizing the long-term reward which is defined as the negative value of AoI. The simulation results show that the proposed DDPG can achieve a considerable performance gain in terms of the average AoI compared with both the policy gradient algorithm~\cite{Sutton1999PolicyGM} and the greedy algorithm.

Different from the DDPG used in~\cite{Wu2020AoIMF}, the authors in~\cite{Hu2020CooperativeIO} propose a CA2C method to minimize the sensing tasks’ AoI in multi-UAV sensing scenarios. The system model consists of a BS, multiple UAVs, and multiple sensing tasks. Each task involves a sensing target. The UAVs first sense the target and then transmit the sensory data to the BS for further process. The spectrum resource is assumed to support all the UAVs' data transmissions. Thus, different from~\cite{Wu2020CellularUC}, the authors only need to consider the SSP, which is highly related to the sensing locations of the UAVs. As a consequence, the UAV's problem is to select a sensing task and the corresponding sensing locations so that it can complete the sensing task to minimize the accumulated AoI of the tasks over a certain period. The problem is formulated as an MG in which the agent is the UAV. The state includes the index of the time slot, the locations of UAVs at the beginning of the time slot, the amount of sensing data of each UAV to be transmitted, the AoI of all the tasks, and the selected task and sensing locations of each UAV in the previous time slot. For cooperation, a common reward for all UAVs is defined as the AoI reduction achieved by the selected actions. 

Each UAV adopts the MARL-based CA2C method to learn the joint task selection and trajectory design policy. Thus, the hybrid action for each UAV includes the discrete task selection action and the continuous trajectory design action. To deal with such hybrid action, the CA2C first inputs the system state and a one-hot indicator for a task to the actor network to calculate the sensing location of the input task, then the sensing location and the system state are fed to the critic network to estimate the $Q$-value of the input task. After estimating the $Q$-values of the tasks, the task with a maximum $Q$-value will be selected by the UAV for execution. The simulation results show that the proposed CA2C algorithm achieves a better performance in terms of the normalized accumulated AoI compared with the baseline algorithms such as DDPG, DQN, and the greedy algorithm. Moreover, cooperative UAVs can obtain a better AoI compared with non-cooperative UAVs.

\textbf{Summary: }This section reviews the applications of MARL for UAV trajectory design in both UAV-aided cellular networks and multi-UAV sensing networks. We observe that the UAV trajectory design is usually coupled with transmit power allocation and frequency resources or sub-channel allocation. These joint optimization issues are typically solved by the CTDE or IL-based MARL methods such as multi-agent MDQN, DQN, DDPG, and CA2C, and they mostly can achieve the locations of all the UAVs from the BS to make action decisions. This is beneficial for the UAVs to solve the non-stationarity issue. We may note that the centralized training of CTDE mostly occurs in the local UAV. However, the UAVs may have limited computing capacity to efficiently deal with complex state information and learn an MARL model. Besides, the energy consumption of UAVs also needs to be taken into consideration to design an energy-efficient trajectory policy.

%table VIII

\begin{table*}[]\scriptsize\centering
\renewcommand\arraystretch{1.5}
\caption{A summary of approaches applying MARL for trajectory design for UAV-aided networks.}
\label{tab:trajectory}
\begin{tabular}{|m{0.5cm}<{\centering}|m{0.6cm}<{\centering}|m{3cm}<{\centering}|m{1.5cm}<{\centering}|m{1.5cm}<{\centering}|m{1.5cm}<{\centering}|m{1cm}<{\centering}|m{1.5cm}<{\centering}|m{2cm}<{\centering}|m{1cm}<{\centering}|}
\hline
\rowcolor[HTML]{9B9B9B}
\textbf{Issues}            & \textbf{Works}      & \textbf{Main Findings}                    & \textbf{Model}     & \textbf{Learning Schemes}      & \textbf{Algorithms}                & \textbf{Private or Shared Policy} & \textbf{Cooperation or Competition}  & \textbf{Learning Objectives}   & \textbf{Network Models}                \\ \hline
\multirow{6}{*}{{\rotatebox{90}{Trajectory design for UAV-aided networks\quad \quad \quad \quad \quad \quad \quad \quad}}} 
                                                   & ~\cite{Ding2021TrajectoryDA}   & \textcolor{black}{The joint access control of GUs and trajectory design of UAVs is investigated} & Dec-POMDP & CTDE                 & AG-PAMADDPG  & Private                  & Mixed &  Gu's throughput maximization for GUs and network fair throughput maximization for UAVs & Cellular network     \\ \cline{2-10}
                                                   & ~\cite{Zhong2020MultiAgentRL}   & \textcolor{black}{Proposes a joint trajectory design and power allocation policy for each UAV to serve the partitioned mobile users}   & Dec-POMDP                            & CTDE          & Multi-agent MDQN      & Shared                   & Cooperation                   &Network throughput maximization & Cellular network     \\ \cline{2-10}
                                                   & ~\cite{Wang2021MultiAgentDR}  & \textcolor{black}{Solves the trajectory design problem for an MEC-enabled network with MARL}      & Dec-POMDP                          & CTDE                 & MADDPG    & Private                  & Cooperation                   & Maximizing regional fairness of UEs and the UE-load fairness of UAVs, while minimizing the total consumption of UEs& Multi-UAV assisted MEC network \\ \cline{2-10}
                                                   & ~\cite{Wu2020CellularUC}  & \textcolor{black}{Considers the trajectory design in UAV sensing networks where each UAV has to change its location to complete the transmission and sensing tasks}      & POMDP                            & IL & DQN       & Private                  & Competition                &The total number of valid transmissions maximization   & UAV sensing network            \\ \cline{2-10}
                                                   & ~\cite{Wu2020AoIMF}   & \textcolor{black}{Studies minimizing the AoI in UAV sensing networks by optimizing the trajectory of the UAV}             & POMDP                            & CTDE & DDPG      & Private                  & Competition              &AoI minimization     & UAV sensing network            \\ \cline{2-10}
                                                   & ~\cite{Hu2020CooperativeIO} & \textcolor{black}{Jointly optimizes task selection and trajectory design for UAV sensing networks with MARL}     & MG                            & IL & CA2C      & Private                  & Cooperation          & Cumulated AoI minimization         & UAV sensing network            \\ \hline
\end{tabular}
\end{table*}

\section{Network Security}\label{Network Security}

Cyber attacks such as eavesdropping attacks, jamming attacks, and distributed denial of service (DDoS) are common security issues in the future internet. Recently, adaptive RL-based cyber-attack defending approaches, e.g.,~\cite{Han2017TwodimensionalAC},~\cite{Liu2018AntiJammingCU}, have been proposed that allows legitimate network entities to interact with the network environments to learn an optimal policy for secure communication. However, such an approach aims to learn an optimal security policy for a single network entity and does not account for other entities' security policies, which may result in a local optimal policy. On the contrary, MARL enables multiple network entities to cooperate with each other to effectively combat the attacks. As a result, MARL has recently been proposed as a promising solution to the attacks in the future Internet. In this section, we review applications of MARL for network security. In particular, we present and discuss the applications of MARL for addressing common attacks such as eavesdropping attacks, jamming attacks, and DDoS attacks.

\subsection{Eavesdropping Attack}
\begin{figure}[h]
  \centering
  \includegraphics[scale=0.8]{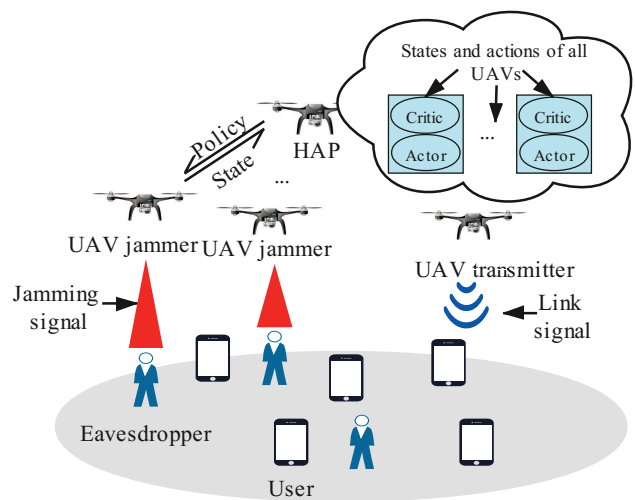}
  \caption{Eavesdropping attack defense model in UAV-aided communication networks based on MARL.}
  \label{jamming}
\end{figure}

The UAV-aided communication networks use the air-to-ground (A2G) links, which can be easily wiretapped by the Ground Eavesdroppers (GEs). The authors in~\cite{Zhang2020MultiAgentDR} propose to use an MADRL framework that enables friendly UAV jammers
to cooperate with each other to learn jamming policies to prevent the GEs from eavesdropping. The system model consists of one UAV transmitter and multiple UAV jammers. The UAV transmitter communicates with ground users (GUs) that can be wiretapped by the GEs. The UAV jammers transmit jamming signals to prevent the GEs from eavesdropping the legitimate communications. The objective is to maximize the sum secure rate of all the GUs by jointly optimizing the trajectory of the UAVs, the transmit power, and the jamming power. The MADDPG is then adopted in which the training of the policies of all the UAVs are centralized at the high altitude platform (HAP) with strong computing capability that is shown in Fig. \ref{jamming}. The state of each UAV includes (i) the locations of all the UAVs, (ii) its transmitting or jamming power, and (iii) the secure rate of a certain user. The reward for the UAV transmitter is defined as the difference between the secure rate and the transmitting power penalty, while the reward for the UAV jammer is the difference between the secure rate and the jamming power penalty. To realize the centralized training, each UAV can upload its state and action (the flying direction, the transmitting power, and the jamming power) to the HAP. The UAV can then download the trained actor network for its local action execution.

As an extension of the work in~\cite{Zhang2020MultiAgentDR}, the work in~\cite{Zhang2020UAVEnabledSC} combines the MADDPG algorithm with an attention mechanism. The attention mechanism enables each UAV, i.e., the agent, to pay attention to the important observations, which improves the learning efficiency. In particular, the UAV transmitter pays more attention to the GUs, while the UAV jammers are more likely to put more attention on the GEs and other jammers. To this end, the proposed continuous action attention MADDPG (CAA-MADDPG) adds the attention layer to its critic network, and the attention layer can assign more attention weight to the information of interest to maximize the long-term overall secure rate. As a result, the attention mechanism can significantly improve the learning efficiency, and accelerate the convergence speed of the algorithm. The simulation results show that the proposed CAA-MADDPG outperforms the MADDPG in terms of the overall secure rate.

\subsection{Jamming Attack}With the jamming attack, an attacker can transmit jamming signals to interfere with the legitimate communication channels. The legitimate users cannot use the interfered channels for reliable communication. The authors in~\cite{aref2017multi} adopt the independent $Q$-learning method to address the anti-jamming channel selection problem, where each agent learns independently to select a channel for communication and simply treats other agents as a part of the environment. Thus, it may fail to achieve a global optimal solution. 

To address the issue, the authors in~\cite{Yao2019ACM} propose to adopt the MARL algorithm to provide a cooperative anti-jamming solution in multi-user environments. The system model consists of multiple legitimate users and one jammer. There is a set of channels that are available to users, and the jammer launches a jamming attack to one channel at each time slot. Meanwhile, co-channel interference occurs when more than one user chooses the same channel. To improve the reliability of communications, the authors model the sequential channel-selection processes of legitimate users as an MG to jointly optimize the co-channel interference and anti-jamming to maximize the user's long-term reward. In the MG, each user is treated as an agent and a multi-agent $Q$-learning algorithm is then adopted to learn a decentralized anti-jamming policy for each agent. In this setting, the $Q$-value of each agent is determined by the observed jamming channel and the joint action of all the agents. The reward for each agent is equal to "$1$" if there is no co-channel interference and jamming happens in the selected channel, otherwise, it is set to "$0$". To obtain the global optimal joint action, the agent can exchange $Q$-values with other agents, and calculate the sum $Q$-values of all the agents. Then, the joint action that can maximize the sum $Q$-values will be selected as the optimal action. The simulation results show that the multi-agent $Q$-learning can achieve a better-normalized rate compared with the independent $Q$-learning method~\cite{aref2017multi} and the sensing-based method.

The multi-agent $Q$-learning method used in~\cite{Yao2019ACM} can also be found in~\cite{Xu2018InterferenceAwareCA} to deal with the cooperative anti-jamming channel selection problem in UAV communication networks, where one UAV cluster is denoted as a user. Each user can sense the channel interference and aim to learn an optimal anti-jamming channel selection policy so as to maximize its utility. Here, the utility of each user is a function of the user’s throughput and the cost of channel switching and cooperation. To solve the anti-jamming issue, according to the co-channel interference level, the collaborative manner or the independent manner is adopted to learn the optimal anti-jamming channel selection policy for the user. Specifically, if the sensed co-channel interference energy exceeds the defined threshold, the users adopt the multi-agent $Q$-learning algorithm to learn a cooperative anti-jamming channel selection policy, where the users can share the joint $Q$-tables with each other to realize the coordination. The $Q$-value in the joint $Q$-table is estimated by the system state (all the users' transmission channels and the jamming channel) and the joint actions of all the users. Otherwise, if the sensed co-channel interference energy does not reach the threshold, then it can be ignored. Each user can learn the optimal policy independently since the user is only influenced by the anti-jamming interference, and it no longer interacts with other users. The simulation results demonstrate that the proposed $Q$-learning-based algorithm outperforms the single $Q$-learning method and the sensing-based method in terms of the normalized utility of the user.

The anti-jamming in the ultra-dense IoT networks is investigated in~\cite{Wang2020MeanFR}. The authors consider an MEC-enabled ultra-dense IoT network that is divided into multiple edge subnetworks. Each subnetwork has an edge server (ES) that can provide computing services for multiple IoT devices. Each ES has the spectrum sense ability and is connected with a cloud server via the wire connection. The problem of each ES is to learn an optimal anti-jamming channel selection policy so as to maximize its transmission rate. \textcolor{black}{Considering the high density of ESs, a mean-field DeepMellow method is proposed in which the user's joint $Q$-function is approximated by a mean-field $Q$-function~\cite{yang2018mean}, where a mean-field action is used to approximate the joint actions of other ESs. How the mean-field $Q$-function handles a large amount of IoT devices can be found in Section \ref{MARLalgorithms}.} This can significantly reduce the computational complexity and improve the scalability of the algorithm. Furthermore, different from~\cite{yang2018mean}, the mellowmax function~\cite{Kim2019DeepMellowRT} is used to estimate the mean-field value function of the next state in the process of updating the $Q$-value to reduce the estimation bias of the $Q$-value. The system state includes the history of the spectrum sensing results. To deal with such a huge system state, the $Q$-function is appropriated by a DNN. All the ESs share a centralized DNN that is trained in the cloud server by using the experiences of all ESs. The simulation results show that the proposed mean-field DeepMellow algorithm is able to obtain a higher average reward compared with the mean-field $Q$-learning~\cite{yang2018mean} and the individual $Q$-learning~\cite{mnih2015human-level}.

\subsection{DDoS Attack} A DDoS attack is a distributed, dynamic, and large-scale cyber attack that can exhaust the network resources of the target system. An attacker can launch the DDoS attack through maliciously tampering and manipulating a large number of compromised nodes, i.e., bots, sending large amounts of network traffic to the target system. This can cause the target system to be paralyzed and result in a denial of service to legitimate requests. The DDoS attack can be hard to defend since the malicious bots can imitate legitimate network users. Moreover, the distributed and dynamic nature of DDoS may also impose a challenge to DDoS defense.

The authors in~\cite{Xia2019ANS} and~\cite{XiaShimingBaiDefendingNetworkTraffic} propose to adopt the MARL framework for collaborative DDoS defending. The system model involves hosts, routers, and servers. The hosts can act as attackers or legitimate users. The routers can decide how many packets are received from hosts or other routers to be forwarded to the next-hop or servers. The authors define aggregate traffic as the packets received by routers or servers in a period to evaluate the ability of routers or servers. The aggregate traffic may include both the legitimate packets and the malicious packets. The objective is to maximize the average legitimate traffic rate of all the routers while the aggregate traffic received by servers should not exceed the load limit. The problem is modeled as an MG, where each router is treated as an agent. The action of each agent is to decide whether to send the aggregate traffic. The authors in~\cite{Xia2019ANS} propose to use a fully centralized collaborative method, namely Centralized Reinforcement Learning Router Throttling (CRLRT) to train the agents. Each agent first transmits its aggregate traffic information to the central DDPG agent for joint action decision-making. Then, the joint action is distributed to the specific agent for execution. The simulations show that the proposed CRLRT can achieve a better legitimate traffic rate than that of Multi-agent Router Throttling (MART)~\cite{malialis2013multiagent} that only depends on its local information without communicating with other routers under fixed attack location settings as well as dynamic attack location settings. Besides, it outperforms the rule-based algorithms such as Server-initiated Router Throttling (SIRT) and Fair Router Throttle (SRT) proposed in ~\cite{yau2005defending}. However, frequently exchanging information with the central DDPG may incur huge communication overhead and increase the decision-making latency.

To overcome the issues of CRLRT, the authors in~\cite{XiaShimingBaiDefendingNetworkTraffic} introduce an MADDPG-like decentralized ComDDPG method to train agents. Different from MADDPG, the ComDDPG consists of multiple decentralized actors for action execution and one central critic for centralized training. During the offline training, the critic can make use of the global state including the traffic information of both routers and servers, the residual resources, and the waiting queue length of servers, to train the agents’ decentralized policies. After the training process finishes, each agent only needs its aggregate traffic information to make online decisions and no longer needs to communicate with the central critic, which significantly reduces the signaling overhead and improves the decision efficiency. The simulation results show that the proposed ComDDPG outperforms the distributed DQN demonstrating the significance of communication. Moreover, it can also achieve similar performance to the centralized DDPG.

\textbf{Summary: }This section reviews the applications of MARL to combat common security issues in the future internet including eavesdropping attacks, jamming attacks, and DDoS attacks. The reviewed works are summarized in Table \ref{tab:security}. As seen from the table, most of the works adopt the CTDE framework to learn a decentralized defense policy in networks such as wireless networks, IoTs, and Internet, since centralized policy may incur high computational complexity to make action decisions for all the network entities, e.g., routers, UAVs, or IoT devices, especially in large scale networks. We may also note that almost all the works adopt a cooperative manner to defend the cyber-attacks by enabling network entities to share information with others, which can result in a better global result than independent learning. However, the communication for cooperation among network entities may not be trusty and reliable under cyber-attacks, thus, safe cooperation should also be taken into account to design an ultra-reliable defense policy.

%IX

\begin{table*}[]\scriptsize\centering
\renewcommand\arraystretch{1.5}
\caption{A summary of approaches applying MARL for network security.}
\label{tab:security}
\begin{tabular}{|m{0.5cm}<{\centering}|m{0.6cm}<{\centering}|m{3cm}<{\centering}|m{1.5cm}<{\centering}|m{1.5cm}<{\centering}|m{1.5cm}<{\centering}|m{1cm}<{\centering}|m{1.5cm}<{\centering}|m{2cm}<{\centering}|m{1cm}<{\centering}|}
\hline
\rowcolor[HTML]{9B9B9B}
\textbf{Issues}            & \textbf{Works}      & \textbf{Main Findings}                    & \textbf{Model}     & \textbf{Learning Schemes}      & \textbf{Algorithms}                & \textbf{Private or Shared Policy} & \textbf{Cooperation or Competition}  & \textbf{Learning Objectives}   & \textbf{Network Models}                \\ \hline
\multirow{7}{*}{{\rotatebox{90}{Network security\quad \quad \quad \quad \quad \quad \quad \quad \quad \quad \quad \quad \quad \quad \quad}}} & ~\cite{Zhang2020MultiAgentDR}  & \textcolor{black}{Proposes a cooperative MADRL algorithm to learn jamming policies for friendly UAV jammers to prevent the GEs from eavesdropping}      & Dec-POMDP & CTDE                                                                         & MADDPG     & Private                                                         & Cooperation                                               & Maximizing the sum secure rate of all the GUs                & UAV network                      \\ \cline{2-10}
                                  & ~\cite{Zhang2020UAVEnabledSC}  & \textcolor{black}{Applies attention mechanism to the trajectory design of friendly UAV jammers to improve the learning efficiency}     & Dec-POMDP & CTDE                                                                         & CAA-MADDPG & Private                                                         & Cooperation                                                           & Maximizing the sum secure rate of all the GUs & UAV network                      \\ \cline{2-10}
                                  & ~\cite{Yao2019ACM}         & \textcolor{black}{Proposes a cooperative anti-jamming MARL algorithm for multi-user environments}        & MG    & CTDE           & multi-agent $Q$-learning & Shared    & Cooperation                                                  & Long-term reward maximization            & Wireless network \\ \cline{2-10}
                                  & ~\cite{Xu2018InterferenceAwareCA}   & \textcolor{black}{Investigates the anti-jamming channel selection problem with MARL for UAV communication networks} & MG   & CTDE if the interference beyond the threshold, otherwise, IL & multi-agent $Q$-learning & Shared if the interference beyond threshold, otherwise, private & Cooperation if the interference beyond threshold, otherwise, competition & User's utility maximization & UAV network                      \\ \cline{2-10}
                                  & ~\cite{Wang2020MeanFR}    & \textcolor{black}{Solves the anti-jamming channel selection problem using MARL in MEC-enabled ultra-dense IoT networks}               & MG    & CTDE      & mean-field DeepMellow & Shared               & Cooperation                                                          & Transmission rate maximization   & IoT \\ \cline{2-10}
                                  & ~\cite{Xia2019ANS} & \textcolor{black}{Proposes a fully centralized MARL framework for cooperative DDoS defending} &MG    & Fully centralized                                                                  & CRLRT       & Shared                                                          & Cooperation                                                          &Average legitimate traffic rate maximization    & Internet  \\ \cline{2-10}
                                  & ~\cite{XiaShimingBaiDefendingNetworkTraffic}  &  \textcolor{black}{Considers decentralized DDoS defending where each agent has a decentralized actor and shares a central critic with other agents}              & MG    & CTDE                        & ComDDPG    & Private                                                         & Cooperation                                                     &Average legitimate traffic rate maximization       & Internet  \\ \hline
\end{tabular}
\end{table*}

\section{Challenges, Open Issues, and Future Research Directions}\label{Challenges_Open}
In this section, we discuss challenges, open issues, and future research directions of applying MARL to the future Internet. 
\subsection{Challenges}
\emph{\textbf{1) Synchronization of multiple agents in MARL: }}The centralized training of MARL raises a requirement of the state synchronization among agents. To synchronize the states of all the agents, an ideal collection thread can be deployed~\cite{He2020DeepHopOE}. However, the future networks become distributed and decentralized, and thus it is difficult to maintain such a thread that can synchronize the global state information for each learning agent at the same time. Furthermore, the unreliability of wireless communications, e.g., of V2V links and IoT links, imposes a challenge to the synchronous communications of state information exchange among the agents. As such, the agents may need to use old state information for the learning that can lead to a non-optimal policy. 

\emph{\textbf{2) Learning performance loss caused by action selection: }}To simplify action space and speed up the learning convergence, the current works typically discretize the continuous action space, such as transmit power of communication links and moving directions of UAVs. This can reduce the learning performance due to the fact that the discrete action space may not always include the optimal actions. Although the policy-based algorithm can work with the continuous action values, the unreasonable output mechanism, e.g., the unreasonable action clip mechanism for file amount selection proposed in~\cite{Pedersen2021DynamicCC}, can still reduce the learning performance of the agent. For example, the action performance loss for coded caching is well analyzed and discussed in~\cite{Wu2020MultiAgentRL}.

\emph{\textbf{3) Malicious packet knowledge: }}In security solutions, such as~\cite{Xia2019ANS} and~\cite{XiaShimingBaiDefendingNetworkTraffic}, MARL is used to enable each router to learn a collaborative traffic control policy that aims to maximize the average legitimate traffic rate of all the routers. For this, the rewards in these solutions are defined based on the assumption of the knowledge of malicious packets. However, it is not practical in reality, the malicious packets are difficult to be figured out by the router since they can imitate the legitimate packets through leveraging some hacking tools. This imposes a challenge on the application of MARL to address security issues in real networks.

\emph{\textbf{4) Privacy and security of the use of MARL: }}To coordinate multiple agents, e.g., mobile users, in MARL, information sharing among the agents is required. However, information-sharing concerns the user's privacy and information security. In particular, the states (observations) shared among agents may contain sensitive information such as the user's task profile, battery level, CPU utilization, and locations, and this information can be easily hacked by malicious attackers~\cite{Pan2019HowYA} during transmission. Besides, the DNN based MARL model (i.e., MADRL) is vulnerable to adversarial attacks since the DNN can be easy to confuse and make harmful action decisions when the input of the DNN is maliciously tampered~\cite{goodfellow2014explaining}. Thus, to construct a credible and safe MARL model in real-world network applications, the vulnerability of MARL to adversarial attacks needs to be addressed.

\subsection{Open Issues}
\emph{\textbf{1) Multi-agent state information extraction: }}With MARL, each agent can use the current or historical state information, e.g., from its own observation or other agents' observations, for its action selection. However, using directly the state information without pre-processing may lead to poor convergence of MARL. Thus, state information extraction is an important task that pre-processes the network information to waken the unrelated or harmful state information. There exist several techniques for state information extraction in reviewed works, such as GNN~\cite{He2020ResourceAB} and attention mechanism~\cite{Kaviani2021RobustAS}. The GNN is a promising technique for MARL for the reason that it can simply abstract the agent as a node and the relationship between agents as the edge, which can efficiently describe the multi-agent environment state. However, with the consideration of other agents, these state information extraction techniques typically need to frequently collect other agents' information that may cause high communication overhead. Thus, an efficient state information extraction design for network communication is still an open issue.

\emph{\textbf{2) Mixed reward design and multi-objective optimization: }}A well-designed reward function can significantly improve the agent's performance and accelerate the learning speed. However, the future networks can consist of multiple network entities, e.g., BSs and users, with different objectives, e.g., throughput maximization or energy efficiency maximization. Thus, an effective reward design is still an open issue. The reviewed works typically set a global reward (i.e., a common reward) for all the agents or a local reward for an individual agent. Both the global reward and the local reward have advantages and shortcomings: the global reward can encourage collaboration among agents, but may also breed lazy agents, while the local reward motivates each agent to earn more rewards, but intensifies the fierce competition for the limited resource. Given these reasons, the mixed reward that combines the global reward and the local reward is proposed, which can improve the performance of learning agents~\cite{Mao2020RewardDI}. Furthermore, the conflicting objectives challenge the reward design, which is common in network transmission objectives such as maximizing the network throughput while minimizing the interference to each other. Hence, it is critical to design an appropriate reward to optimize the trade-off between different objectives.

\subsection{Future Research Directions}

\emph{\textbf{1) MARL for spectrum management in NB-IoT: }}NB-IoT is a promising cellular technique that can provide low cost, low power, and long-range connectivity to a large amount of IoT devices~\cite{ratasuk2016nb}. However, such a massive number of IoT devices can result in serious co-channel interference. Moreover, serious interference can also come from the existing networks, e.g., 5G. As a result, reliable communications in NB-IoT cannot be guaranteed. This raises the spectrum management for NB-IoT. In this case, MARL can be used as an effective solution to the spectrum management that allows each IoT device to decide the spectrum so as to minimize interference.

\emph{\textbf{2) MARL for testing attacks on blockchain incentive mechanisms: }}Blockchains require participants to consume substantial storage, computation, and electricity resources to guarantee the correctness and activity of other users' transactions. Thus, most blockchains are based on incentive mechanisms to motivate users to participate in blockchain consensus protocols, where the users are paid by the cryptocurrency. Therefore, incentive mechanisms are necessary to sustain the operation of permissionless blockchains. However, the attacker can exploit the bugs of poorly-designed incentive mechanisms to achieve illegal profits. Multiple works investigate such attacks to cryptocurrency~\cite{Carlsten2016OnTI}. However, they mostly focus on theory analysis. MARL can be used to test the vulnerabilities of the incentive mechanism. The reason is that it allows rational users to interact with both unknown incentives and other users to learn the Nash equilibrium, which can discover the potential vulnerabilities that can be exploited by attackers to gain more profits.

\emph{\textbf{3) MARL for auction: }}Action theory has been shown its effectiveness in solving the spectrum resource allocation in wireless networks~\cite{wang2017auction}. In particular, the auction enables each agent, e.g., the mobile user, to make the spectrum selection from the perspective of economics, where the mobile users act as buyers and bid for the spectrum resource based on their demands. However, the classic auctions do not allow the agents to learn their optimal decisions from the environment. Fortunately, MARL enables each agent to learn an optimal auction policy not only from the environment but also from other bidders. Therefore, MARL can be used for auction in multi-agent environments.

\section{Conclusions}\label{Conclusions}
In this paper, we have presented a comprehensive survey of the applications of MARL in the future internet. First, we have introduced the backgrounds, challenges, and advanced algorithms of single-agent reinforcement learning and multi-agent reinforcement learning. Afterward, we have reviewed, analyzed, and compared the recent research contributions that use MARL to address the emerging issues in the future internet. These issues involve network access, transmit power control, computation offloading, content caching, packet routing, trajectory design for UAV-aided networks, and network security. Finally, we have outlined the key challenges, open issues, and future directions that can help the readers to further explore research topics in this area.

\bibliographystyle{IEEEtran}
\bibliography{cite}

\end{document}